\documentclass[a4paper,11pt]{article}
\usepackage{latexsym,amssymb,enumerate,amsmath,epsfig,amsthm,dsfont,bm}
\usepackage[margin=1in]{geometry}
\usepackage{setspace,color}
\usepackage{tikz}
\usepackage{floatrow}
\usepackage{graphicx,subfig}
\usepackage[ruled]{algorithm2e}
\usepackage{algorithmic}

\usepackage{epstopdf}
\usepackage{grffile}

\newcommand{\bp}{\mathbf{p}}

\newcommand{\RR}{\mathds{R}}

\newcommand{\bu}{\mathbf{u}}
\newcommand{\bv}{\mathbf{v}}
\newcommand{\bx}{\mathbf{x}}
\newcommand{\bq}{\mathbf{q}}

\newcommand{\bmu}{\bm{\mu}}
\newcommand{\blambda}{\bm{\lambda}}
\newcommand{\bbf}{\mathbf{f}}
\newcommand{\bG}{\mathbf{G}}

\newcommand{\bM}{\mathbf{M}}
\newcommand{\zero}{\mathbf{0}}
\newcommand{\diver}{\mathrm{div}}
\newcommand{\Real}{\mathrm{Real}}
\newcommand{\cF}{\mathcal{F}}
\newcommand{\cS}{\mathcal{S}}
\newcommand{\cI}{\mathcal{I}}
\newcommand{\cH}{\mathcal{H}}
\newcommand{\cL}{\mathcal{L}}
\newcommand{\cV}{\mathcal{V}}

\DeclareMathOperator*{\argmin}{arg\,min}

\newtheorem{remark}{Remark}[section]

\title{A Color Elastica Model for Vector-Valued Image Regularization
}

\author{Hao Liu \thanks{School of Mathematics, Georgia Institute of Technology, Atlanta, GA 30332, USA. Email: hao.liu@math.gatech.edu.}
	, Xue-Cheng Tai \thanks{Department of Mathematics, Hong Kong Baptist University, Kowloon Tong, Hong Kong. Email: xuechengtai@hkbu.edu.hk.}
	, Ron Kimmel \thanks{Computer Science Department, Technion, Haifa, Israel. Email: ron@cs.technion.ac.il.}
	, Roland Glowinski \thanks{Department of Mathematics, University of Houston, Honston, TX 77204, USA. Email: roland@math.uh.edu, and Department of Mathematics, Hong Kong Baptist University, Kowloon Tong, Hong Kong. }
}

\begin{document}
	
	\maketitle
	

		\begin{abstract}
			Models related to the Euler's elastica energy have proven to be useful for many applications including image processing. 
			Extending elastica models to color images and multi-channel data is a challenging task, as stable and consistent numerical solvers for these geometric models often involve high order derivatives.
			Like the single channel Euler's elastica model and the total variation (TV) models, geometric measures that involve high order derivatives could help when considering image formation models that minimize elastic properties.
			In the past, the Polyakov action from high energy physics has been successfully applied to color image processing.
			Here, we introduce an addition to the Polyakov action for color images that minimizes the color manifold curvature.
			The color image curvature is computed by applying of the Laplace--Beltrami operator to the color image channels.
			When reduced to gray-scale images, while selecting appropriate scaling between space and color, the proposed model minimizes the Euler's elastica operating on the image level sets.
			Finding a minimizer for the proposed nonlinear geometric model is a challenge we address in this paper.
			Specifically, we present an operator-splitting method to minimize the proposed functional.
			The non-linearity is decoupled by introducing three vector-valued and matrix-valued variables.
			The problem is then converted into solving for the steady state of an associated initial-value problem.
			The initial-value problem is time-split into three fractional steps, such that each sub-problem has a closed form solution, or can be solved by fast algorithms.
			The efficiency and robustness of the proposed method are demonstrated by systematic numerical experiments.
		\end{abstract}

	
	\section{Introduction}
	In image processing and computer vision, a fundamental question related to denoising and enhancing the quality of given images is what is the appropriate metric.
	In the literature, image regularization for gray-scale images has been extensively studied.
	One class of such models takes advantage of the Euler elastica energy defined by \cite[p.63]{glowinski2017}
	\begin{align}
		E(v)=\int_{\Omega} \left(1+\beta\left( \nabla\cdot \frac{\nabla v}{|\nabla v|}\right)^2\right)|\nabla v|d\bx,
	\end{align}
	where $v:\Omega \rightarrow \mathbb{R}^+$  is a gray-scale image given as a function on a bounded domain $\Omega\subset\mathbb{R}^2$, $d\bx=dx_1dx_2$, with $x_1,x_2$ the coordinates of the generic point $\bx$ of $\Omega$, while $a$ and $b$ are two positive scalar model parameters.
	
	Euler's elastica  has a wide range of uses in image processing, such as image denoising \cite{grimm2017discrete,tai2011fast,zhang2017fast}, image segmentation \cite{bae2017augmented,duan2014two,zhang2016new,zhu2013image}, image inpainting \cite{shen2003euler,tai2011fast,yashtini2016fast}, and image segmentation with depth \cite{esedoglu2003segmentation,nitzberg1993filtering,zhu2006segmentation}, to name just a few.
	An important image denoising model, incorporating the Euler's elastica energy, is
	\begin{align}
		\min_v \int_{\Omega} \left(1+\beta\left( \nabla\cdot \frac{\nabla v}{|\nabla v|}\right)^2\right)\nabla v|d\bx +\frac{1}{2\eta}\int_{\Omega} |f-v|^2d\bx,
		\label{eq.greyelastica}
	\end{align}
	where $f:\Omega \rightarrow \mathbb{R}^+$ is the input image  we would like to enhance and denoise.
	To find the minimizer of (\ref{eq.greyelastica}), one class of methods is based on the augmented Lagrangian method.
	In \cite{tai2011fast}, an augmented Lagrangian method was proposed for image denoising, inpainting and zooming.
	Based on the method in \cite{tai2011fast}, \cite{duan2011fast} suggested a fast augmented Lagrangian method and \cite{zhang2017fast} proposed a linearized augmented Lagarangian method.
	A split-Bregman method was suggested in \cite{yashtini2016fast} to solve a linearized elastica model introduced in \cite{bae2010graph}.
	Recently, an almost `parameter free' operator-splitting method was presented in \cite{deng2019new}; this method is efficient, robust, and has less parameters to adjust compared to augmented Lagrangian based methods.
	
	A color image is a vector-valued signal which can be represented, for example, by three RGB channels or four CMYK channels.
	While there are many papers and models that deal with gray-scale image regularization, most of them can not be trivially extended to handle color images when taking the geometric properties of the color image into consideration.
	Inspired by an early paper \cite{di1986note} that focused on regularization functionals on vector-valued images, \cite{sapiro1996anisotropic} proposed an anisotropic diffusion framework and \cite{blomgren1998color} introduced a color TV norm for vector-valued images. 
	Another generalization of the scalar TV, known as the vectorial total variation, was proposed in \cite{sapiro1996vector} and studied in \cite{goldluecke2012natural}.
	A total curvature model was proposed in \cite{tan2018color}, in which the color TV term in \cite{blomgren1998color} is replaced by the sum of the curvatures of each channel.
	A generic anisotropic diffusion framework which unifies several PDE based methods was studied in \cite{tschumperle2005vector}.
	The Beltrami framework was proposed in \cite{sochen1998general}, it considers the image as a two-dimensional manifold embedded in a five-dimensional space-color $(x,y,R,G,B)$ space.
	The image is regularized by minimizing the Polyakov action \cite{polyakov1989quantum}, a surface area related functional.
	Evolving the image according to the Euler--Lagrange equation of the functional gives rise to the Beltrami flow.
	To accelerate the convergence rate of the Beltrami flow, a fixed-point method was used in \cite{bar2007deblurring} and a vector-extrapolation method was explored in \cite{rosman2009efficient}.
	In \cite{spira2007short}, the authors used a short-time kernel of the Beltrami flow to selectively smooth two-dimensional images on manifolds.
	Later on, a semi-implicit operator-splitting method \cite{rosman2011semi} and an ALM based method \cite{rosman2010polyakov} were proposed to minimize the Polyakov action for processing images.
	In \cite{zosso2014primal}, the authors used a primal-dual projection gradient method to solve a simplified Beltrami functional.
	A fidelity-Beltrami-sparsity model was proposed in \cite{wang2013fidelity}, in which a sparsity penalty term was added to the Polyakov action.
	A generalized gradient operator for vector bundles was introduced in \cite{batard2014covariant}, which is applied in vector-valued image regularization.
	In \cite{bresson2006multiscale}, the Beltrami framework has been applied in active contour for image segmentation. 
	Finally, in \cite{roussos2010tensor,wetzler2011efficient} the Beltrami framework was extended to deal with patch based geometric flows, or anisotropic diffusion in patch space.

	Most existing models for vector-valued image regularization contain only first order derivatives, like the Polyakov action \cite{sochen1998general}, color TV \cite{blomgren1998color} and vectorial TV \cite{goldluecke2012natural}. 
	This maybe insufficient when considering image formation models containing geometric properties, like the curvature as captured in the single channel Euler's elastica model.
	The Polyakov action was shown to be a natural measure which minimization leads to selective smoothing filters for color images.
	To enrich the variety of image formation models, here, we propose a \emph{color elastica} model which incorporates the Polyakov action and a Laplace--Beltrami operator acting  on the image channels, as a way to regularize color images.
	While the Polyakov action measures the area of the image manifold, the surface elastica can be computed by applying the Laplace--Beltrami operator on the color channels.
	It describes a second order geometric structure that allows for more flexibility of the color image manifold and could relax the over-restrictive area minimization.
	The new model is a natural extension of the Euler elastica model (\ref{eq.greyelastica}) to color images, which takes the geometrical relation between channels into account.
	Compared to the Polyakov action model, the suggested model is more challenging to minimize since the Laplace--Beltrami term involves high order nonlinear terms.
	Here, we use an operator-splitting method to find the minimizer of the proposed model.
	To decouple the non-linearity, three vector-valued and matrix-valued variables are introduced.
	Then, finding the minimizer of the proposed model can be shown to be equivalent to solving a time-dependent PDE system until the steady state is reached.
	The PDE system is time-discretized by the operator-splitting method such that each sub-problem can be solved efficiently.
	To the best of our knowledge, this is the first numerical method to solve the color elastica model.
	Unlike deep learning methods, the suggested model does not require training, though the model itself could be used as an unsupervised loss in general settings.
	Numerical experiments show that the proposed model is effective in selectively smoothing color images, while keeping sharp edges. 
	
		While the model we propose may not outperform some existing methods like the BM3D \cite{dabov2007image} and some other deep learning methods \cite{lefkimmiatis2017non,yang2017bm3d,kobler2020total} on vector-valued image regularization, it manifests properties which have simple geometric interpretations of images. 
		The proposed model provides a new perspective on how to regularize vector-valued images and can be used in many other image processing applications, like image inpainting and segmentation, or as a semi-supervised regularization loss in network based denoising and deblurring. In addition, the model and algorithm proposed here could be also important for many other applications with vector-valued functions that need to be regularized. Many inverse problems, surface processing and image construction problems need this.

	The rest of this paper is structured as follows:
	In Section \ref{sec.Polyakov}, we briefly review the Polyakov action and the Beltrami framework.
	The color elastica model is introduced in Section \ref{sec.model}.
	We derive the operator-splitting scheme in Section \ref{sec.splitting} and describe its finite difference discretization in Section \ref{sec.discretization}.
	In Section \ref{sec.experiments}, we demonstrate the efficiency, and robustness of the proposed method by systematic numerical experiments.
	We conclude the paper in Section \ref{sec.conclusion}.
	
	\section{The Polyakov action}
	\label{sec.Polyakov}
	In this section we briefly review the Riemannian geometry relevant to the Polyakov action as applied to color images.
	Since the Polyakov action is adopted from high energy physics, we temporarily follow the Einstein's summation notation convention in this section.
	We first introduce the metric on the Riemannian manifold.
	Denote a two-dimensional Riemannian manifold embedded in $\RR^d$ by $\Sigma$.
	Let $(\sigma^1,\sigma^2)$ be a local coordinate system of $\Sigma$ and $F(\sigma^1,\sigma^2)=(F^1(\sigma^1,\sigma^2),...,F^d(\sigma^1,\sigma^2))$ be the embedding of $\Sigma$ into $\RR^d$.
	Given a metric $\{g_{ij}\}$ for $i,j=1,2$, of $\Sigma$ which is a symmetric positive definite matrix-valued metric tensor, the squared geodesic distance $(ds)^2$ on $\Sigma$ between two points $(\sigma^1,\sigma^2)$ and $(\sigma^1+d\sigma^1,\sigma^2+d\sigma^2)$ is defined by,
	\begin{align*}
		(ds)^2= g_{\mu\nu}d\sigma^{\mu} d\sigma^{\nu}\equiv g_{11}(d\sigma^1)^2+2g_{12}d\sigma^1 d\sigma^2 + g_{22}(d\sigma^2)^2.
	\end{align*}
	
	Assume $F$ embeds $\Sigma$ into a $d$-dimensional Riemannian manifold $M$ with metric $\{h_{ij}\},i,j=1,...,d$. If $\{h_{ij}\}$ is known, we can deduce the metric of $\Sigma$ by the pullback,
	\begin{align}
		g_{ij}(\sigma^1,\sigma^2)=h_{\mu\nu}\partial_i F^{\mu} \partial_j F^{\nu},
		\label{eq.inducedMetric}
	\end{align}
	where $\partial_i$ denotes the partial derivative with respect to the $i^{\mbox{th}}$ coordinate, $\partial_i \equiv \partial/\partial\sigma^i$.
	
	
	Consider a color image $\bbf(\sigma^1,\sigma^2)=(f_1,f_2,f_3)$ where $(\sigma^1,\sigma^2)$ denote the spatial coordinates in the image domain, and $f_1,f_2,f_3$ denote the RGB color components.
	$\bbf$ can be considered as a two-dimensional surface embedded in the five-dimensional space-feature space: $F= (\sqrt{\alpha}\sigma^1, \sqrt{\alpha}\sigma^2, f_1, f_2, f_3)$, where $\alpha>0$ is a scalar parameter controlling the ratio between the space and color.
	Assuming the spatial and color spaces to be Euclidean, we have  $h_{ij}=\delta_{ij}$ on the space-feature space, where $\delta_{ij}$ is the Kronecker delta, $\delta_{ij}=1$ if $i=j$ and $\delta_{ij}=0$ otherwise.
	Set $\sigma^1=x_1,\sigma^2=x_2$.
	According to (\ref{eq.inducedMetric}), the metric on the image manifold is
	\begin{align}
		\bG\equiv\{g_{ij}\}=\begin{pmatrix}
			\alpha+\sum_{k=1}^3(\partial_1 f_k)^2 & \sum_{k=1}^3 \partial_1 f_k\partial_2f_k\\
			\sum_{k=1}^3 \partial_1 f_k \partial_2 f_k & \alpha+\sum_{k=1}^3(\partial_2 f_k)^2
		\end{pmatrix}.
		\label{eq.colorMetric}
	\end{align}
	
	Denote $(\Sigma,g)$ the image manifold and its metric, and  $(M,h)$ its space-feature space and the metric on it. The weight of the mapping $F:\Sigma\rightarrow M$ is measured by
	\begin{align}
		S(F,g_{ij},h_{\mu\nu})=\int d^m\sigma \sqrt{g}\|dF\|_{g,h}^2
		\label{eq.Polyakov}
	\end{align}
	for $i,j=1,...,\dim(\Sigma)$ and $\mu,\nu=1,...,\dim(M)$, where $ g=\det(\bG)$
	and $\|dX\|_{g,h}^2=\partial_{i} F^{\mu}\partial_j F^{\nu} g^{ij} h_{\mu\nu}$ with $F=(F^1,...,F^{\dim(M)})$ and $\{g^{ij}\}$ being the inverse of the image metric $\{g_{ij}\}$.
	With $m=2,h_{ij}=\delta_{ij}$, the functional is known as the Polyakov action \cite{polyakov1989quantum} in String Theory.
	
	Substituting $h_{ij}=\delta_{ij}$ and (\ref{eq.colorMetric}) into (\ref{eq.Polyakov}) gives rise to
	\begin{align}
		S(F)=\int \sqrt{g}d\bx \mbox{\,\,\,\,\,\,\,\,\, where\,\, } g=\det(\bG).
		\label{eq.colorS}
	\end{align}
	Minimizing (\ref{eq.colorS}) by variational gradient descent, yields the Beltrami flow \cite{kimmel1997high},
	\begin{align}
		F_t^k=\frac{1}{\sqrt{g}} \nabla\cdot (\sqrt{g} \bG^{-1}\nabla F^k),
		\label{eq.bel}
	\end{align}
	where $\nabla F^k$ is a column vector by convention. The term on the right hand side of (\ref{eq.bel}) is known as the Laplace--Beltrami operator applied to $F^k$.
	
	\paragraph{Notations} In the rest of the paper, we will use regular letters to denote scalar-valued functions, and  bold letters to denote vector-valued and matrix-valued functions.
	\section{Formulation of the color elastica model}\label{sec.model}
	We assume the domain $\Omega=[0,L_1]\times[0,L_2]$ to be a rectangle. Periodic boundary condition is a common condition in image processing. In this paper, we assume all variables (scalar-valued, vector-valued and matrix-valued) have the periodic boundary condition. 
	Let $\bbf=(f_1,...,f_d)^T$ be the given $d$-dimensional signal to be smoothed, in which each component $f_k: \Omega\rightarrow \RR^+$ represents a channel.
	Based on the Polyakov action (\ref{eq.colorS}), we define the color elastica model as,
	\begin{equation}
		\min_{\bv\in \cV^d}\int_{\Omega}\left ( 1+\beta\sum_{k=1}^d (\Delta_g v_k)^2\right ) \sqrt{g} d\bx +\frac{1}{2\eta} \sum_{k=1}^d \int_{\Omega} |v_k-f_k|^2d\bx,
		\label{eq.model}
	\end{equation}
	where $\bv=(v_1, ..., v_d)^T: \Omega\rightarrow \RR^d$, $\beta, \eta>0$ are weight parameters, $\cV$ denotes the Sobolev space with periodic boundary condition $\cH^2_P(\Omega)$ on $\Omega$:
	\begin{align*}
		\cH^2_P(\Omega)=&\big\{f\in \cH^2(\Omega): f(0,:)=f(L_1,:), f(:,0)=f(:,L_2), \\
		&\quad \nabla f(0,:)= \nabla f(L_1,:), \nabla f(:,0)= \nabla f(:,L_2)\big\}.
	\end{align*}
	Similar to (\ref{eq.colorMetric}), let $\bG=(g_{kr})_{1\leq k,r \leq 2}$ be a symmetric positive-definite metric matrix depending on $\bv$, where,
	\begin{align*}
		g_{11}=\alpha+\sum_{k=1}^d \left| \frac{\partial v_i}{\partial x_1}\right|^2,\ g_{12}=g_{21}=\sum_{k=1}^d \frac{\partial v_k}{\partial x_1} \frac{\partial v_k}{\partial x_2},\ g_{22}=\alpha+\sum_{k=1}^d \left| \frac{\partial v_k}{\partial x_2}\right|^2,
	\end{align*}
	for some $\alpha>0$.
	We denote $g=\det \bG$. $\Delta_g$ in (\ref{eq.model}) is the Laplace--Beltrami operator associated to $\bG$,
	\begin{equation}
		\Delta_g \phi\,=\,\frac{1}{\sqrt{g}} \nabla \cdot (\sqrt{g}\bG^{-1}\nabla\phi),\,\,\,\, \forall \phi\in V.
	\end{equation}
	Here, we consider the case $d=3$ for color images: $\bbf=(f_1,f_2,f_3)^T$ whose components represent the RGB channels. Then $\bG$ is the metric of the surface $F(x_1,x_2)=(\sqrt{\alpha}x_1,\sqrt{\alpha}x_2,v_1,v_2,v_3$ induced by the denoised image $\bv$.
	
	%
	
	%
	\begin{remark}
		As described in Section \ref{sec.Polyakov}, $\alpha$ represents the ratio between the spatial and feature coordinates.
		As $\alpha$ goes to $0$, the ratio vanishes and the space-feature space reduces to the three-dimensional color space.
	\end{remark}
	\begin{remark}
		When $f$ is a gray-scale image, dividing the first integral by $\sqrt{\alpha}$, model (\ref{eq.model}) reduces to
		\begin{align}
			&\tilde{E}_{\alpha}(v)\equiv\int_{\Omega} \sqrt{\alpha+(v_x^2+v_y^2)}d\bx +\beta \int_{\Omega} \frac{1}{\sqrt{\alpha+(v_x^2+v_y^2)}}\left( \nabla\cdot \frac{\nabla v}{\sqrt{\alpha+v_x^2+v_y^2}}\right)^2d\bx\nonumber\\
			&\hspace{1cm} +\frac{1}{2\eta}\int_{\Omega} |v-f|^2d\bx.
			\label{eq.gray}
		\end{align}
		As $\alpha \rightarrow 0$, (\ref{eq.gray}) becomes
		\begin{align}
			&\tilde{E}_{0}(v)=\int_{\Omega} |\nabla v|d\bx +\beta \int_{\Omega} \frac{1}{|\nabla v|}\left( \nabla\cdot \frac{\nabla v}{|\nabla v|}\right)^2d\bx +\frac{1}{2\eta}\int_{\Omega} |v-f|^2d\bx
			\label{eq.gray1}
		\end{align}
		which gives a variant of the elastica model (\ref{eq.greyelastica}), except that in (\ref{eq.gray1}) the weights depend on $|\nabla v|$.
		In other words, along the edges, the effect of the elastica term would be reduced. 
		%
	\end{remark}

		\begin{remark}
			In the one channel case, the Polyakov action relates closely to existing models.
			As $\alpha \rightarrow 0$, the Polyakov action converges to the TV model, and the corresponding gradient flow converges to a level set mean curvature flow. As $\alpha \rightarrow \infty$, the gradient flow of the Polyakov action converges to a diffusion process, i.e., Gaussian smoothing.
			In the multi-channel case, the Polyakov action contains interactions between channels. By letting $\alpha \rightarrow 0$, the Polyakov action does not reduce to the color TV model suggested in \cite{blomgren1998color}. 
			The color area minimization also tries to remove potential torsion of the image manifold and align the gradient directions of different channels. 
			Correspondingly, its gradient flow converges to the mean curvature flow in the color space. 
			Thus, the Polyakov action can be thought of as a natural extension of the TV model in the multi-channel case. 
			By adding the Laplace-Betrami operator (LBO) acting on the color channels to the Polyakov action, the proposed color elastica model becomes a natural extension of the Euler's elastica model, since it reduces to the Euler's elastica model in one-channel case as $\alpha \rightarrow 0$.
			Moreover, when $\alpha \rightarrow 0$ the measure we end up with captures the area of the image manifold in the $RGB$ space for which the minimization is a mean curvature flow.
			The new term we add is the LBO acting on the $RGB$ coordinates which is, in fact, the mean curvature vector of the $RGB$ surface, here denoted by $\Delta_g {\bf v}$.
			Thus, we end up minimizing the squared value of the mean curvature $\int_\Omega (1+\beta|\Delta_g {\bf v}|^2)da$ for $da = \sqrt{g}d{\bf x}$.
		\end{remark}

	The energy in (\ref{eq.model}) is nonlinear and is difficult to minimize, since it contains the determinate and inverse of a Hessian matrix.
	To simplify the nonlinearity, we introduce three vector-valued and matrix-valued variables.
	For $k=1,2,3$ and $r=1,2$, let us denote by $q_{kr}$ the real valued function $\frac{\partial v_k}{\partial x_r}$ and by $\bq$ the $3\times 2$ Jacobian matrix
	\begin{align*}
		\bq=\begin{pmatrix}
			q_{11} & q_{12}\\
			q_{21} & q_{22}\\
			q_{31} & q_{32}
		\end{pmatrix} =\nabla \bv.
	\end{align*}
	Denote $\bq_k=\begin{pmatrix}
		q_{k1} & q_{k2}
	\end{pmatrix}, k=1,2,3$. We introduce the $3\times 2$ matrix $\bmu=\sqrt{g}\bq\bG^{-1}$ with
	\begin{equation}
		\bmu_k=\sqrt{g}\bq_k\bG^{-1}
		\label{eq.mu}
	\end{equation}
	for each row of $\bmu$. We immediately have $\bq_k=\frac{1}{\sqrt{g}}\bmu_k\bG$.
	We shall use  $\bM(\bq)$ to be the matrix-valued function defined by
	\begin{align}
		\bM(\bq) =   \begin{pmatrix}
			\alpha+q_{11}^2+q_{21}^2+q_{31}^2 & q_{11}q_{12}+q_{21}q_{22}+q_{31}q_{32}\\
			q_{11}q_{12}+q_{21}q_{22}+q_{31}q_{32} & \alpha+q_{12}^2+q_{22}^2+q_{32}^2
		\end{pmatrix},
	\end{align}
	and denote $\det \bM(\bq)$ by $m(\bq)$, that is,
	\[m(\bq) = \det \bM(\bq). \]
	Define $\bv_{\bq}=\{(v_{\bq})_k\}_{k=1}^3$ to be the solution of
	\begin{equation}
		\begin{cases}
			\nabla^2(v_{\bq})_k\,=\,\nabla \cdot \bq_k &\mbox{ in } \Omega,\\
			(v_{\bq})_k \mbox{ has the periodic boundary condition} &\mbox{ on } \partial\Omega,\\
			\int_{\Omega} (v_{\bq})_k dx\,=\,\int_{\Omega} f_k dx\\
			\mbox{for } k=1,2,3.
		\end{cases}
		\label{eq.v}
	\end{equation}
	
	%
	Denote 
	\begin{align*}
		&\cL_P^2=\{f\in \cL^2(\Omega): f(0,:)=f(L_1,:), f(:,0)=f(:,L_2),\\
		&\cH_P^1=\{f\in \cH^1(\Omega): f(0,:)=f(L_1,:), f(:,0)=f(:,L_2).
	\end{align*}
	We define the sets $\Sigma_f$ and $S_{\bp}$ as
	\begin{align}
		&\Sigma_f=\Big\{\bq\in (\cL^2_P(\Omega))^{3\times 2}: \exists \bv\in (\cH^1_P(\Omega))^3 \mbox{ such that } \bq=\nabla \bv \nonumber\\
		&\hspace{5cm} \mbox{ and } \int_{\Omega} (v_k-f_k) dx=0 \mbox{ for }k=1,2,3\Big\},\\
		&S_{\bG}=\left\{(\bq,\bmu)\in \left((\cL^2_P(\Omega))^{3\times 2},(\cL^2_P(\Omega))^{3\times 2}\right): \bmu_k=\sqrt{g}\bq_k\bG^{-1}, k=1,2,3 \mbox{ with }g=\det \bG\right\}
	\end{align}
	and their corresponding indicator functions as
	\begin{align*}
		I_{\Sigma_f}(\bq)=\begin{cases}
			0, &\mbox{ if }\bq\in\Sigma_f,\\
			\infty, &\mbox{ otherwise},
		\end{cases}\hspace{1cm}
		I_{S_{\bG}}(\bq,\bmu)=\begin{cases}
			0, &\mbox{ if }(\bq,\bmu)\in S_{\bG},\\
			\infty, &\mbox{ otherwise}.
		\end{cases}
	\end{align*}
	If $(\bp,\blambda,\bG)$ is the minimizer of the following energy
	\begin{equation}
		\begin{cases}
			\min\limits_{\bmu,\bq}\displaystyle\int_{\Omega} \left(\sqrt{g}+\frac{\beta}{\sqrt{g}} \sum_{k=1}^3 |\nabla\cdot \bmu_k|^2 \right) d\bx +\frac{1}{2\eta} \sum_{k=1}^3 \displaystyle\int_{\Omega} |(v_{\bq})_k-f_k|^2 d\bx \\
			\hspace{2cm}+I_{\Sigma_f}(\bq)+I_{S_{\bG}}(\bq,\bmu),\\
			\bG=\bM(\bq),
		\end{cases}
		\label{eq.model1}
	\end{equation}
	then $\bv$ solving (\ref{eq.v}) minimizes (\ref{eq.model}).
	
	In (\ref{eq.model1}), we rewrite the Laplace--Beltrami term as $\nabla\cdot \bmu_k$ where $\bmu$ linearly depends on $\bq$. The metric $\sqrt{g}=\det \bM(\bq)$ is also a function of $\bq$. In the fidelity term, $\bv$ which solves (\ref{eq.v}) can be uniquely determined by $\bq$. Thus the functional only depends on $\bmu$ and $\bq$ under some constraints on their relations. We then remove the constraints by incorporating the functional with the two indicator function $I_{\Sigma_f}(\bq)$ and $I_{S_{\bG}}(\bq,\bmu)$. The resulting problem (\ref{eq.model1}) is an unconstrained optimization problem with respect to $\bmu$ and $\bq$ only.

	\begin{remark}
		While the periodic boundary condition is used in this paper, we would like to mention that zero Neumann boundary condition is another natural condition to consider. The model and algorithm proposed in this paper can be adapted to zero Neumann boundary condition with minor modification.
	\end{remark}
	

	\section{Operator splitting method}\label{sec.splitting}
	In this section, we derive the operator-splitting scheme to solve (\ref{eq.model1}). We refer the readers to \cite{glowinski2017splitting} for a complete introduction of the operator-splitting method and to \cite{glowinski2019fast} for its applications in image processing.
	\subsection{Optimality condition}
	
	Denote
	\begin{align}
		&J_1(\bq,\bmu,\bG)=\int_{\Omega} \left(\sqrt{g}+\frac{\beta}{\sqrt{g}} \sum_{k=1}^3 |\nabla\cdot \bmu_k|^2 \right) d\bx,\\
		&J_2(\bq)=\frac{1}{2\eta} \sum_{k=1}^3 \int_{\Omega} |(v_{\bq})_k-f_k|^2 d\bx
	\end{align}
	with $g=\det \bG$.
	If $(\bp,\blambda,\bG)$ is the minimizer of (\ref{eq.model1}), it satisfies
	\begin{equation}
		\begin{cases}
			\partial_{\bq}J_1(\bp,\blambda) +\partial_{\bq}J_2(\bp)+\partial I_{\Sigma_f}(\bp)+\partial_{\bq} I_{S_{\bp}}(\bp,\blambda) \ni 0,\\
			\partial_{\bmu} J_1(\bp,\blambda) +\partial_{\bmu} I_{S_{\bG}}(\bp,\blambda) \ni0,\\
			\bG-\bM(\bp)=0,
		\end{cases}
		\label{eq.optimal}
	\end{equation}
	where $\partial_{\bq}$ denotes the partial derivative with respect to $\bq$ if its operand is a smooth function, and the sub-derivative if its operand is an indicator function.
	To solve (\ref{eq.optimal}), we introduce the artificial time and associate it with the initial value problem
	\begin{equation}
		\begin{cases}
			\frac{\partial \bp}{\partial t}+\partial_{\bq}J_1(\bp,\blambda,\bG) +\partial_{\bq}J_2(\bp,\bG)+\partial I_{\Sigma_f}(\bp)+\partial_{\bq} I_{S_{\bG}}(\bp,\blambda) \ni 0,\\
			\gamma_1\frac{\partial \blambda}{\partial t}+\partial_{\bmu} J_1(\bp,\blambda) +\partial_{\bmu} I_{S_{\bG}}(\bp,\blambda) \ni0,\\
			\frac{\partial \bG}{\partial t}+\gamma_2(\bG-\bM(\bp))=0,\\
			(\bp(0),\blambda(0),\bG(0))=(\bp^0,\blambda^0,\bG^0),
		\end{cases}
		\label{eq.ivp}
	\end{equation}
	where $(\bp^0,\blambda^0,\bG^0)$ is the initial condition which is supposed to be given, and $\gamma_1, \gamma_2$ are positive constants controlling the evolution speed of $\blambda$ and $\bG$, respectively. The choice of $\gamma_1$ will be discussed in Section \ref{sec.gammachoice}.
	\subsection{Operator-splitting scheme}
	Similar to \cite{glowinski2019finite,liu2019finite,glowinski2020numerical}, we use an operator-splitting method to time discretize (\ref{eq.ivp}). One simple choice is the Lie scheme \cite{glowinski2017}. We denote $n$ as the iteration number, $\det\bG^n$ by $g^n$, the time step by $\tau$ and set $t^n = n \tau$. Since $J_1(\bq,\bmu,\bG)$ only contains $g$, when there is no ambiguity, we write $J_1(\bq,\bmu,\bG)$ and $J_1(\bq,\bmu,g)$ interchangeably. We update $\bp,\blambda,\bG$ as follows,\\
	\emph{\underline{Initialization}}
	\begin{equation}
		\mbox{Initialize }\bp^0, \bG^0, \blambda^0 \mbox{ and compute } g^0=\det\bG^0.
		\label{eq.split.0}
	\end{equation}
	\emph{\underline{Fractional step 1}}\\
	Solve
	\begin{equation}
		\begin{cases}
			\begin{cases}
				\frac{\partial \bp}{\partial t}+\partial_{\bq} J_1(\bp,\blambda,\bG) \ni \zero,\\
				\gamma_1 \frac{\partial \blambda}{\partial t}+ \partial_{\bmu}J_1(\bp,\blambda,\bG)= \zero,\\
				\frac{\partial \bG}{\partial t}+\frac{\gamma_2 }{3}(\bG-\bM(\bp))=\zero,
			\end{cases}
			\mbox{ in } \Omega\times (t^n,t^{n+1}),\\
			(\bp(t^n),\blambda(t^n),\bG(t^n))=(\bp^n,\blambda^n,\bG^n)
		\end{cases}
		\label{eq.split.1}
	\end{equation}
	and set $\bp^{n+1/3}=\bp(t^{n+1}),\blambda^{n+1/3}=\blambda(t^{n+1}),\bG^{n+1/3}=\bG(t^{n+1}), g^{n+1/3}=\det \bG^{n+1/3}$.
	
	\noindent\emph{\underline{Fractional step 2}}\\
	Solve
	\begin{equation}
		\begin{cases}
			\begin{cases}
				\frac{\partial \bp}{\partial t}+\partial_{\bq} I_{S_{\bG^{n+1/3}}}(\bp,\blambda) \ni \zero,\\
				\gamma_1 \frac{\partial \blambda}{\partial t}+ \partial_{\bmu}I_{S_{\bG^{n+1/3}}}(\bp,\blambda)\ni \zero,\\
				\frac{\partial \bG}{\partial t}+\frac{\gamma_2 }{3}(\bG-\bM(\bp))=\zero
			\end{cases}
			\mbox{ in } \Omega\times (t^n,t^{n+1}),\\
			(\bp(t^n),\blambda(t^n),\bG(t^n))=(\bp^{n+1/3},\blambda^{n+1/3},\bG^{n+1/3})
		\end{cases}
		\label{eq.split.2}
	\end{equation}
	and set $\bp^{n+2/3}=\bp(t^{n+1}),\blambda^{n+2/3}=\blambda(t^{n+1}),\bG^{n+2/3}=\bG(t^{n+1}), g^{n+2/3}=\det \bG^{n+2/3}$.
	
	\noindent\emph{\underline{Fractional step 3}}\\
	Solve
	\begin{equation}
		\begin{cases}
			\begin{cases}
				\frac{\partial \bp}{\partial t}+\partial_{\bq} J_2(\bq)+ \partial_{\bq} I_{\Sigma_f}(\bp) \ni \zero,\\
				\gamma_1 \frac{\partial \blambda}{\partial t}= \zero,\\
				\frac{\partial \bG}{\partial t}+\frac{\gamma_2 }{3}(\bG-\bM(\bp))=\zero
			\end{cases}
			\mbox{ in } \Omega\times (t^n,t^{n+1}),\\
			(\bp(t^n),\blambda(t^n),\bG(t^n))=(\bp^{n+2/3},\blambda^{n+2/3},\bG^{n+2/3})
		\end{cases}
		\label{eq.split.3}
	\end{equation}
	and set $\bp^{n+1}=\bp(t^{n+1}),\blambda^{n+1}=\blambda(t^{n+1}),\bG^{n+1}=\bG(t^{n+1}), g^{n+1}=\det \bG^{n+1}$.
	
	Scheme (\ref{eq.split.1})-(\ref{eq.split.3}) is only semi-discrete since we still need to solve the three sub-initial value problems. There is no difficulty in updating $\bG$ in (\ref{eq.split.1})-(\ref{eq.split.3}) if $\bM(\bp)$ is fixed, since we have the exact solution $\bG(t^{n+1})=e^{-\gamma_2\tau/3}G(t^n)+(1-e^{-\gamma_2\tau/3})M(\bp)$. To solve other subproblems, we suggest to use the Marchuk-Yanenko type discretization:\\
	Initialize $\bp^0, \bG^0, \blambda^0 \mbox{ and compute } g^0=\det\bG^0$.
	
	For $n\geq0$, we update $(\bp^n,\blambda^n,\bG^n,g^n)\rightarrow (\bp^{n+1/3},\blambda^{n+1/3},\bG^{n+1/3},g^{n+1/3})$ \\
	$ \rightarrow (\bp^{n+2/3},\blambda^{n+2/3},\bG^{n+2/3},g^{n+2/3}) \rightarrow (\bp^{n+1},\blambda^{n+1},\bG^{n+1},g^{n+1})$ as:
	\begin{align}
		&\begin{cases}
			\frac{\bp^{n+1/3}-\bp^n}{\tau} + \partial_{\bq} J_1(\bp^{n+1/3},\blambda^n,m(\bp^{n+1/3})) \ni \zero,\\
			\bG^{n+1/3}=e^{-\gamma_2\tau/3}\bG^n+(1-e^{-\gamma_2\tau/3})\bM(\bp^{n+1/3}),\\
			g^{n+1/3}=\det \bG^{n+1/3},\\
			\frac{\blambda^{n+1/3}-\blambda^n}{\tau} + \partial_{\bmu} J_1(\bp^{n+1/3},\blambda^{n+1/3},g^{n+1/3}) = \zero.
		\end{cases} \label{eq.split1.1}\\
		&\begin{cases}
			\frac{\bp^{n+2/3}-\bp^{n+1/3}}{\tau} + \partial_{\bq} I_{S_{\bG^{n+1/3}}}(\bp^{n+2/3},\blambda^{n+2/3}) \ni \zero,\\
			\gamma_1\frac{\blambda^{n+2/3}-\blambda^{n+1/3}}{\tau} + \partial_{\bmu} I_{S_{\bG^{n+1/3}}}(\bp^{n+2/3},\blambda^{n+2/3}) \ni \zero,\\
			\bG^{n+2/3}=e^{-\gamma_2\tau/3}\bG^{n+1/3}+(1-e^{-\gamma_2\tau/3})\bM(\bp^{n+2/3}),\\
			g^{n+2/3}=\det \bG^{n+2/3},
		\end{cases}\label{eq.split1.2}\\
		&\begin{cases}
			\frac{\bp^{n+1}-\bp^{n+2/3}}{\tau} +\partial_{\bq} J_2(\bp^{n+1})+ \partial_{\bq} I_{\Sigma_f}(\bp^{n+1}) \ni \zero,\\
			\bG^{n+1}=e^{-\gamma_2\tau/3}\bG^{n+2/3}+(1-e^{-\gamma_2\tau/3})\bM(\bp^{n+1}),\\
			g^{n+1}=\det \bG^{n+1},\\
			\blambda^{n+1}=\blambda^{n+2/3}.
		\end{cases}\label{eq.split1.3}
	\end{align}
	In the following subsections, we discuss the solution of each of the sub-PDE systems (\ref{eq.split1.1})-(\ref{eq.split1.3}).
	\begin{remark}
		Scheme (\ref{eq.split.0})-(\ref{eq.split.3}) is an approximation of the gradient flow. Thus the convergence rate of the proposed scheme closely relates to that of the gradient flow together with an approximation error. It is shown that when there is only one variable and the operators in each step is smooth enough, the approximation error is of $O(\tau)$ (see \cite{chorin1978product} and \cite[Chapter 6]{glowinski2003finite}). In this article, due to the complex structures of $J_1,J_2, I_{\Sigma_f}$ and $I_{S_{\bG}}$, we cannot use existing tools to analyze the convergence rate and need to study it separately.
	\end{remark}

	\subsection{On the solution of (\ref{eq.split1.1})}
	If $\bp^{n+1/3}$ is the minimizer of the following problem, then  the Euler-Lagrangian equation for it is exactly (\ref{eq.split1.1}) and this means $\bp^{n+1/3}$ solves (\ref{eq.split1.1}):
	\begin{align}
		\bp^{n+1/3}=&\argmin_{\bq\in (\cL^2_P(\Omega))^{3\times 2}} \Bigg[ \frac{1}{2\tau} \int_{\Omega} |\bq-\bp^n|^2dx \nonumber\\
		&\quad +\int_{\Omega} \left( \sqrt{m(\bq)} +\frac{\beta}{\sqrt{m(\bq)}} \sum_{k=1}^3 |\nabla\cdot \blambda^n_k|^2\right) d\bx \Bigg].
		\label{eq.frac1.p}
	\end{align}
	This problem can be solved by Newton's method. The functional in (\ref{eq.frac1.p}) is in the form of
	\begin{equation}
		E_1=\frac{1}{2\tau} \int_{\Omega} |\bq-\bp|^2dx +\int_{\Omega} \left( s_1\sqrt{m(\bq)} +\frac{\beta s_2}{\sqrt{m(\bq)}} \right) d\bx
	\end{equation}
	for some $s_1,s_2\geq 0, \bp\in (\cL^2_P(\Omega))^{3\times 2}$.
	The first and second order variation of $E_1$ with respect to $q_{kr},k=1,2, r=1,2,3$ are
	\begin{align}
		&\frac{\partial E_1}{\partial q_{kr}}(\bq,\bp)=\frac{1}{\tau}(q_{kr}-p_{kr}) +\frac{1}{2}\left(s_1(m(\bq))^{-\frac{1}{2}}-\beta s_2(m(\bq))^{-\frac{3}{2}}\right) \frac{\partial m(\bq)}{\partial q_{kr}},\\
		&
		\begin{aligned}
			\frac{\partial^2 E_1}{\partial q_{kr}^2}(\bq,\bp)=&\frac{1}{\tau}+\frac{1}{2}\left(s_1(m(\bq))^{-\frac{1}{2}}-\beta s_2 (m(\bq))^{-\frac{3}{2}}\right) \frac{\partial^2 m(\bq)}{\partial q_{kr}^2}\\
			&\quad\quad  +\frac{1}{2}\left(-\frac{1}{2}s_1(m(\bq))^{-\frac{3}{2}}
			+ \frac{3}{2}\beta s_2 (m(\bq))^{-\frac{5}{2}}\right)\left(\frac{\partial m(\bq)}{\partial q_{kr}}\right)^2
		\end{aligned}
	\end{align}
	with
	\begin{align}
		&\frac{\partial m(\bq)}{\partial q_{k1}}=2g_{22}q_{k1}-2g_{12}q_{k2}, & &\frac{\partial m(\bq)}{\partial q_{k2}}=2g_{11}q_{k2}-2g_{12}q_{k1},\\
		&\frac{\partial^2 m(\bq)}{\partial q_{k1}^2}=2g_{22}-2q_{k2}^2,& &\frac{\partial^2 m(\bq)}{\partial q_{k2}^2}=2g_{11}-2q_{k1}^2
	\end{align}
	for $k=1,2,3$. From an initial guess of $\bq^0$, $q_{kr}$ is updated by
	\begin{equation}
		q_{kr}^{\omega+1}=q_{kr}^\omega-\frac{\frac{\partial E_1}{\partial q_{kr}}(\bq^\omega,\bp)}{\frac{\partial^2 E_1}{\partial q_{kr}^2} (\bq^\omega,\bp)}
		\label{eq.update.q}
	\end{equation}
	until $\max_{k,r}|q_{kr}^{\omega+1}-q_{kr}^{\omega}|_{\infty}<tol$ for some stopping criterion $tol$. Then we set $q^{n+1/3}_{kr}=q^*_{kr}$ where $q^*_{kr}$ is the converged variable.

		\begin{remark}
			In our algorithm, the initial guess of $\mathbf{q}^0$ is set to be $\mathbf{p}^n$, the minimizer of (\ref{eq.frac1.p}) in the previous iteration. 
			As long as the time step is small enough, $\mathbf{p}^{n}$ and $\bm{\lambda}^{n}$ are close to $\mathbf{p}^{n-1}$ and $\bm{\lambda}^{n-1}$, respectively. 
			Thus, the functional in (\ref{eq.frac1.p}) does not change too much from that in the previous iteration. It is reasonable to expect its minimizer is close to $\mathbf{p}^n$, and $\mathbf{p}^n$ is a good initial guess. This is verified in our numerical experiments.
		\end{remark}

	For $\blambda^{n+1/3}$, it is the solution to
	\begin{equation}
		\begin{cases}
			\blambda^{n+1/3}=\{\blambda^{n+1/3}_k\}_{k=1}^3 \in (\cH^1_P(\Omega))^{3\times 2},\\
			\gamma_1\displaystyle\int_{\Omega} \blambda_k^{n+1/3}\cdot \bmu_k dx+ 2\beta\tau \displaystyle\int_{\Omega} \frac{(\nabla\cdot \blambda_k^{n+1/3})(\nabla \cdot \bmu_k)}{\sqrt{g^{n+1/3}}}dx=\gamma_1 \int_{\Omega} \blambda_k^n \cdot \bmu_kdx,\\
			\forall \bmu_k\in (\cH^1_P(\Omega))^2, k=1,2,3,
		\end{cases}
	\end{equation}
	which is equivalent to
	\begin{equation}
		\begin{cases}
			\gamma_1\blambda_k^{n+1/3}-2\beta\tau\nabla\left(\frac{\nabla\cdot \blambda_k^{n+1/3}}{\sqrt{g^{n+1/3}}}\right)=\gamma_1\blambda_k^n &\mbox{ in } \Omega,\\
			\blambda_k^{n+1/3} \mbox{ has the periodic boundary condition} &\mbox{ on } \partial\Omega,\\
			k=1,2,3.
		\end{cases}
		\label{eq.frac1.pde.lambda.fft}
	\end{equation}
	Problem (\ref{eq.frac1.pde.lambda.fft}) (and (\ref{eq.frac3.pde.u.fft}) in Section \ref{sec.solvev}) is a simple linear PDE. In the case that $\Omega$ is a rectangle, there are many fast solvers (like sparse Cholesky, conjugate gradient and cyclic reduction to name a few) for this kind of problems. 
	
	%
	
	
	\subsection{On the solution of (\ref{eq.split1.2}) and the choice of $\gamma_1$}
	\label{sec.gammachoice}
	The solution $(\bp^{n+2/3},\blambda^{n+2/3})$ is the minimizer of
	\begin{equation}
		(\bp^{n+2/3},\blambda^{n+2/3})=\argmin_{(\bq,\bmu)\in S_{\bG^{n+1/3}}} \int_{\Omega} \left( \left| \bq-\bp^{n+1/3}\right|^2 +\gamma_1 \left| \bmu-\blambda^{n+1/3}\right|^2 \right)dx .
		\label{eq.frac2.lambda}
	\end{equation}
	Recall that
	$$S_{\bG}=\left\{(\bq,\bmu)\in \left((\cL^2_P(\Omega))^{3\times 2},(\cL^2_P(\Omega))^{3\times 2}\right), \bmu_k=\sqrt{g}\bq_k\bG^{-1}, k=1,2,3 \mbox{ with }g=\det \bG\right\}.$$
	Thus, we have $\bq=\frac{1}{\sqrt{g^{n+1/3}}}\bmu\bG^{n+1/3}$. Substituting this into (\ref{eq.frac2.lambda}), the right hand side becomes a functional of $\bmu$ only:
	\begin{equation}
		E_2=\int_{\Omega} \left( \left| \frac{1}{\sqrt{g^{n+1/3}}}\bmu\bG^{n+1/3}-\bp^{n+1/3}\right|^2 +\gamma_1 \left| \bmu-\blambda^{n+1/3}\right|^2 \right)dx.
		\label{eq.frac2.lambda.1}
	\end{equation}
	For simplicity, we temporally use $g, g_{kr},p_{kr},\lambda_{kr}$ to denote $g^{n+1/3}, g_{kr}^{n+1/3},p_{kr}^{n+1/3},\lambda_{kr}^{n+1/3}$ in this subsection. Computing the variation of $E_2$ with respect to $\bmu_k$ gives
	\begin{equation}
		\frac{\partial E_2}{\partial \bmu_k}= \begin{pmatrix}
			A_{11} & A_{12}\\ A_{21} & A_{22}
		\end{pmatrix}
		\begin{pmatrix}
			\mu_{k1} \\ \mu_{k2}
		\end{pmatrix}-
		\begin{pmatrix}
			b_1\\ b_2
		\end{pmatrix}
	\end{equation}
	with
	\begin{align*}
		&A_{11}=\frac{2g_{11}^2+2g_{12}^2}{g}+2\gamma_1,\  A_{12}=A_{21}=\frac{2g_{11}g_{12}+2g_{12}g_{22}}{g}, \ A_{22}=\frac{2g_{12}^2+2g_{22}^2}{g}+2\gamma_1\\
		&b_1=\frac{2g_{11}p_{k1}+2g_{21}p_{k2}}{\sqrt{g}}+2\gamma_1\lambda_{k1},\ b_2=\frac{2g_{12}p_{k1}+2g_{22}p_{k2}}{\sqrt{g}}+2\gamma_1\lambda_{k2}.
	\end{align*}
	By the optimality condition, we get that
	\begin{equation}
		\begin{pmatrix}
			\lambda_{k1}^{n+2/3}\\ \lambda_{k2}^{n+2/3}
		\end{pmatrix}
		=\frac{1}{A_{11}A_{22}-A_{12}A_{21}}
		\begin{pmatrix}
			A_{22}b_1-A_{12}b_2\\ -A_{21}b_1+A_{11}b_2
		\end{pmatrix}
		\label{eq.frac2.lambda.update}
	\end{equation}
	for $k=1,2,3$. And $\bp_k^{n+1/3}$ is computed as
	\begin{equation}
		\bp_k^{n+2/3}=\frac{1}{\sqrt{g^{n+1/3}}}\blambda_k^{n+2/3} \bG^{n+1/3}.
		\label{eq.frac2.p.update}
	\end{equation}
	
	For the choice of $\gamma_1$, we want to chose $\gamma_1$ such that the two terms in (\ref{eq.frac2.lambda}) are balanced. Since $\bmu=\sqrt{g}\bp\bG^{-1}$, we have
	\begin{align}
		\frac{\partial \bmu}{\partial t}=\sqrt{g}\frac{\partial \bp}{\partial t}\bG^{-1}.
	\end{align}
	The integrand in (\ref{eq.frac2.lambda}) can be approximated as
	\begin{align}
		\left| \bp^{n+2/3}-\bp^{n+1/3}\right|^2 +\gamma_1 \left| \blambda^{n+2/3}-\blambda^{n+1/3}\right|^2\approx \tau^2 \left(\left| \frac{\partial \bp}{\partial t}(t^{n+1/3})\right|^2+\gamma_1 \left| \frac{\partial \bmu}{\partial t}(t^{n+1/3}) \right|^2\right).
	\end{align}
	Denote the eigenvalues of $\bG$ by $\rho_1,\rho_2$. To balance the two terms, the above estimations suggest using $\gamma_1=\sqrt{g}/\min\{\rho_1,\rho_2\}$. Since $g=\det \bG=\rho_1\rho_2$, we approximate $\min\{\rho_1,\rho_2\}\approx \sqrt{g}$ which gives rise to $\gamma_1=1$.
	\subsection{On the solution of (\ref{eq.split1.3})}
	\label{sec.solvev}
	In (\ref{eq.split1.3}), since $v_{\bq,g}$ is the solution to (\ref{eq.v}), we write $\bp^{n+1}$ as $\nabla \bu^{n+1}$ for $\bu^{n+1}=(u_1^{n+1},u_2^{n+1},u_3^{n+1})^T$. If $\bp^{n+1}$ minimizes (\ref{eq.split1.3}), $\bu^{n+1}$ is the solution to
	\begin{equation}
		\begin{cases}
			u^{n+1}\in (\cH^1_P(\Omega))^3,\\
			\sum_{k=1}^3\left(\eta \displaystyle\int_{\Omega} \nabla u_k^{n+1} \cdot \nabla v_k dx+\tau \displaystyle\int_{\Omega} u^{n+1}_kv_kdx\right)=\\
			\hspace{1cm}\sum_{k=1}^3\left(\eta \displaystyle\int_{\Omega} \bp_k^{n+2/3} \cdot \nabla v_k dx +\tau\displaystyle\int_{\Omega} f_kv_kdx\right),\\
			\forall v\in (\cH^1_P(\Omega))^3,\\
			\displaystyle\int_{\Omega} u_k^{n+1}d\bx=\int_{\Omega} f_kd\bx,\ k=1,2,3.
		\end{cases}
	\end{equation}
	Equivalently, $\bu^{n+1}$ is the weak solution of
	\begin{equation}
		\begin{cases}
			-\eta \nabla^2u_k^{n+1}+\tau u_k^{n+1}=-\eta \nabla\cdot \bp_k^{n+2/3}+\tau f_k &\mbox{ in } \Omega,\\
			u_k^{n+1} \mbox{ has the periodic boundary condition} &\mbox{ on } \partial\Omega, \\
			k=1,2,3.
		\end{cases}
		\label{eq.frac3.pde.u.fft}
	\end{equation}
	Problem (\ref{eq.frac3.pde.u.fft}) can be solved by FFT. Once it has been solved,  $\bp^{n+1}$ is updated as $\bp^{n+1}=\nabla \bu^{n+1}$.
	
	Our algorithm is summarized in Algorithm \ref{alg.1}.
	\begin{algorithm}
		\caption{\label{alg.1}An algorithm to solve (\ref{eq.model1})}
		\begin{algorithmic}
			\STATE {\bf Input:} The input image $\bbf$, parameters $\beta,\eta,\tau,\gamma_1,\gamma_2$.
			\STATE {\bf Output:} Denoised image $\bu^*$.
			\STATE {\bf Initialization:} n=0, $\bu^0, \bp^0,\blambda^0 , \bG^0, g^0=m(\bG^0)$.
			\WHILE{$\|\bu^{n}-\bu^{n-1}\|_{\infty}>tol$}
			\STATE 1. Solve (\ref{eq.split1.1}) using (\ref{eq.update.q}) and (\ref{eq.frac1.pde.lambda.fft}) to obtain $(\bp^{n+1/3}, \blambda^{n+1/3}, \bG^{n+1/3}, g^{n+1/3})$.
			\STATE 2. Solve (\ref{eq.split1.2}) using (\ref{eq.frac2.lambda.update}) and (\ref{eq.frac2.p.update}) to obtain $(\bp^{n+2/3}, \blambda^{n+2/3}, \bG^{n+2/3}, g^{n+2/3})$.
			\STATE 3. Solve (\ref{eq.split1.3}) using (\ref{eq.frac3.pde.u.fft}) to obtain $(\bu^{n+1},\bp^{n+1}, \blambda^{n+1}, \bG^{n+1}, g^{n+1})$.
			\STATE Set $n=n+1$.
			\ENDWHILE
			\STATE Set $\bu^*=\bu^{n}$.
		\end{algorithmic}
	\end{algorithm}
	\begin{remark}
		We remark that the proposed algorithm can be easily extended to solve image deblurring problems. Consider the following deblurring model (a variant of (\ref{eq.model}))
		\begin{equation}
			\min_{\bv\in \cV^d}\int_{\Omega}\left[ 1+\beta\sum_{k=1}^d |\Delta_g v_k|^2\right] \sqrt{g} d\bx +\frac{1}{2\eta} \sum_{k=1}^d \int_{\Omega} |Kv_k-f_k|^2d\bx,
			\label{eq.model.deblurring}
		\end{equation}
		where $Kv_i=k(x_1,x_2)\star v_i$ with $k(x_1,x_2)$ being the blurring kernel and $\star$ denoting convolution.
		Define the set $\tilde{\Sigma}_f$ as
		\begin{align*}
			\tilde{\Sigma}_f=&\Big\{\bq\in (\cL^2_P(\Omega))^{3\times 2}: \exists \bv\in (\cH^1_P(\Omega))^3 \mbox{ such that } \bq=\nabla v \\
			&\hspace{1cm} \mbox{ and } \int_{\Omega} (K^*Kv_k-K^*f_k) dx=0 \mbox{ for }k=1,2,3\Big\}.
		\end{align*}
		If $(\bp,\blambda,\bG)$ is the minimizer of the following energy
		\begin{equation}
			\begin{cases}
				\min\limits_{\bmu,\bq} \displaystyle\int_{\Omega} \left(\sqrt{g}+\frac{\beta}{\sqrt{g}} \sum_{k=1}^3 |\nabla\cdot \bmu_k|^2 \right) d\bx +\frac{1}{2\eta} \sum_{k=1}^3 \displaystyle\int_{\Omega} |K(v_{\bq})_k-f_k|^2 d\bx \\
				\hspace{2cm}+I_{\tilde{\Sigma}_f}(\bq)+I_{S_{\bG}}(\bq,\bmu),\\
				\bG=\bM(\bq),
			\end{cases}
			\label{eq.model1.deblurring}
		\end{equation}
		then $\bv$ solving 
		\begin{equation}
			\begin{cases}
				\nabla^2(v_{\bq})_k\,=\,\nabla \cdot \bq_k &\mbox{ in } \Omega,\\
				(v_{\bq})_k \mbox{ has the periodic boundary condition} &\mbox{ on } \partial \Omega,\\
				\displaystyle\int_{\Omega} K^*K(v_{\bq})_k dx\,=\,\displaystyle\int_{\Omega} K^*f_k dx,\\
				k=1,2,3
			\end{cases}
		\end{equation}
		minimizes (\ref{eq.model.deblurring}). Scheme (\ref{eq.split1.1})-(\ref{eq.split1.3}) can be applied to solve (\ref{eq.model1.deblurring}) by replacing $I_{\Sigma_f}(\bq)$ by $I_{\tilde{\Sigma}_f}(\bq)$ and $J_2$ by
		\begin{align*}
			\tilde{J}_2(\bq,\bG)=\frac{1}{2\eta} \sum_{k=1}^3 \int_{\Omega} |K(v_{\bq})_k-f_k|^2 d\bx
		\end{align*}
		in (\ref{eq.split1.3}). The solution to the new subproblem can be solved as $\bp^{n+1}=\nabla \bu^{n+1}$, where
		\begin{equation}
			\begin{cases}
				-\eta \nabla^2u_k^{n+1}+\tau K^*Ku_k^{n+1}=-\eta \nabla\cdot \bp_k^{n+2/3}+\tau K^*f_k & \mbox{ in } \Omega,\\
				u_k^{n+1} \mbox{ has the periodic boundary condition} &\mbox{ on } \partial\Omega, \\
				k=1,2,3
			\end{cases}
			\label{eq.deblurring.frac3.pde.u.fft}
		\end{equation}
		with $K^*u_i=k^*(x_1,x_2)\star u_i$ and $k^*$ being the $\cL^2$ adjoint of $k$. (\ref{eq.deblurring.frac3.pde.u.fft}) can be solved by FFT using the celebrated convolution theorem.
	\end{remark}
	\section{Space discretization}\label{sec.discretization}
	\subsection{Basic discrete operators}
	We assume $\Omega\in \RR^2$ to be a rectangle with $M\times N$ pixels. $x_1,x_2$ are used to denote the two spatial directions along which all functions are assumed to be periodic. We use spatial step $\Delta x_1=\Delta x_2=h$. For a vector-valued function $\bbf=(f_1,f_2,f_3)^T$ (resp. scalar-valued function $f$) defined on $\Omega$, we denote its $(i,j)$-th pixel by $\bbf(i,j)=(f_1(i,j),f_2(i,j),f_3(i,j))^T$ (resp. $f(i,j)$). By taking into account the periodic boundary condition, the backward ($-$) and forward ($+$) approximation for $\partial f/\partial x_1$ and $\partial f/\partial x_2$  are defined as
	\begin{align*}
		&\partial_1^- f(i,j)=\begin{cases}
			\frac{f(i,j)-f(i-1,j)}{h}, &\ 1<i\leq M,\\
			\frac{f(1,j)-f(M,j)}{h}, &\ i=1,
		\end{cases}
		& \partial_1^+ f(i,j)=\begin{cases}
			\frac{f(i+1,j)-f(i,j)}{h}, &\ 1\leq i< M,\\
			\frac{f(1,j)-f(M,j)}{h}, &\ i=M,
		\end{cases}\\
		&\partial_2^- f(i,j)=\begin{cases}
			\frac{f(i,j)-f(i,j-1)}{h}, &\ 1<i\leq N,\\
			\frac{f(i,1)-f(i,N)}{h}, &\ i=1,
		\end{cases}
		& \partial_2^+ f(i,j)=\begin{cases}
			\frac{f(i,j+1)-f(i,j)}{h}, &\ 1\leq i< N,\\
			\frac{f(i,1)-f(i,N)}{h}, &\ i=N.
		\end{cases}
	\end{align*}
	Based on the above notations, the backward ($-$) and forward ($+$) approximation of the gradient $\nabla$ and the divergence $\diver$ are defined as
	\begin{align*}
		\nabla^{\pm}f(i,j)=(\partial_1^{\pm}f(i,j),\partial_2^{\pm}f(i,j)),\ \diver^{\pm}\bp(i,j)=\partial_1^{\pm}p_1,(i,j)+  \partial_2^{\pm}p_2(i,j).
	\end{align*}

	In our discretization, only $\nabla^+$ and $\diver^-$ are used so that $\diver^-\nabla^+f$ recovers the central difference approximation of $\nabla^2f$.
	
	Define the shifting and identity operator as
	\begin{align}
		\mathcal{S}_1^{\pm}f(i,j)=f(i\pm 1,j),\quad \mathcal{S}_2^{\pm}f(i,j)=f(i,j\pm 1),\quad \mathcal{I} f(i,j)=f(i,j),
	\end{align}
	and denote the discrete Fourier transform and its inverse by $\cF$ and $\cF^{-1}$, respectively. We have
	\begin{align*}
		\cF (\mathcal{S}_1^{\pm}f)(i,j)=e^{\pm 2\pi\sqrt{-1}(i-1)/M}\cF(f)(i,j),\ \cF (\mathcal{S}_2^{\pm}f)(i,j)=e^{\pm 2\pi\sqrt{-1}(j-1)/N}\cF(f)(i,j).
	\end{align*}
	
	We use $\Real(\cdot)$ to denote the real part of its argument.
	\subsection{Numerical solution for $\bp^{n+1/3}$ in (\ref{eq.frac1.p})}
	In this subsection we discretize the updating formula of $\bp^{n+1/3}$ in problem (\ref{eq.frac1.p}). We set $s_1=1$.
	In $s_2$, the divergence is approximated by $\nabla\cdot \blambda_k^n=\diver^- \blambda_k^n$ for $k=1,2,3$. $s_2$ is computed as
	\begin{align*}
		s_2(i,j)=\sum_{k=1}^3|\diver^- \blambda_k^n(i,j)|^2.
	\end{align*}
	Then $\partial E_1/\partial q_{kr}, \partial^2 E_1/\partial q_{kr}^2, \partial m/\partial q_{kr}$ and $\partial^2 m/\partial q_{kr}^2$ can be computed pointwise. Set an initial condition $\bq^{(0)}$. Then $\bq$ is updated through (\ref{eq.update.q}) until it converges. Denoting the converged variable by $\bq^{*}$, we update $\bp^{n+1/3}=\bq^*$. In our algorithm, $\bq^{(0)}=\bq^n$ is used.
	
	\subsection{Numerical solution for $\blambda^{n+1/3}$ in (\ref{eq.frac1.pde.lambda.fft})}
	\label{sec.num.frac1.lambda}
	Problem (\ref{eq.frac1.pde.lambda.fft}) is discretized as
	\begin{equation}
		\gamma_1\blambda_k^{n+1/3}-2\beta\tau\nabla^+ \left(\frac{\diver^- \blambda_k^{n+1/3}}{\sqrt{g^{n+1/3}}}\right)=\gamma_1\blambda_k^n
		\label{eq.frac2.lambda.dis}
	\end{equation}
	for $k=1,2,3$.
	Instead of solving (\ref{eq.frac2.lambda.dis}), we use the \emph{frozen coefficient approach} (see \cite{deng2019new,he2020curvature}) to solve
	\begin{equation}
		\gamma_1\blambda_k^{n+1/3}-c_1\nabla^+(\diver^-\blambda_k^{n+1/3})=\gamma_1\blambda_k^n -\nabla^+\left[\left(c_1-2\beta\tau/\sqrt{g^{n+1/3}}\right)\diver^-\blambda_k^{n} \right]
		\label{eq.frac2.lambda.dis.1}
	\end{equation}
	with some properly chosen $c_1$. We suggest to use $c_1=\max_{i,j} 2\beta\tau/\sqrt{g^{n+1/3}(i,j)}$. (\ref{eq.frac2.lambda.dis.1}) can be written in matrix form
	\begin{align}
		\begin{pmatrix}
			\gamma_1-c_1\partial_1^+\partial_1^- & -c_1\partial_1^+\partial_2^-\\
			-c_1\partial_2^+\partial_1^- & \gamma_1-c_1\partial_2^+\partial_2^-
		\end{pmatrix}
		\begin{pmatrix}
			\lambda_{k1}^{n+1/3}\\ \lambda_{k2}^{n+1/3}
		\end{pmatrix}=
		\begin{pmatrix}
			w_1\\ w_2
		\end{pmatrix}
		\label{eq.frac2.lambda.dis.2}
	\end{align}
	with
	\begin{align*}
		&w_1=\gamma_1\lambda_{k1}^n-\partial_1^+\left[\left(c_1-2\beta\tau/\sqrt{g^{n+1/3}}\right)\diver^- \blambda_k^n\right],\\
		&w_2=\gamma_1\lambda_{k2}^n-\partial_2^+\left[\left(c_1-2\beta\tau/\sqrt{g^{n+1/3}}\right)\diver^- \blambda_k^n\right].
	\end{align*}
	The linear system (\ref{eq.frac2.lambda.dis.2}) is equivalent to
	\begin{align}
		\begin{pmatrix}
			\gamma_1-c_1(\cS_1^+-\cI)(\cI-\cS_1^-)/h^2 & -c_1(\cS_1^+-\cI)(\cI-\cS_2^-)/h^2\\
			-c_1(\cS_2^+-\cI)(\cI-\cS_1^-)/h^2 & \gamma_1-c_1(\cS_2^+-\cI)(\cI-\cS_2^-)/h^2
		\end{pmatrix}
		\begin{pmatrix}
			\lambda_{k1}^{n+1/3}\\ \lambda_{k2}^{n+1/3}
		\end{pmatrix}=
		\begin{pmatrix}
			w_1\\ w_2
		\end{pmatrix}.
		\label{eq.frac2.lambda.dis.3}
	\end{align}
	Applying the discrete Fourier transform on both sides gives rise to
	\begin{align*}
		\begin{pmatrix}
			a_{11} & a_{12}\\ a_{21} & a_{22}
		\end{pmatrix}
		\cF\begin{pmatrix}
			\lambda_{k1}^{n+1/3}\\ \lambda_{k2}^{n+1/3}
		\end{pmatrix}=
		\cF\begin{pmatrix}
			w_1\\ w_2
		\end{pmatrix}
	\end{align*}
	where
	\begin{align*}
		&a_{11}=\gamma_1-2c_1(\cos z_i-1)/h^2,\ a_{22}=\gamma_1-2c_1(\cos z_j-1)/h^2,\\
		&a_{12}=-c_1(1-\cos z_i-\sqrt{-1}\sin z_i)(1-\cos z_j+\sqrt{-1}\sin z_j)/h^2,\\
		&a_{21}=-c_1(1-\cos z_j-\sqrt{-1}\sin z_j)(1-\cos z_i+\sqrt{-1}\sin z_i)/h^2,
	\end{align*}
	with $z_i=2\pi(i-1)/M,z_j=2\pi(j-1)/N$ for $i=1,...,M$ and $j=1,...,N$. We have
	\begin{align*}
		\begin{pmatrix}
			\lambda_{k1}^{n+1/3}\\ \lambda_{k2}^{n+1/3}
		\end{pmatrix}=\Real\left(\cF^{-1}\left[\frac{1}{a_{11}a_{22}-a_{12}a_{21}}
		\begin{pmatrix}
			a_{22}\cF(w_1)-a_{12}\cF(w_2)\\
			-a_{21}\cF(w_1)+a_{22}\cF(w_2)
		\end{pmatrix} \right]\right).
	\end{align*}
	\subsection{Numerical solution for $(\bp^{n+2/3},\blambda^{n+2/3})$ in (\ref{eq.frac2.lambda})}
	(\ref{eq.frac2.lambda}) can be written as the energy $E_2$ (in (\ref{eq.frac2.lambda.1})) which is a quadratic functional of $\mu$. $\blambda^{n+2/3}$ can be updated by solving (\ref{eq.frac2.lambda.update}) pointwise. Then we update
	\begin{equation}
		\bp_k^{n+2/3}=\frac{1}{\sqrt{g^{n+1/3}}}\blambda_k^{n+2/3} \bG^{n+1/3}
	\end{equation}
	for $k=1,2,3$.
	\subsection{Numerical solution for $\bp^{n+1}$ in (\ref{eq.frac3.pde.u.fft})}
	We update $\bp^{n+1}$ as $\nabla \bu^{n+1}$ where $\bu^{n+1}=(u_1^{n+1},u_2^{n+1},u_3^{n+1})^T$ is the solution of
	\begin{align}
		\begin{cases}
			-\eta \nabla^2u_k^{n+1}+\tau u_k^{n+1}=-\eta \nabla\cdot \bp_k^{n+2/3}+\tau f_k &\mbox{ in } \Omega,\\
			u_k^{n+1} \mbox{ has the periodic boundary condition} &\mbox{ on } \partial\Omega.\\
		\end{cases}
		\label{eq.frac2.lambda.update.1}
	\end{align}
	We discretize (\ref{eq.frac2.lambda.update.1}) as
	\begin{align}
		-\eta \diver^-(\nabla^+ u_k^{n+1})+\tau u_k^{n+1}=-\eta \diver^- \bp_k^{n+2/3}+\tau f_k
		\label{eq.frac2.lambda.update.2.1}
	\end{align}
	which is equivalent to
	\begin{align}
		\left[-\eta(\cI-\cS_1^-)(\cS_1^+-\cI)/h^2-\eta(\cI-\cS_2^-)(\cS_2^+-\cI)/h^2+c_1\cI\right]u_k^{n+1}=b
		\label{eq.frac2.lambda.update.3}
	\end{align}
	with $b=-\eta \diver^- \bp_k^{n+2/3}+\tau f_k$.
	(\ref{eq.frac2.lambda.update.3}) can be solved efficiently by FFT,
	\begin{align}
		u_k^{n+1}=\Real\left[\cF^{-1}\left( \frac{\cF(b)}{w}\right)\right]
	\end{align}
	where
	\begin{align}
		w(i,j)=\tau\cI-\eta\left(1-e^{-\sqrt{-1}z_i}\right)\left(e^{\sqrt{-1}z_i}-1\right)/h^2- \eta\left(1-e^{-\sqrt{-1}z_j}\right)\left(e^{\sqrt{-1}z_j}-1\right)/h^2
	\end{align}
	with $z_i,z_j$ defined in Section \ref{sec.num.frac1.lambda}. Then we update
	\begin{align}
		\bp_k^{n+1}=\nabla^+ u_k^{n+1}.
	\end{align}
	\subsection{Initialization}
	For initial condition, we first initialize $\bu^0=\bbf$ or $\bu^0=\mathbf{0}$. Then we initialize $\bp^0=(\bp_1^0,\bp_2^0,\bp_3^0)^T$ as
	\begin{align}
		\bu^0=\bbf  \mbox{ and }\bp_k^0=\nabla^+ u_k^0
	\end{align}
	for $k=1,2,3$ and
	\begin{align}
		\bG^0=\bM(\bp^0), g^0=m(\bG^0).
	\end{align}
	$\blambda^0=(\blambda_1^0,\blambda_2^0,\blambda_3^0)^T$ is computed as
	\begin{align}
		\blambda_k^0=\sqrt{g^0}\bp^0\bG^{-1}=\frac{1}{\sqrt{g^0}}\begin{pmatrix}
			g_{22}^0p_{k1}^0-g_{12}^0p_{k2}^0\\
			-g_{21}^0p_{k1}^0+g_{11}^0p_{k2}^0
		\end{pmatrix}.
	\end{align}
	
	\section{Numerical experiments}\label{sec.experiments}
	In this section, we present numerical results that demonstrate the effectiveness of the proposed model and solver.
	All experiments are implemented in MATLAB(R2018b) on a laptop of 8GB RAM and Intel Core i7-4270HQ CPU: \@2.60 GHz.
	For all of the images used, the pixel values are in $[0,1]$.
	For simplicity, $h=1$ is used.
	For all experiments, $tol=10^{-6}$ is used in the Newton method (\ref{eq.update.q}) to update $\bp^{n+1/3}$.
	
	We consider noisy images with Gaussian noise and  Poisson noise.
	For Gaussian noise, the parameter is the standard deviation (denoted by $SD$).
	The larger  $SD$ is, the heavier the noise.
	In Poisson noise case, the parameter is the number of photons (denoted by $P$).
	Images have better quality with more photons.
	When adding Poisson noise, we use the MATLAB function \emph{imnoise}.
	To add Poisson noise with $P$ photons to an image $\bv$, we refer to the function as \texttt{imnoise($\bv \ast P/10^{12}$,'poisson')$\ast 10^{12}/P$}.
	
	In our experiments, when not specified otherwise, we use $\tau=0.05,\gamma_1=1, \gamma_2=3$,
	$c_1=\max_{i,j} 2\beta\tau/\sqrt{g^{n+1/3}(i,j)}$,
	and stopping criterion $\|u^{n+1}-u^n\|_2\leq 10^{-2}$.
	
	\begin{figure}[t!]
		\centering
		\begin{tabular}{ccc}
			(a) &(b) &(c)\\
			\includegraphics[height=0.28\textwidth]{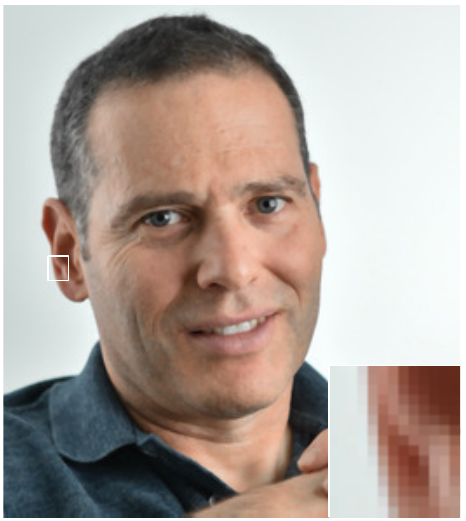}&
			\includegraphics[height=0.28\textwidth]{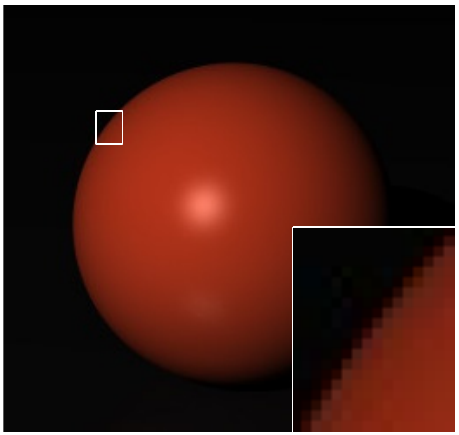}&
			\includegraphics[height=0.28\textwidth]{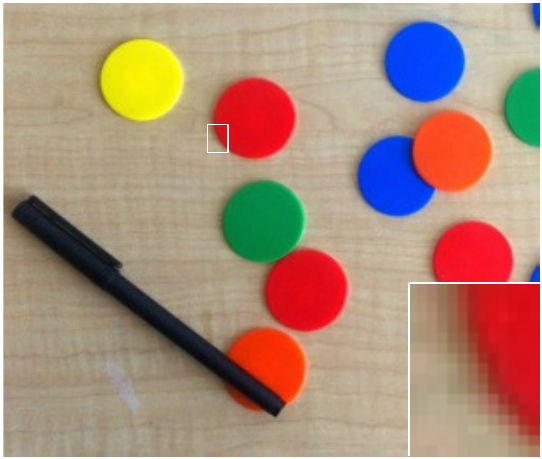}\\
			\includegraphics[height=0.28\textwidth]{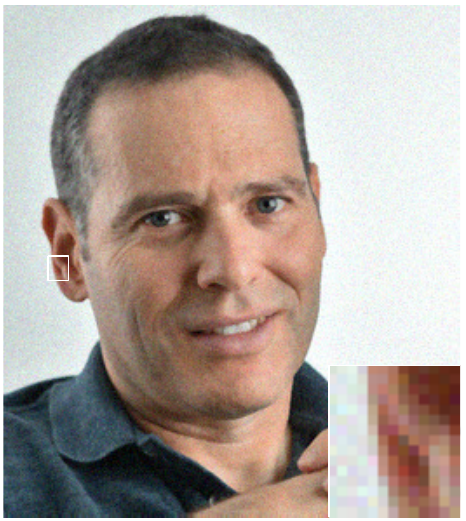}&
			\includegraphics[height=0.28\textwidth]{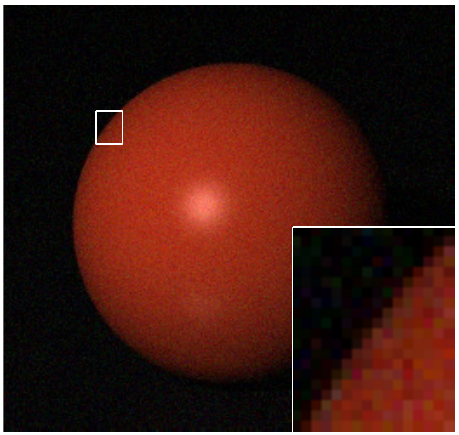}&
			\includegraphics[height=0.28\textwidth]{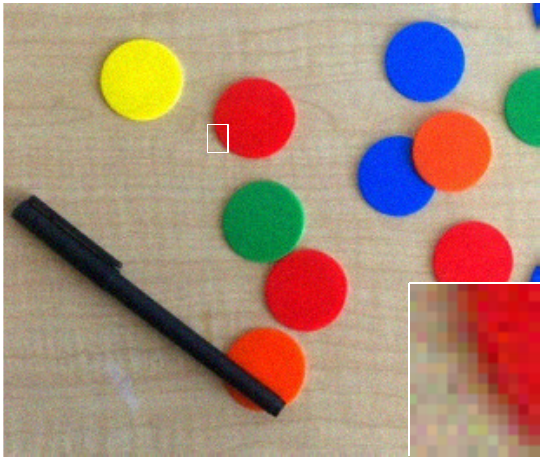}\\
			\includegraphics[height=0.28\textwidth]{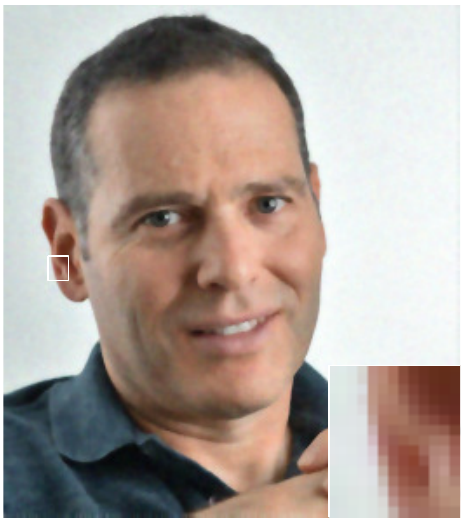}&
			\includegraphics[height=0.28\textwidth]{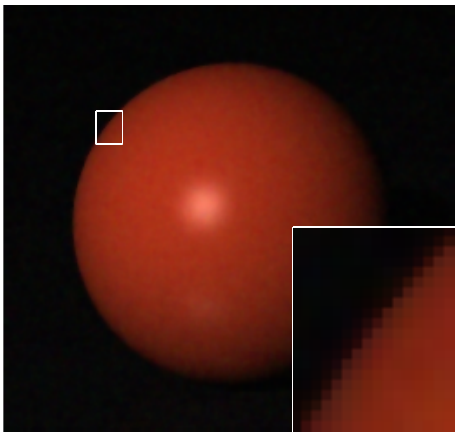}&
			\includegraphics[height=0.28\textwidth]{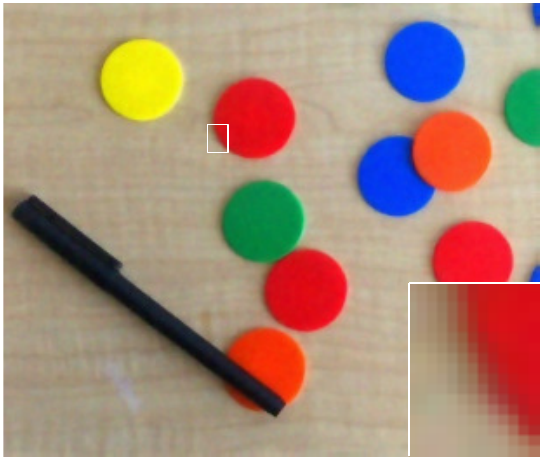}
		\end{tabular}
		\caption{(Gaussian noise with $SD=0.03$. $\alpha=0.01,\eta=0.5$.)
			Image denoising by the proposed method on (a) the portrait, (b) the orange ball and (c) chips.
			The first row shows clean images.
			The second row shows noisy images.
			The third row shows denoised images.
			$\beta=0.01$ is used for the orange ball, and $\beta=0.005$ is used for the portrait and chips.}
		\label{fig.Gaussian.light}
	\end{figure}
	
	\begin{figure}[t!]
		\centering
		\begin{tabular}{ccc}
			\includegraphics[width=0.3\textwidth]{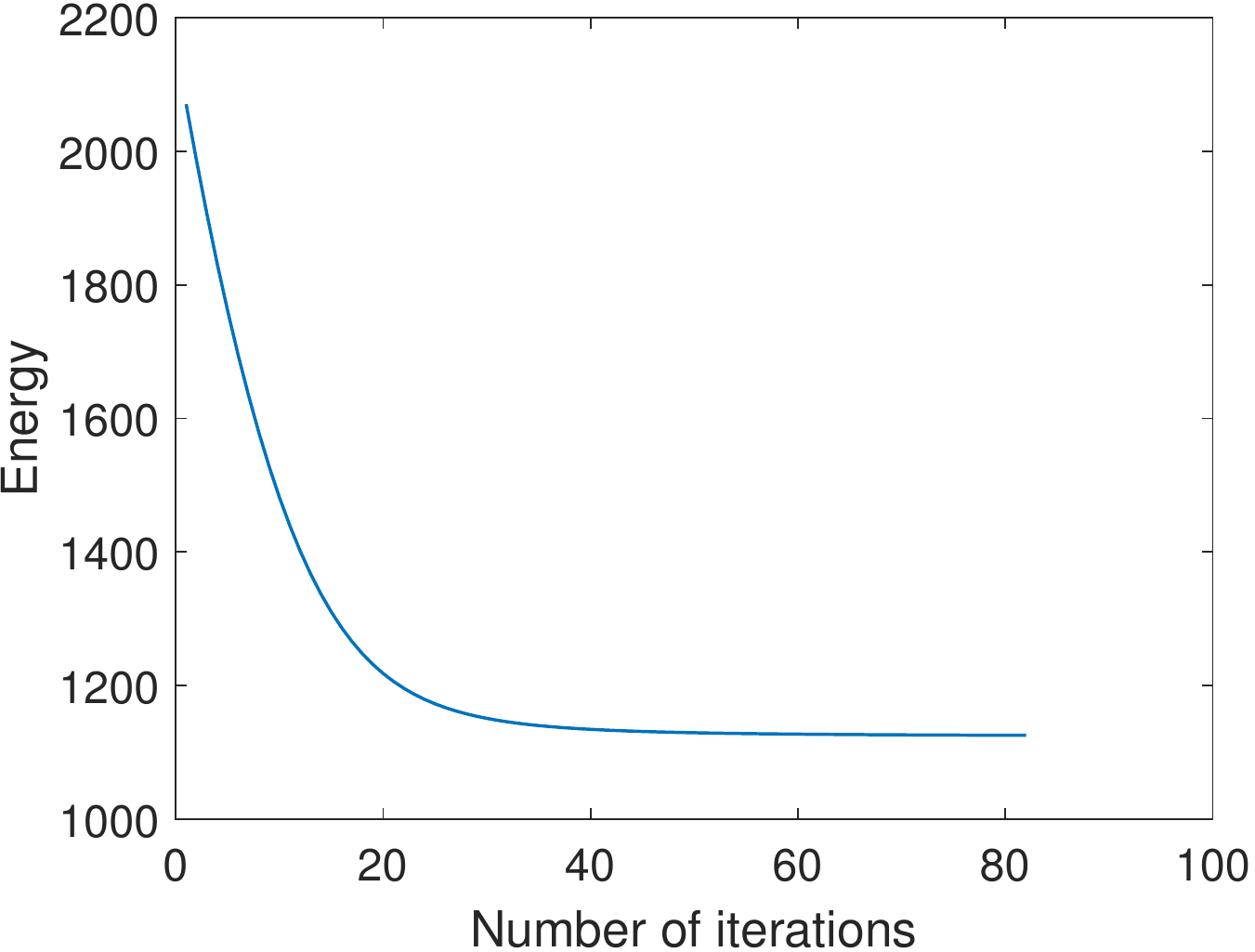}&
			\includegraphics[width=0.3\textwidth]{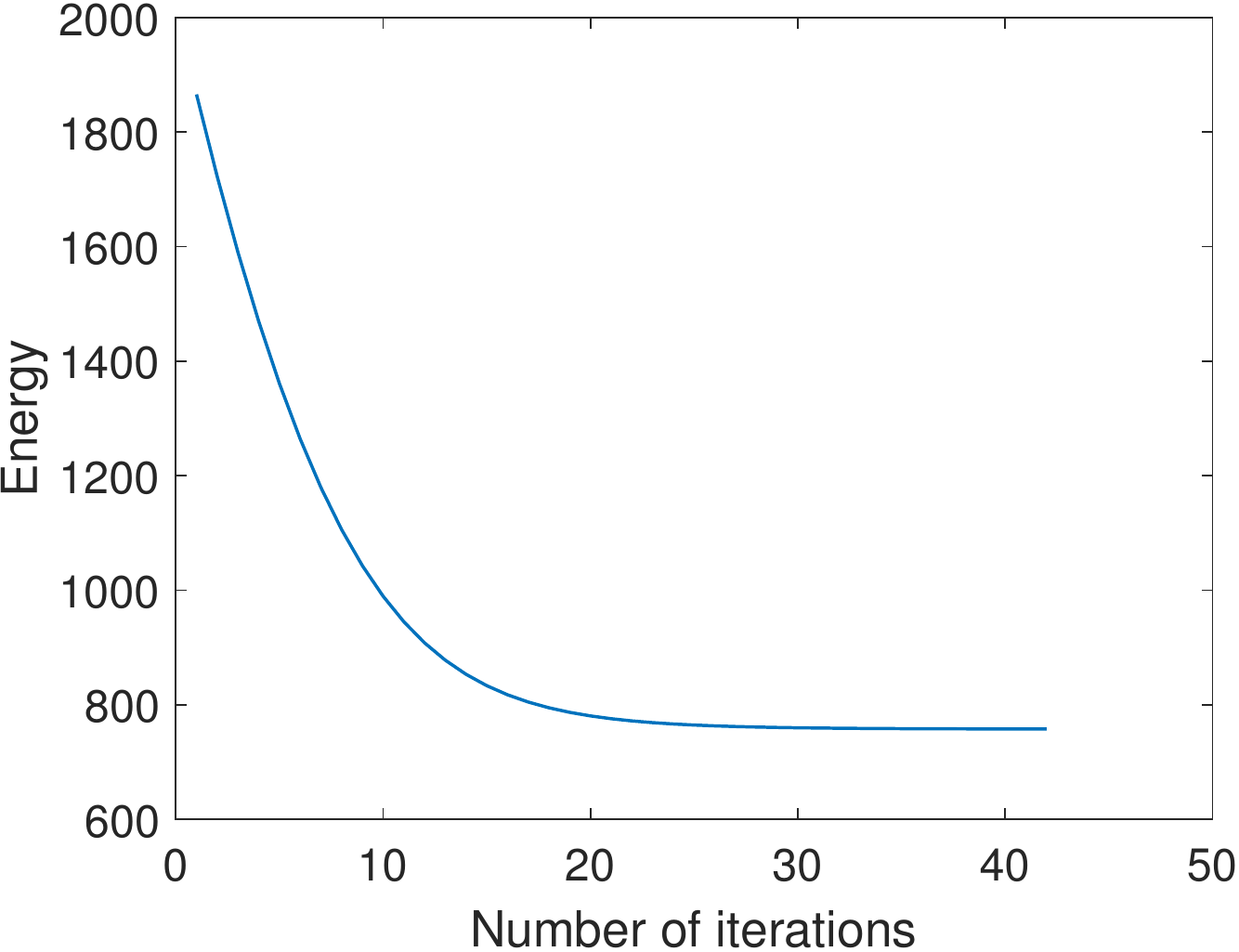}&
			\includegraphics[width=0.3\textwidth]{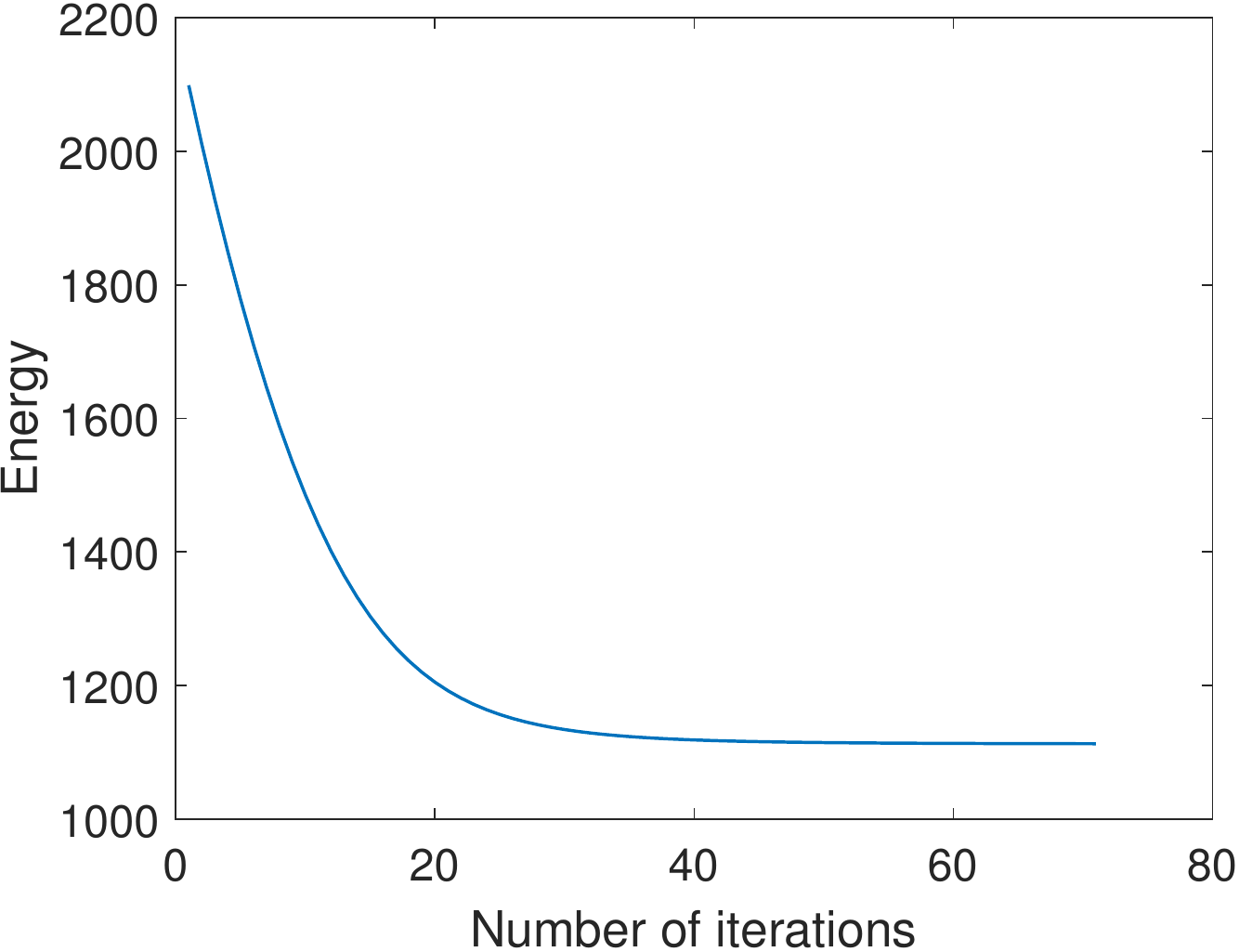}\\
			\includegraphics[width=0.3\textwidth]{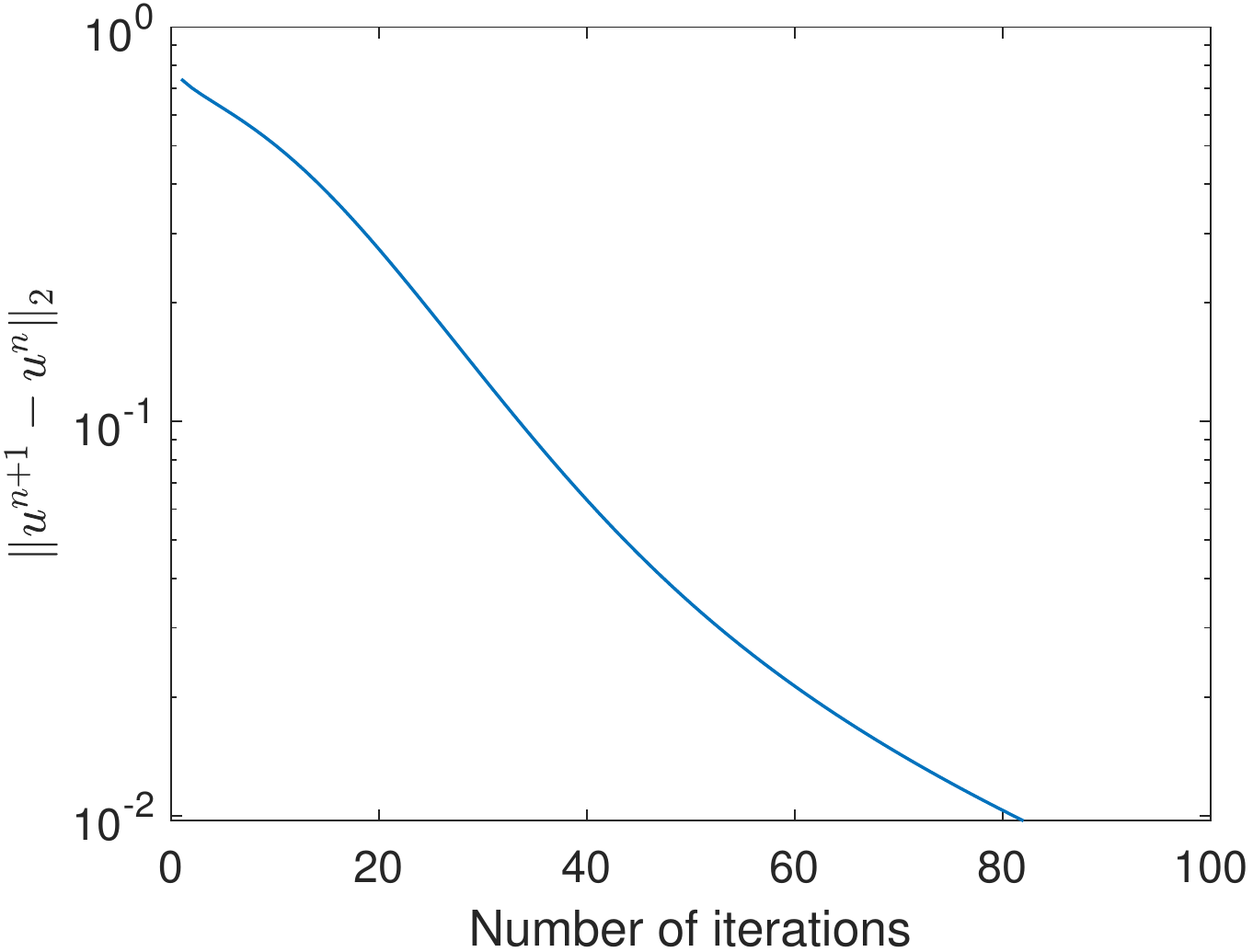}&
			\includegraphics[width=0.3\textwidth]{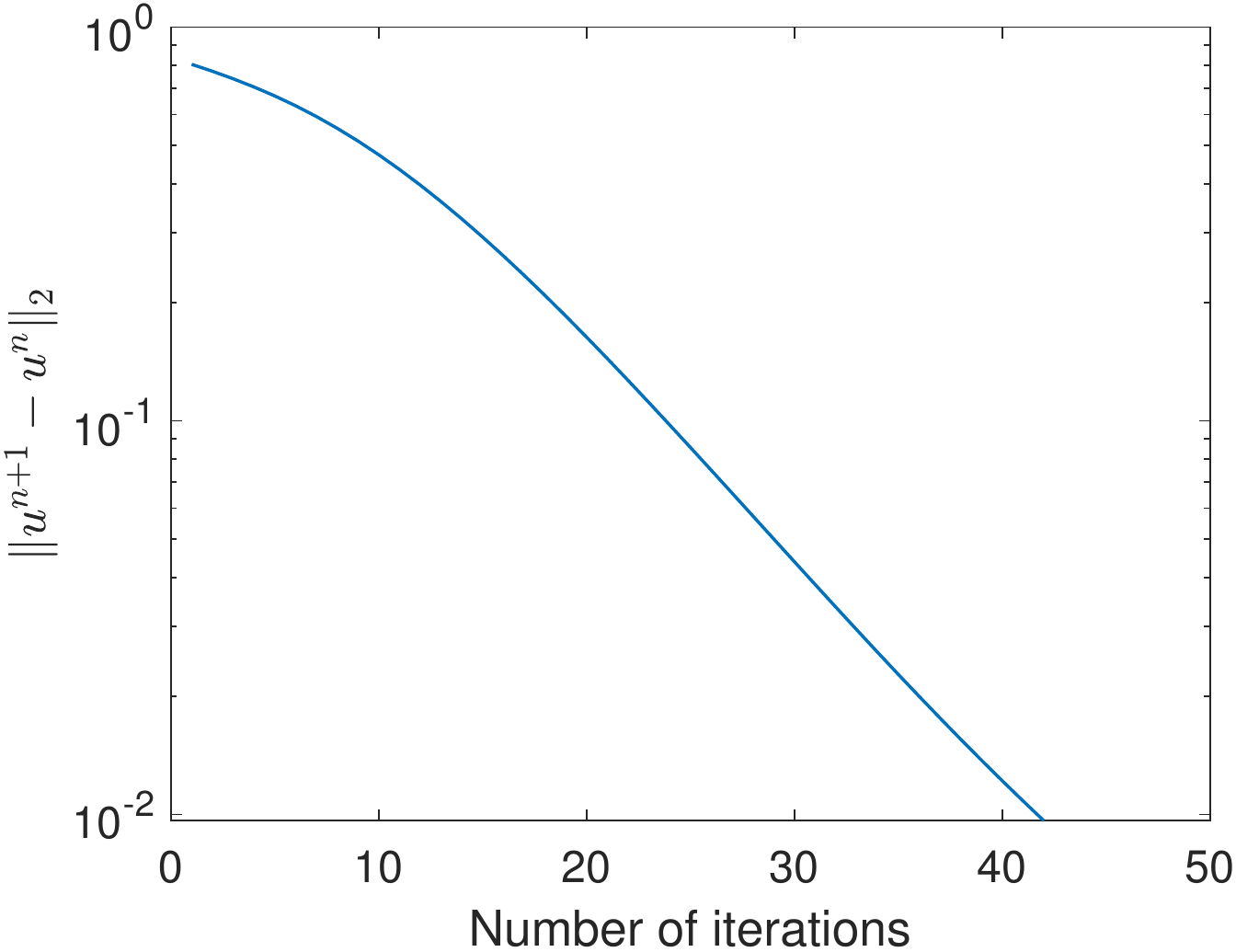}&
			\includegraphics[width=0.3\textwidth]{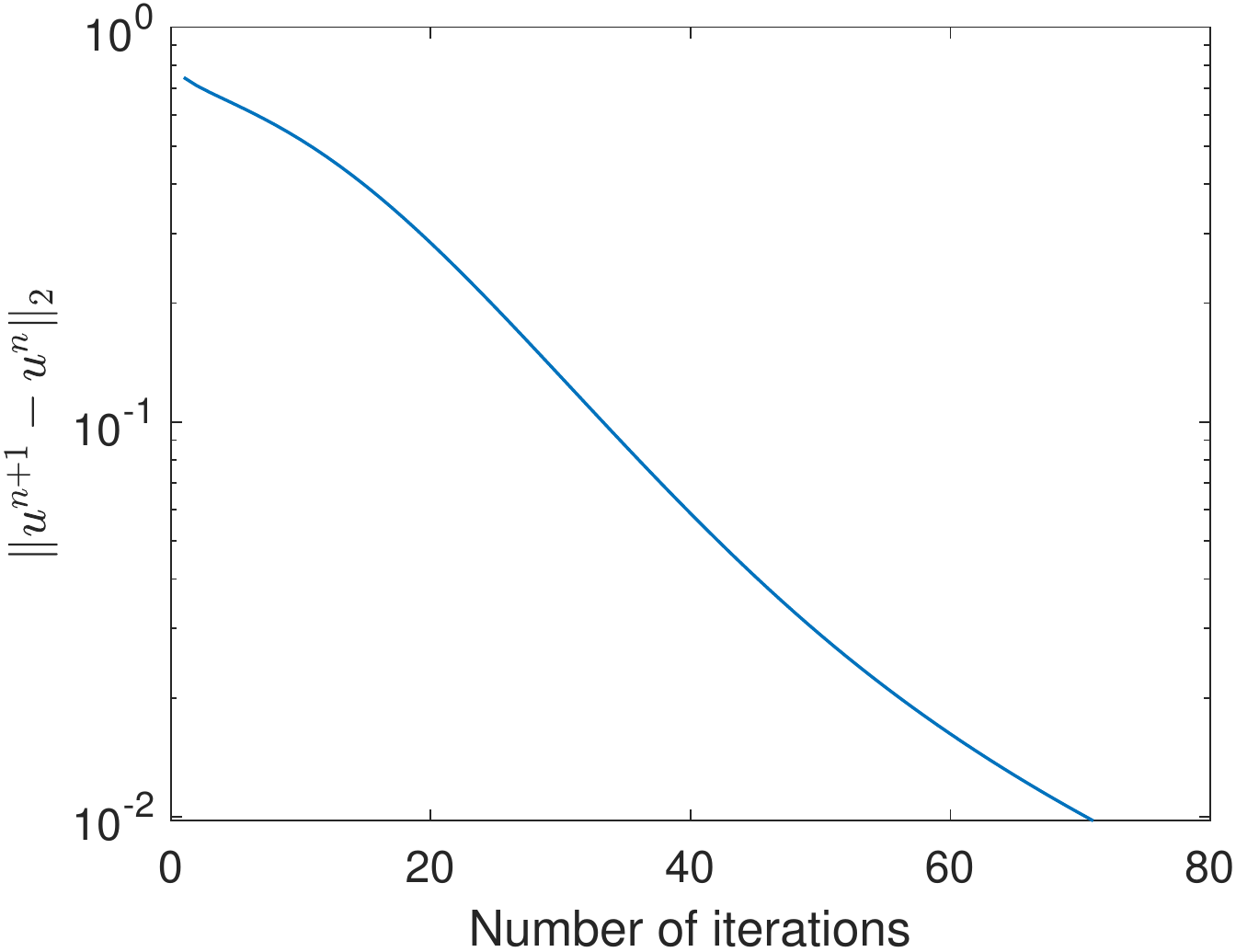}
		\end{tabular}
		\caption{(Gaussian noise with $SD=0.03$. $\alpha=0.01,\eta=0.5$.) The evolution of (first row) the energy and (second row) $\|u^{n+1}-u^n\|_2$ w.r.t. the number of iterations for results in Figure \ref{fig.Gaussian.light}. First column: the portrait. Second column: the orange ball. Third column: chips. $\beta=0.01$ is used for the orange ball, and $\beta=0.005$ is used for the portrait and chips.}
		\label{fig.Gaussian.light.ener}
	\end{figure}
	
	\begin{figure}[t!]
		\centering
		\begin{tabular}{ccc}
			(a) &(b) &(c)\\
			\includegraphics[height=0.28\textwidth]{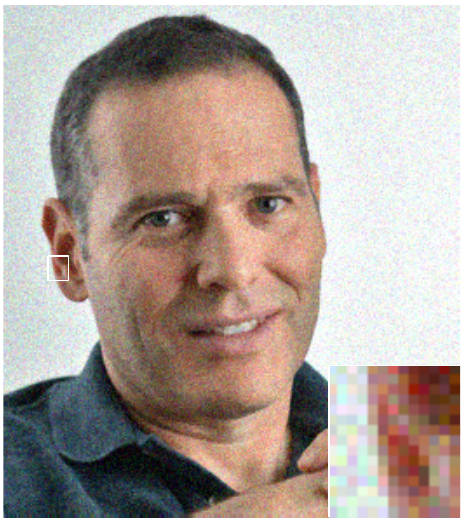}&
			\includegraphics[height=0.28\textwidth]{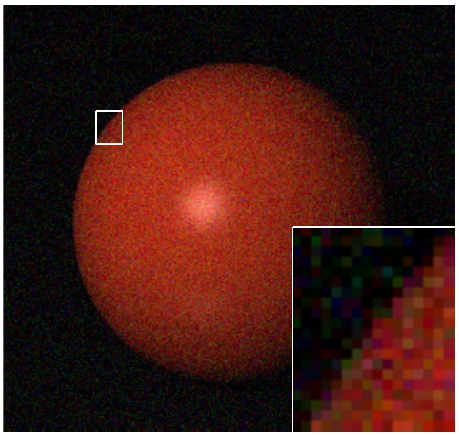}&
			\includegraphics[height=0.28\textwidth]{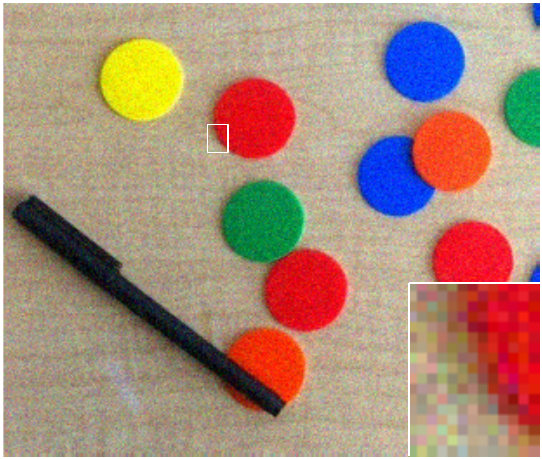}\\
			\includegraphics[height=0.28\textwidth]{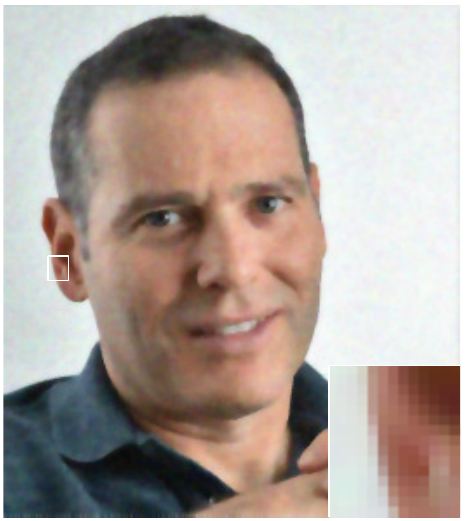}&
			\includegraphics[height=0.28\textwidth]{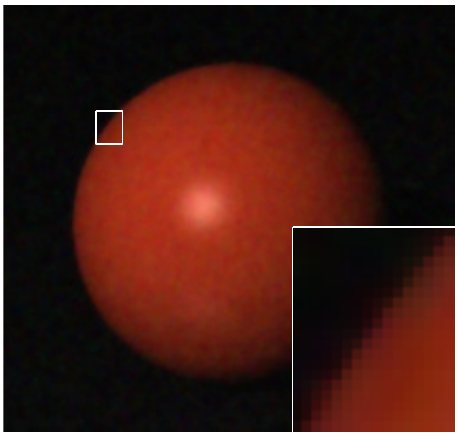}&
			\includegraphics[height=0.28\textwidth]{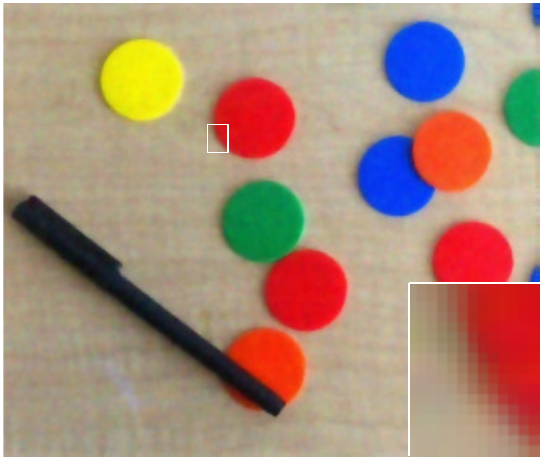}\\
			\includegraphics[width=0.28\textwidth]{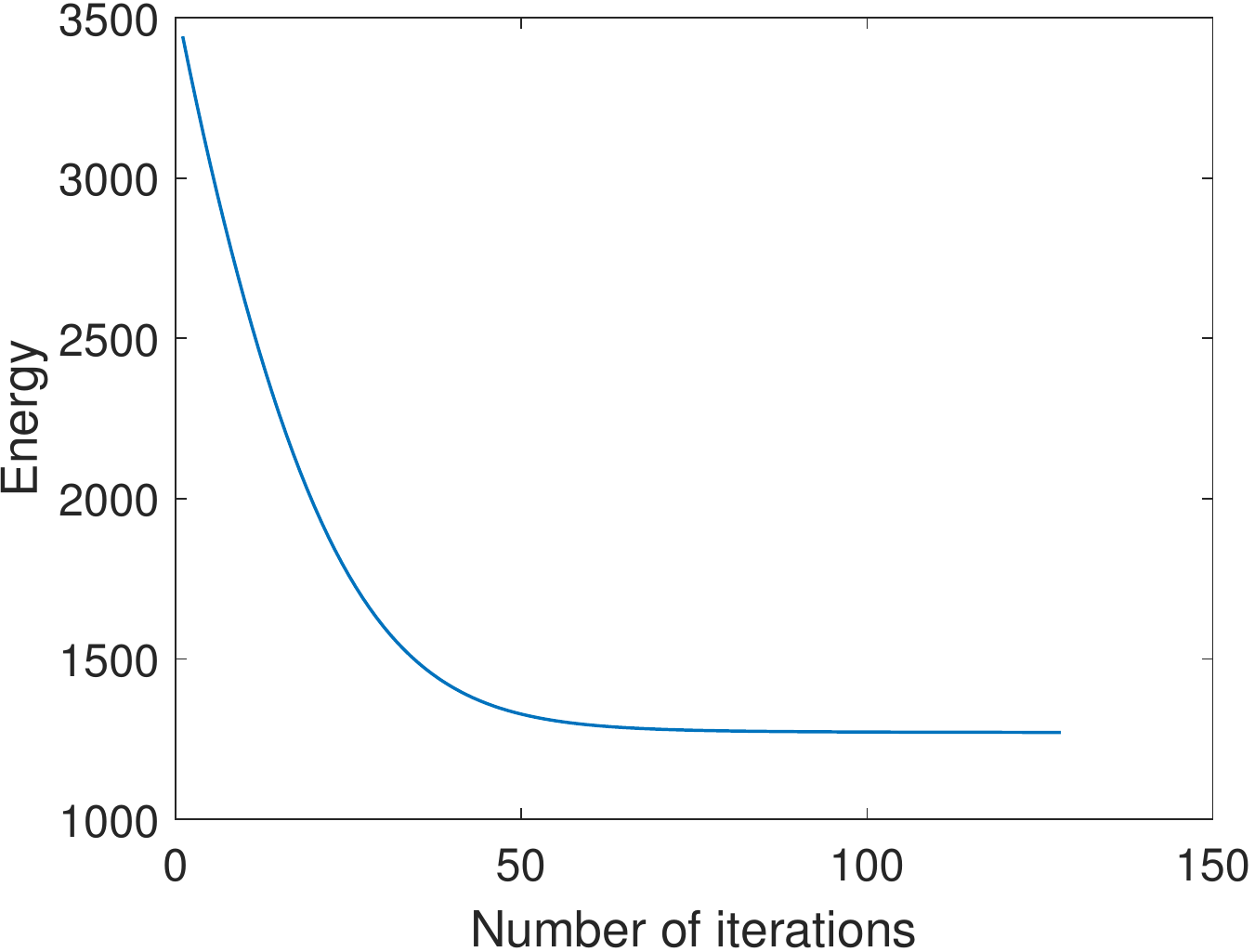}&
			\includegraphics[width=0.28\textwidth]{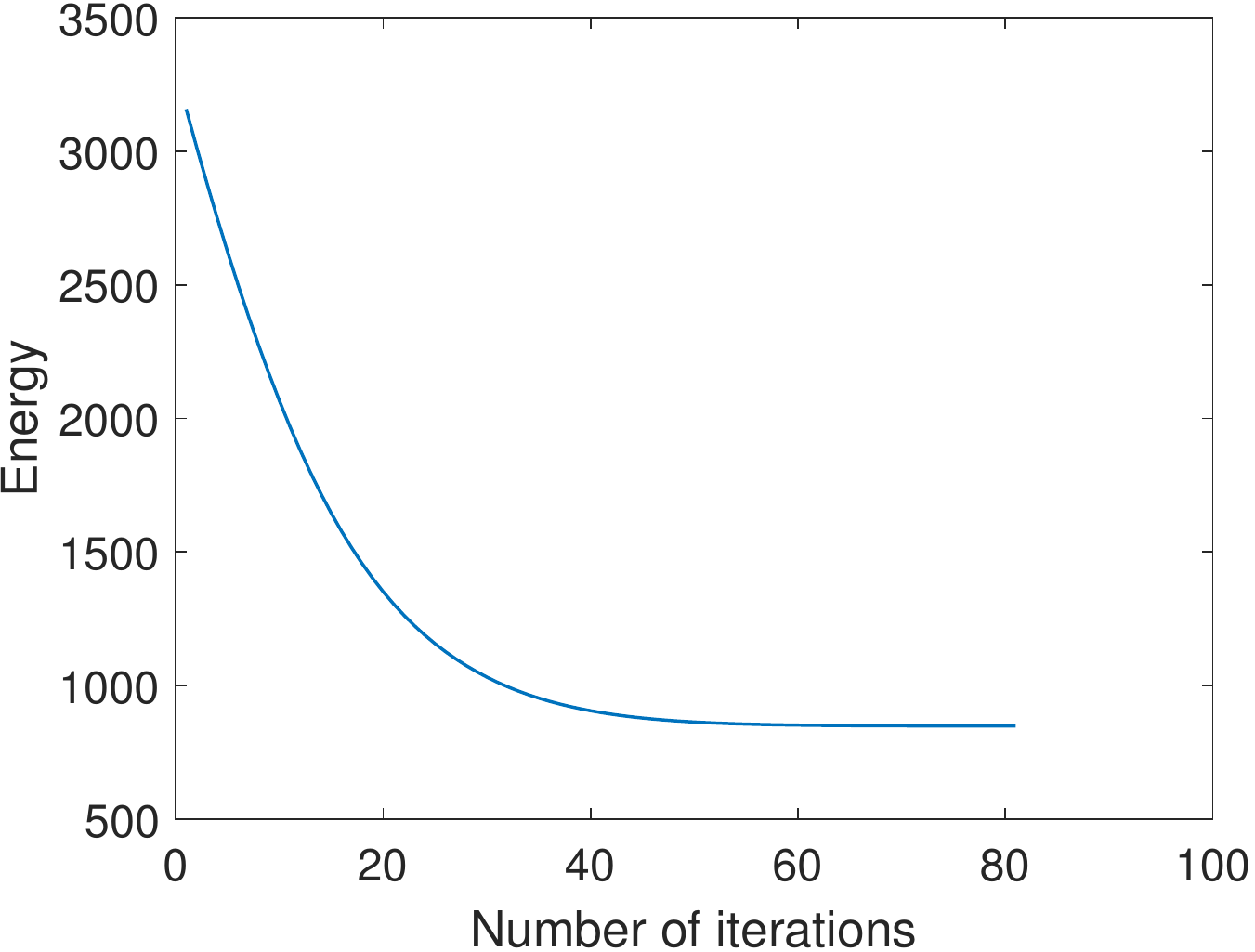}&
			\includegraphics[width=0.28\textwidth]{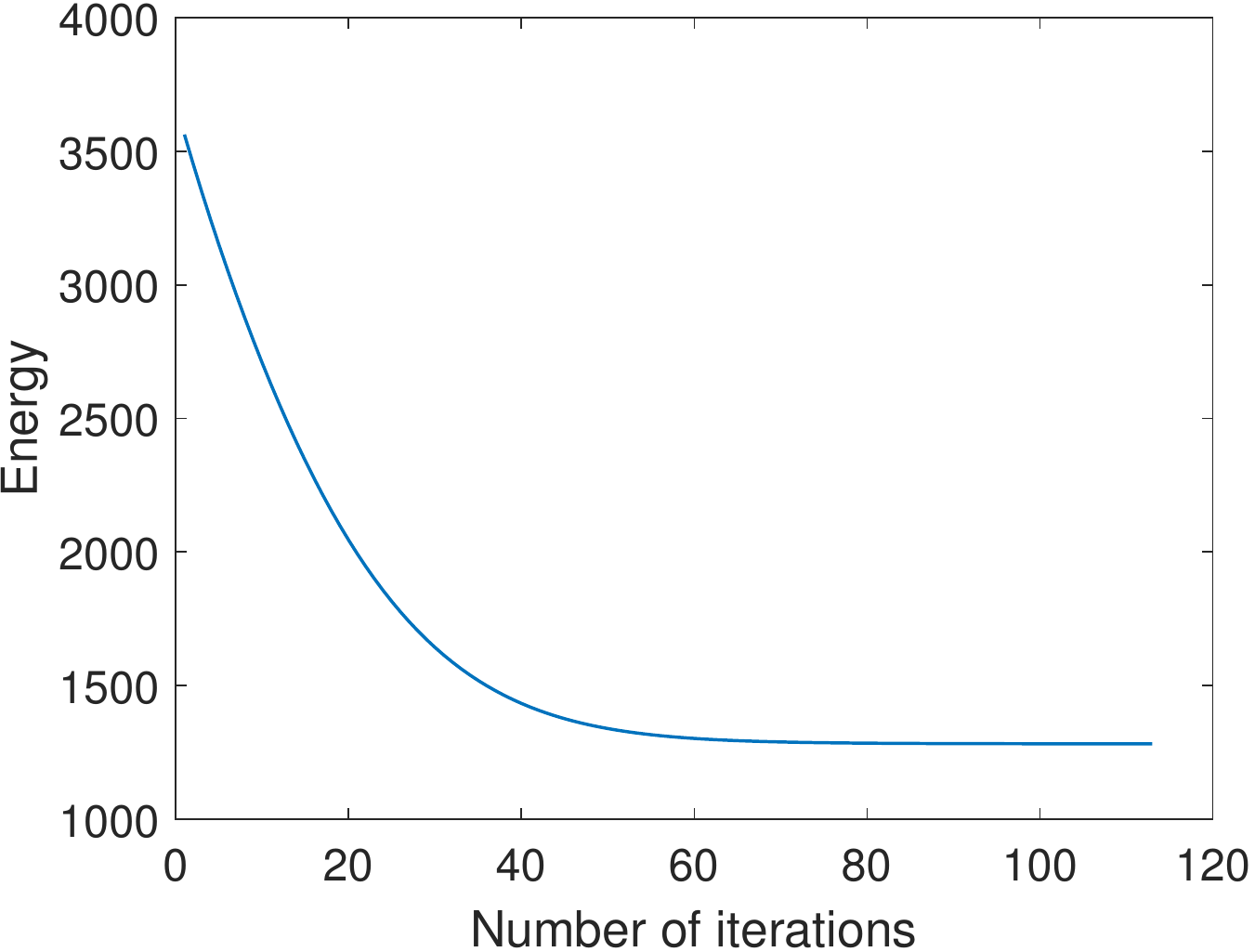}
		\end{tabular}
		\caption{(Gaussian noise with $SD=0.06$. $\alpha=0.01,\eta=1$.) Image denoising by the proposed method on (a) the portrait, (b) the orange ball and (c) chips. The first row shows noisy images. The second row shows denoised images. The third row shows the evolution of the energy w.r.t. the number of iterations. $\beta=0.01$ is used for the orange ball, and $\beta=0.005$ is used for the portrait and chips.}
		\label{fig.Gaussian.heavy}
	\end{figure}
	\begin{figure}[t!]
		\centering
		\begin{tabular}{cc}
			\includegraphics[height=0.28\textwidth]{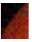}&
			\includegraphics[height=0.28\textwidth]{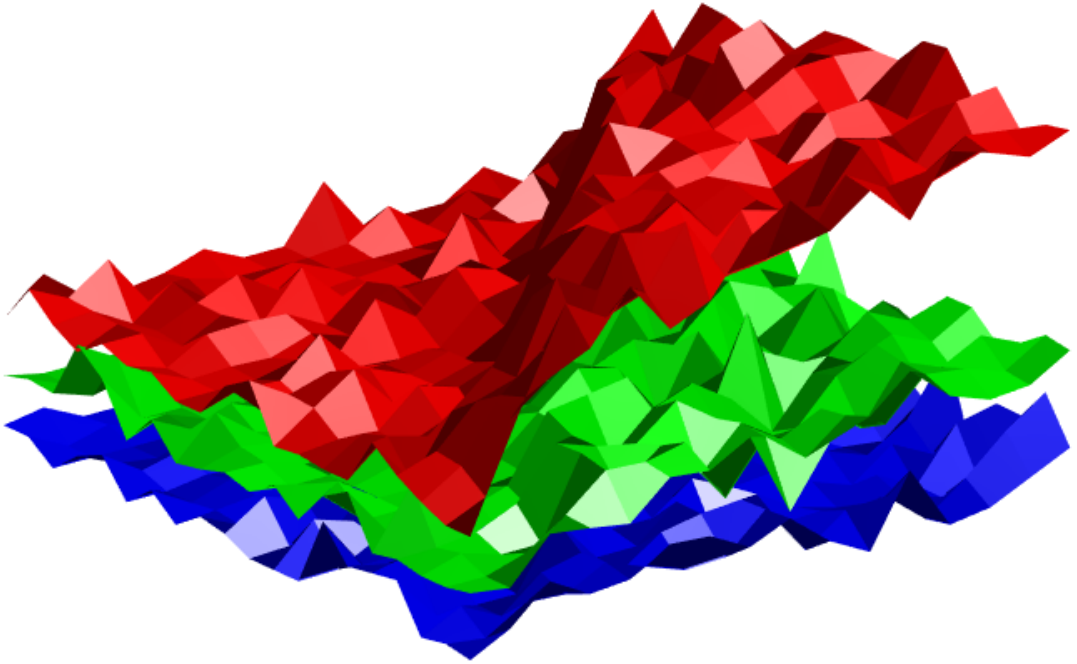}\\
			\includegraphics[height=0.28\textwidth]{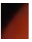}&
			\includegraphics[height=0.28\textwidth]{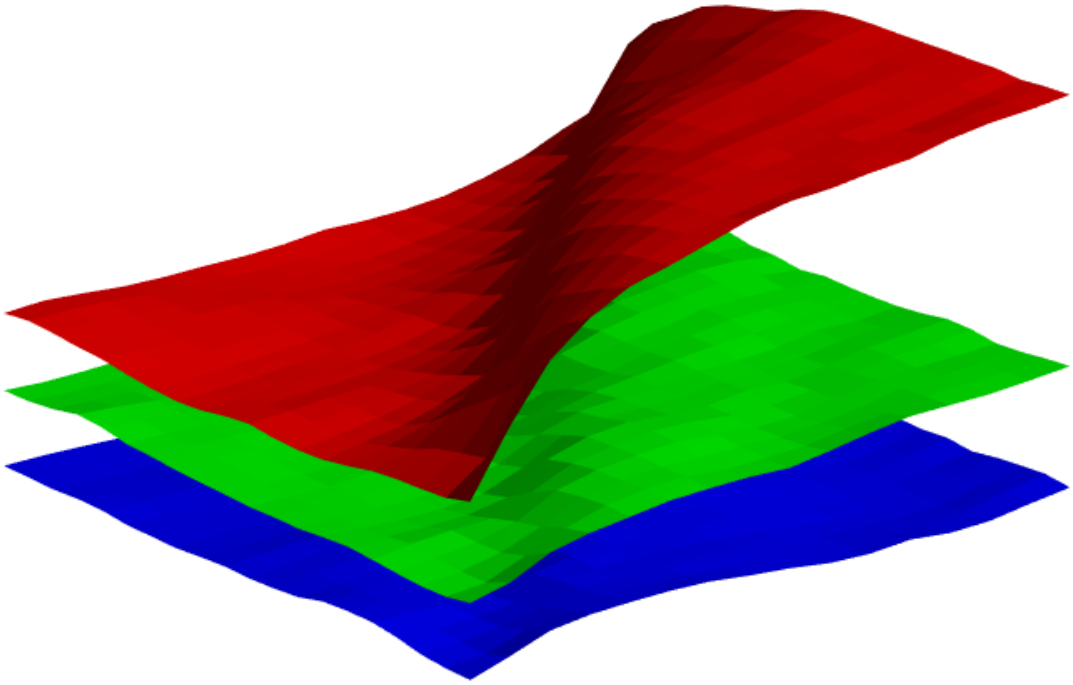}
		\end{tabular}
		\caption{(Gaussian noise with $SD=0.06$. Smoothing effect.) Left: zoomed region of the (first row) noisy and (second row) denoised orange ball in Figure \ref{fig.Gaussian.heavy}. Right: the surface plot of the left images. The RGB channels corresponds to the red, green and blue surface, respectively.}
		\label{fig.Gaussian.ball.surf}
	\end{figure}
	
	\begin{table}[t!]
		\centering
		(a)\\
		\begin{tabular}{c|cc|cc|cc|cc}
			\hline
			& \multicolumn{2}{c|}{Proposed} & \multicolumn{2}{c|}{Polyakov} & \multicolumn{2}{c|}{CTV} & \multicolumn{2}{c}{VTV}\\
			\hline
			Portrait & 33.09 & (0.94) &30.54&(0.91) &31.69 & (0.90) & 32.92 & (0.96)\\
			\hline
			Chips & 33.88 &(0.97) & 31.81 & (0.97) & 32.41 &(0.96) &  33.67 &(0.97)\\
			\hline
		\end{tabular}\vspace{0.2cm}\\
		(b)\\
		\begin{tabular}{c|cc|cc|cc|cc}
			\hline
			& \multicolumn{2}{c|}{Proposed} & \multicolumn{2}{c|}{Polyakov} & \multicolumn{2}{c|}{CTV} & \multicolumn{2}{c}{VTV}\\
			\hline
			Portrait & 128 & (69.73) & 79&(3.12) &99 & (2.62) & 287 & (10.79)\\
			\hline
			Chips & 113 &(61.22) & 90 & (3.77) & 102 &(2.76) &  301 &(14.49)\\
			\hline
		\end{tabular}
		\caption{Corresponding to Figure \ref{fig.Gaussian06.comparison}, (a) The PSNR (SSIM) value of the denoised images; (b) The number of iterations (CPU time in seconds) of different methods. }
		\label{tab.Gaussian06.comparison}
	\end{table}
	
	\begin{figure}[t!]
		\centering
		\begin{tabular}{ccccc}
			Noisy & Proposed & Polyakov & CTV & VTV\\
			\includegraphics[width=0.17\textwidth]{figure/ronZ_G006}&
			\includegraphics[width=0.17\textwidth]{figure/ronZ_G006_eta1_beta5e-3_alpha1e-2}&
			\includegraphics[width=0.17\textwidth]{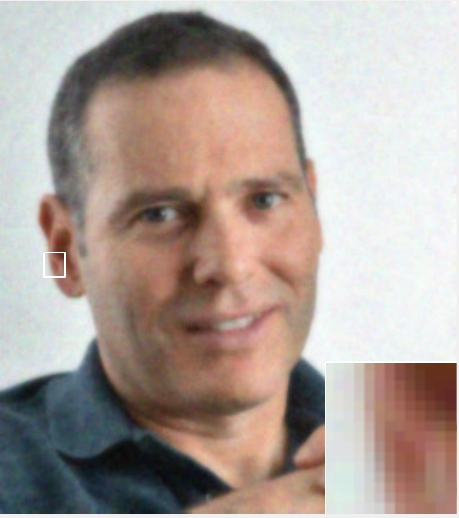}&
			\includegraphics[width=0.17\textwidth]{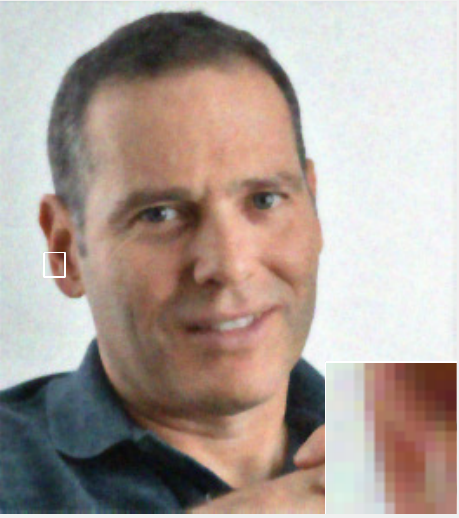}&
			\includegraphics[width=0.17\textwidth]{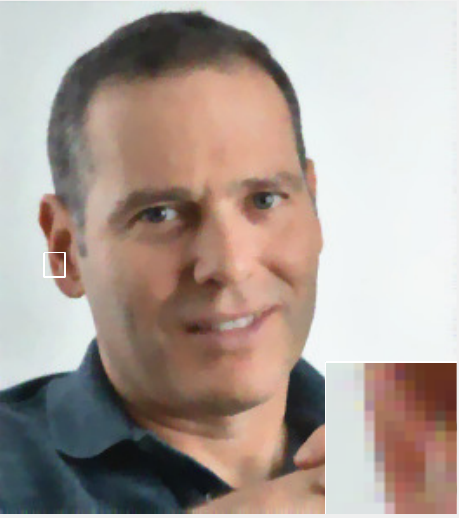}\\
			\includegraphics[width=0.17\textwidth]{figure/chipZ_G006}&
			\includegraphics[width=0.17\textwidth]{figure/chipZ_G006_eta1_beta5e-3_alpha1e-2}&
			\includegraphics[width=0.17\textwidth]{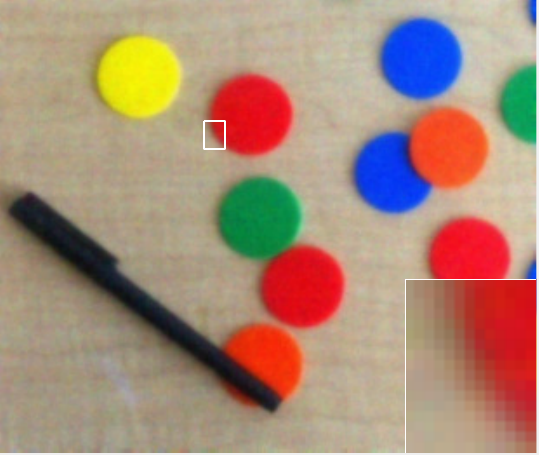}&
			\includegraphics[width=0.17\textwidth]{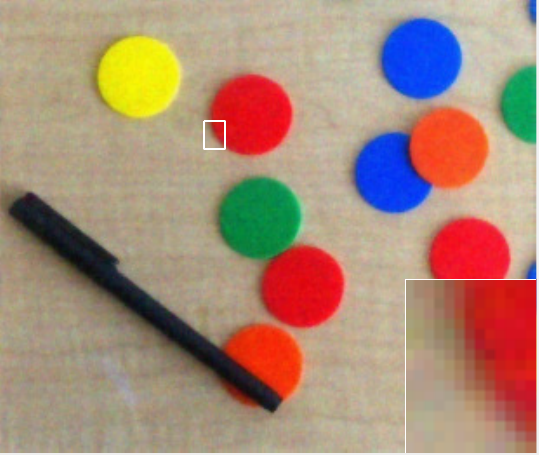}&
			\includegraphics[width=0.17\textwidth]{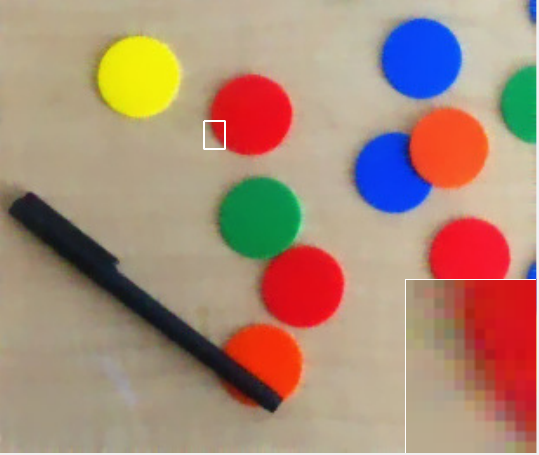}
		\end{tabular}
		\caption{(Gaussian noise with $SD=0.06$.) Comparison of the denoised images by the proposed method and the model of the Polyakov action, CTV, and VTV. First column: noisy images. Second column: results by the proposed method with $\alpha=0.01, \eta=1, \beta=0.005$. Third column: results the Polyakov action model. Fourth column: results by CTV with $\lambda=6$. Fifth column: results by VTV with $\lambda=0.1$.}
		\label{fig.Gaussian06.comparison}
	\end{figure}
	
	\begin{table}[t!]
		\centering
		(a)\\
		\begin{tabular}{c|cc|cc|cc|cc}
			\hline
			& \multicolumn{2}{c|}{Proposed} & \multicolumn{2}{c|}{Polyakov} & \multicolumn{2}{c|}{CTV} & \multicolumn{2}{c}{VTV}\\
			\hline
			Portrait & 27.71 & (0.86) & 26.00&(0.82) &26.30 & (0.80) & 27.33 & (0.88)\\
			\hline
			Vegetables & 27.13 &(0.83) & 26.05 & (0.79) & 25.93 &(0.78) &  26.57 &(0.83)\\
			\hline
		\end{tabular}\vspace{0.2cm}\\
		(b)\\
		\begin{tabular}{c|cc|cc|cc|cc}
			\hline
			& \multicolumn{2}{c|}{Proposed} & \multicolumn{2}{c|}{Polyakov} & \multicolumn{2}{c|}{CTV} & \multicolumn{2}{c}{VTV}\\
			\hline
			Portrait & 153 & (75.88) & 135&(5.56) &118 & (3.17) & 421 & (16.95)\\
			\hline
			Vegetables & 117 &(15.95) & 141 & (1.49) & 125 &(0.75) &  364 &(4.12)\\
			\hline
		\end{tabular}
		\caption{Corresponding to Figure \ref{fig.Gaussian.comparison}, (a) The PSNR (SSIM) value of the denoised images; (b) The number of iterations (CPU time in seconds) of different methods. }
		\label{tab.Gaussian.comparison}
	\end{table}
	
	\begin{figure}[t!]
		\centering
		\begin{tabular}{ccccc}
			Noisy & Proposed & Polyakov & CTV & VTV\\
			\includegraphics[width=0.17\textwidth]{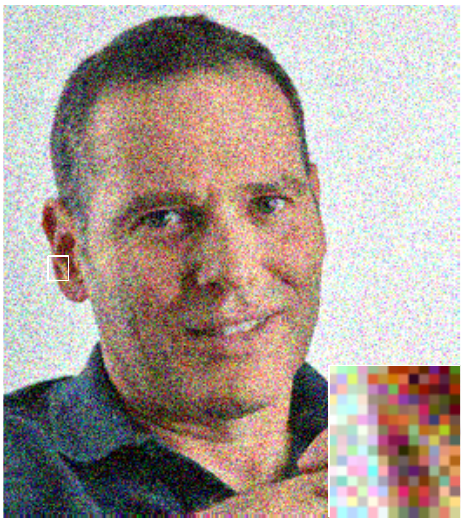}&
			\includegraphics[width=0.17\textwidth]{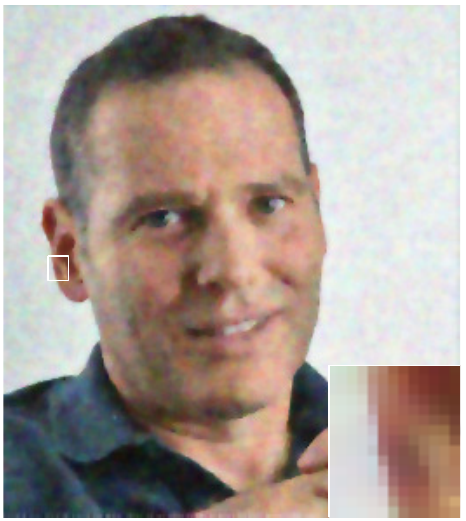}&
			\includegraphics[width=0.17\textwidth]{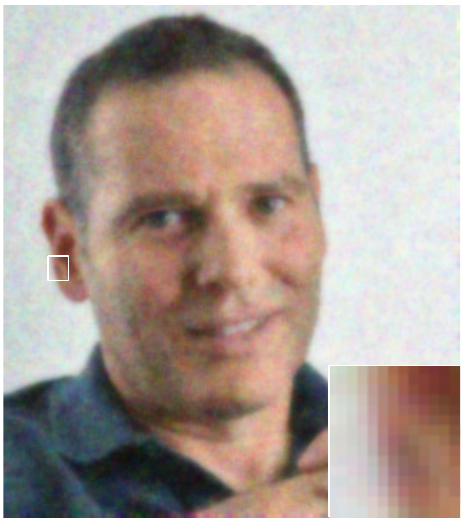}&
			\includegraphics[width=0.17\textwidth]{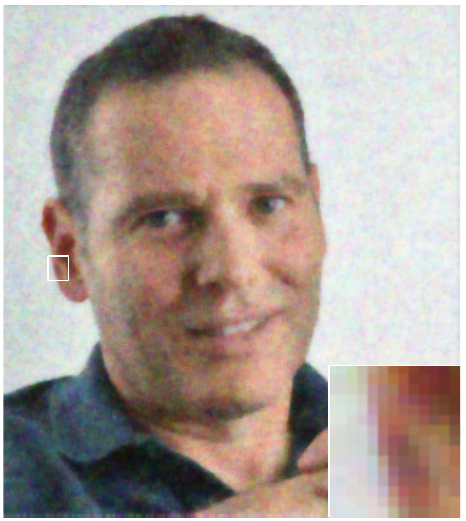}&
			\includegraphics[width=0.17\textwidth]{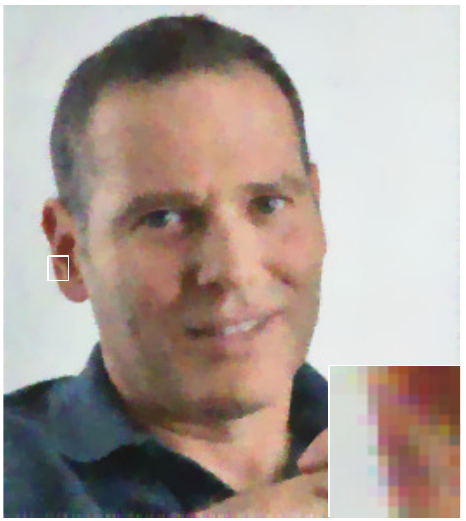}\\
			\includegraphics[width=0.17\textwidth]{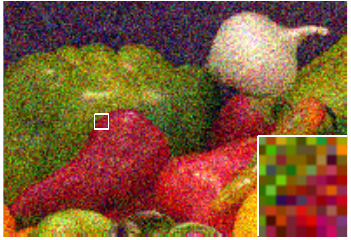}&
			\includegraphics[width=0.17\textwidth]{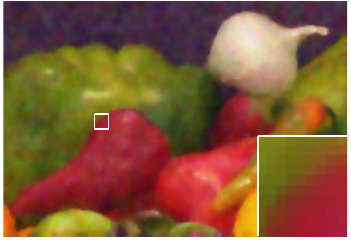}&
			\includegraphics[width=0.17\textwidth]{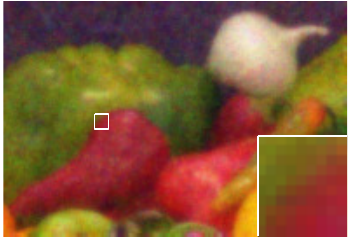}&
			\includegraphics[width=0.17\textwidth]{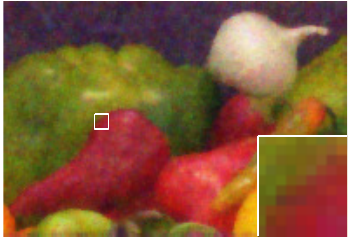}&
			\includegraphics[width=0.17\textwidth]{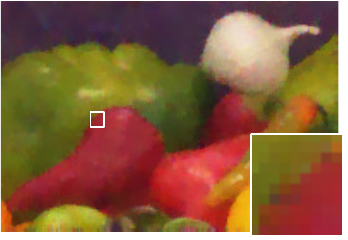}
		\end{tabular}
		\caption{(Gaussian noise with $SD=0.15$.) Comparison of the denoised images by the proposed method and the model of the Polyakov action, CTV, and VTV. First column: noisy images. Second column: results by the proposed method with $\alpha=0.01, \eta=1, \beta=0.02$. Third column: results the Polyakov action model. Fourth column: results by CTV with $\lambda=3$. Fifth column: results by VTV with $\lambda=0.2$.}
		\label{fig.Gaussian.comparison}
	\end{figure}
	\subsection{Image denoising for Gaussian noise}\label{sec.numerical.Gaussian}
	We use the proposed model to denoise images with  Gaussian noise.
	We add Gaussian noise with $SD=0.03$ to three images: (a)  portrait, (b)  orange ball and (c) chips (shown in the first row of Figure \ref{fig.Gaussian.light}).
	The noisy images  and denoised images with $\alpha=0.01,\eta=0.5$ are shown in the second and third rows of Figure \ref{fig.Gaussian.light}, respectively.
	$\beta=0.01$ is used for the orange ball, and $\beta=0.005$ is used for the portrait and chips.
	For all of  three examples, while they are denoised, sharp edges are kept.
	To demonstrate the efficiency of the proposed method, in Figure \ref{fig.Gaussian.light.ener}, we present the evolution of the energy and $\|u^{n+1}-u^n\|_2$ of the three examples shown in Figure \ref{fig.Gaussian.light}.
	For  the three examples, the energy decreases very fast and achieves the minimum within 40 iterations.
	Linear convergence is observed on the evolution of $\|u^{n+1}-u^n\|_2$.
	
	We next add Gaussian noise with $SD=0.06$ to these images.
	The noisy and denoised images are shown in Figure \ref{fig.Gaussian.heavy}.
	The evolution of the energy w.r.t. the number of iterations are also shown in the third row.
	Our method is efficient and performs well.
	The energy achieves its minimum within 70 iterations.
	To better demonstrate the smoothing effect, we select the zoomed region of the orange ball and show the surface plot of each channel (RGB channel correspond the red, green and blue surface) in Figure \ref{fig.Gaussian.ball.surf}.
	The surface plot of the noisy image is shown in the first row, which is very oscillating.
	The surface plot of the denoised image is shown in the second row.
	The surfaces of all three channels are very smooth which verifies the smoothing property of the proposed model.

		We then compare the proposed model with the Polyakov action model \cite{kimmel1997high}, the color total variation (CTV) model \cite{blomgren1998color} and the vectorial total variation (VTV) model \cite{goldluecke2012natural}. For Gaussian noise with $SD=0.06$, the denoised images by these models are shown in Figure \ref{fig.Gaussian06.comparison}. In the proposed model, $\alpha=0.01, \eta=1, \beta=0.005$ is used. In CTV and VTV, we set $\lambda=6$ and $\lambda=0.1$ respectively. Compared to other models, the proposed model smoothens flat regions while keeping sharp edges and textures. To quantify the difference of the denoised images, we compute and compare the PSNR and the SSIM \cite{wang2004image} value of each denoised image in Figure \ref{fig.Gaussian06.comparison} and present these values in Table \ref{tab.Gaussian.comparison}(a). The proposed model and VTV outperform the other two models in both values. For the image of Chips, the result by VTV smoothens out almost all textures of the background. In comparison, these textures are kept by our proposed model. The number of iterations and CPU time of different methods are reported in Table \ref{tab.Gaussian.comparison}(b). Due to the complex structures of the color elastica model, the proposed method is slower than the other three methods.
	
	To further demonstrate the effectiveness of the proposed model, we compare these models on images with heavy Gaussian noise. In Figure \ref{fig.Gaussian.comparison}, the noisy images contaminated by Gaussian noise with $SD=0.15$ are shown in the first column. The results by the proposed method, the Polyakov action, CTV and VTV are shown in Column 2 - Column 5, respectively. In the proposed model, $\alpha=0.01, \eta=1, \beta=0.02$ is used. In CTV and VTV, we set $\lambda=3$ and $\lambda=0.2$ respectively. The PSNR and the SSIM value of each denoised  image are shown in Table \ref{tab.Gaussian.comparison}(a). The proposed model and VTV outperform the other two models in both values. While VTV is powerful in smoothing the flat region of images, the edges in the denoised image are oscillating. Compared to VTV, the proposed model provides results with smooth and sharp edges, and with larger PSNR values. This justifies the effectiveness of the high-order derivatives in the proposed model. The number of iterations and CPU time of different methods are reported in Talbe \ref{tab.Gaussian.comparison}(b). 

	\begin{figure}[t!]
		\centering
		\begin{tabular}{ccc}
			(a) &(b) &(c)\\
			\includegraphics[height=0.28\textwidth]{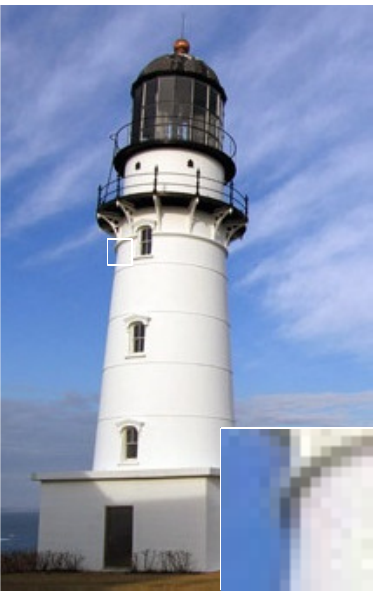}&
			\includegraphics[height=0.28\textwidth]{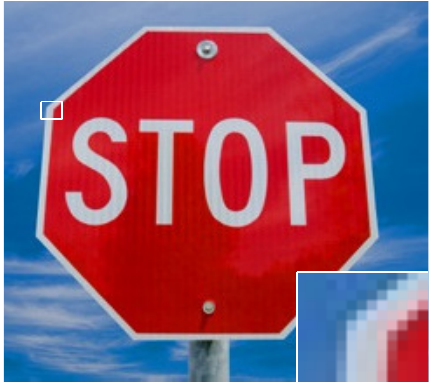}&
			\includegraphics[height=0.22\textwidth]{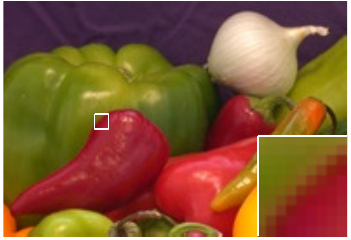}\\
			\includegraphics[height=0.28\textwidth]{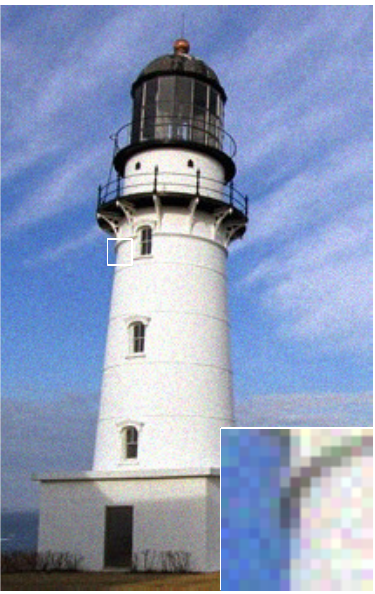}&
			\includegraphics[height=0.28\textwidth]{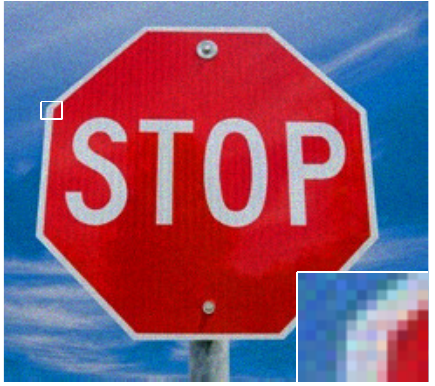}&
			\includegraphics[height=0.22\textwidth]{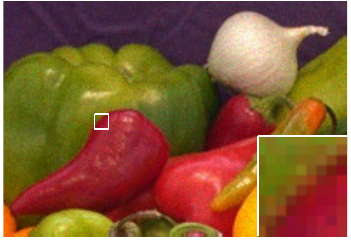}\\
			\includegraphics[height=0.28\textwidth]{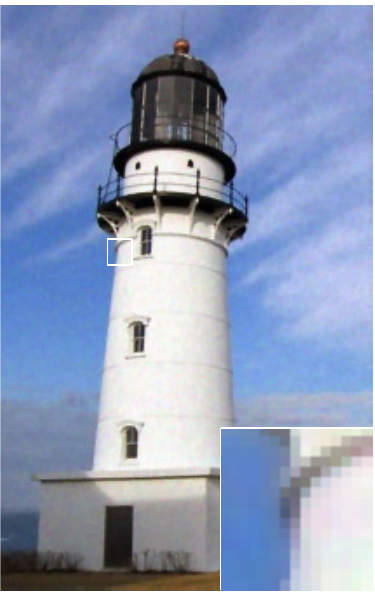}&
			\includegraphics[height=0.28\textwidth]{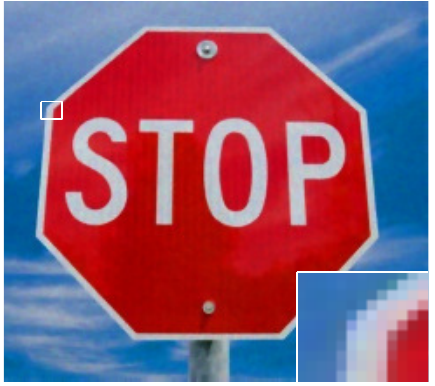}&
			\includegraphics[height=0.22\textwidth]{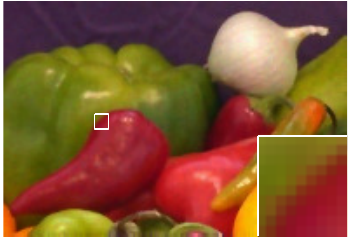}
		\end{tabular}
		\caption{(Possion noise with $P=500$. $\alpha=0.01,\eta=0.3,\beta=0.005$.) Image denoising by the proposed method on (a) the lighthouse, (b) the stop sign and (c) vegetables . The first row shows clean images. The second row shows noisy images. The third row shows denoised images.}
		\label{fig.Poisson.light}
	\end{figure}
	\begin{figure}[t!]
		\centering
		\begin{tabular}{ccc}
			(a) &(b) &(c)\\
			\includegraphics[width=0.3\textwidth]{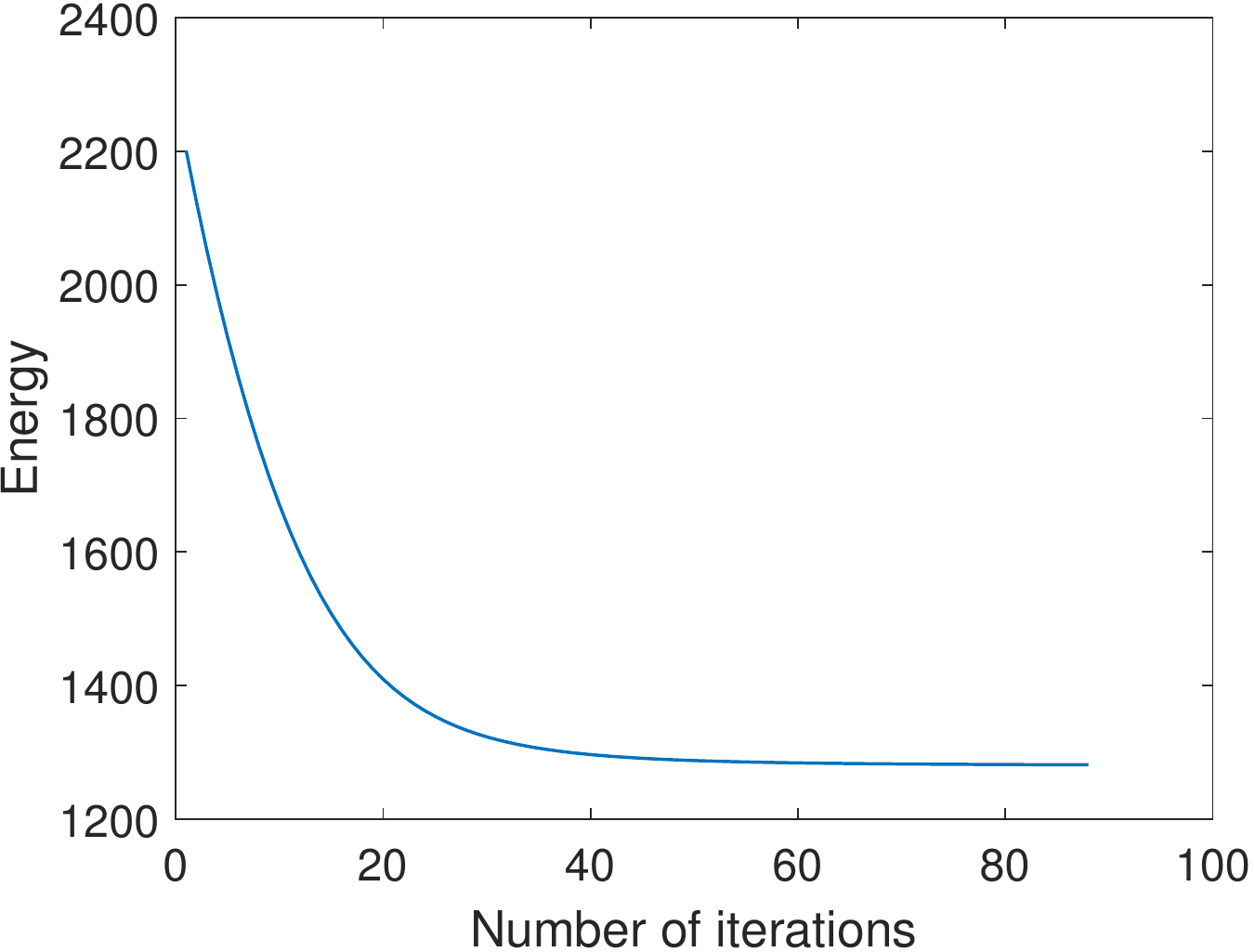}&
			\includegraphics[width=0.3\textwidth]{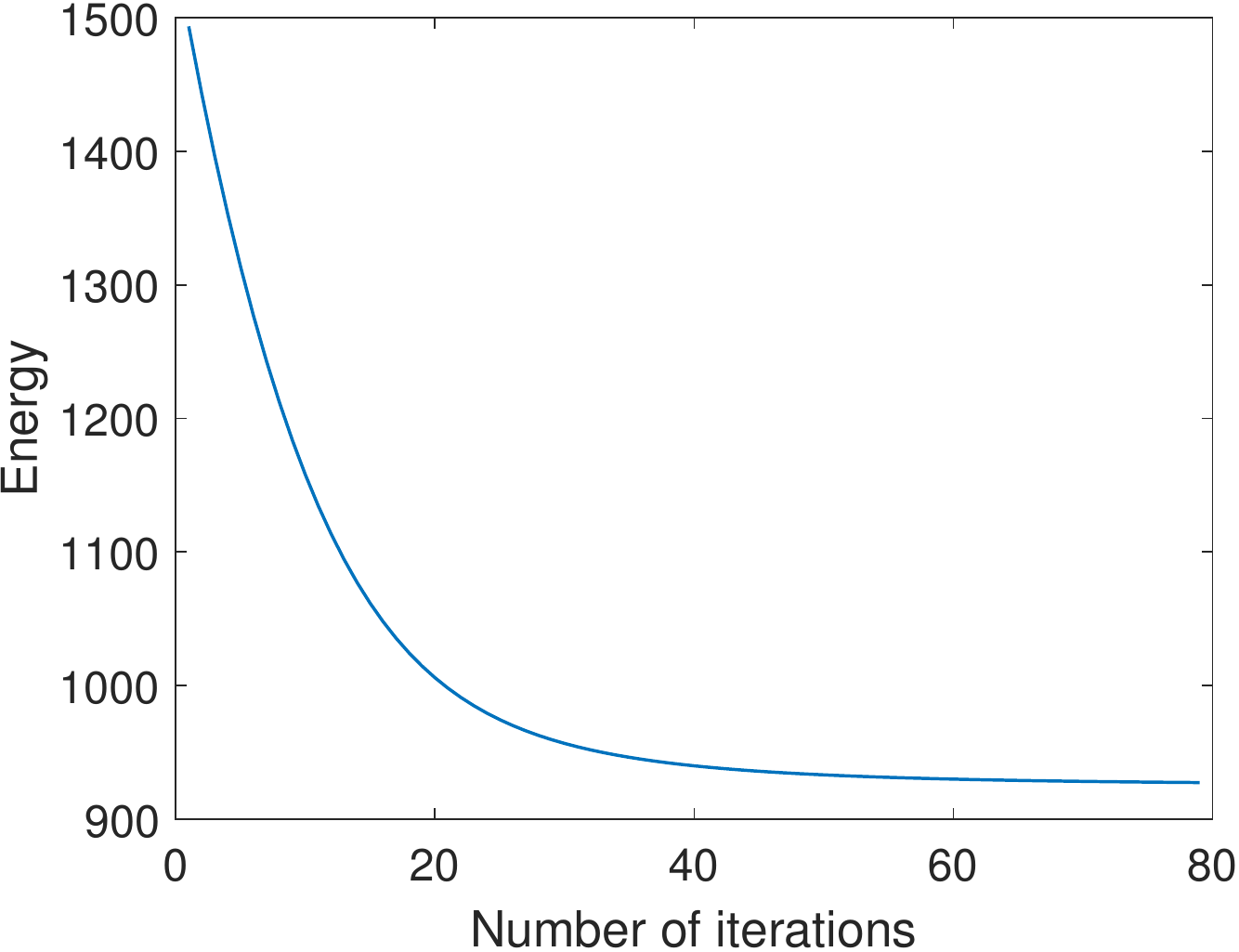}&
			\includegraphics[width=0.3\textwidth]{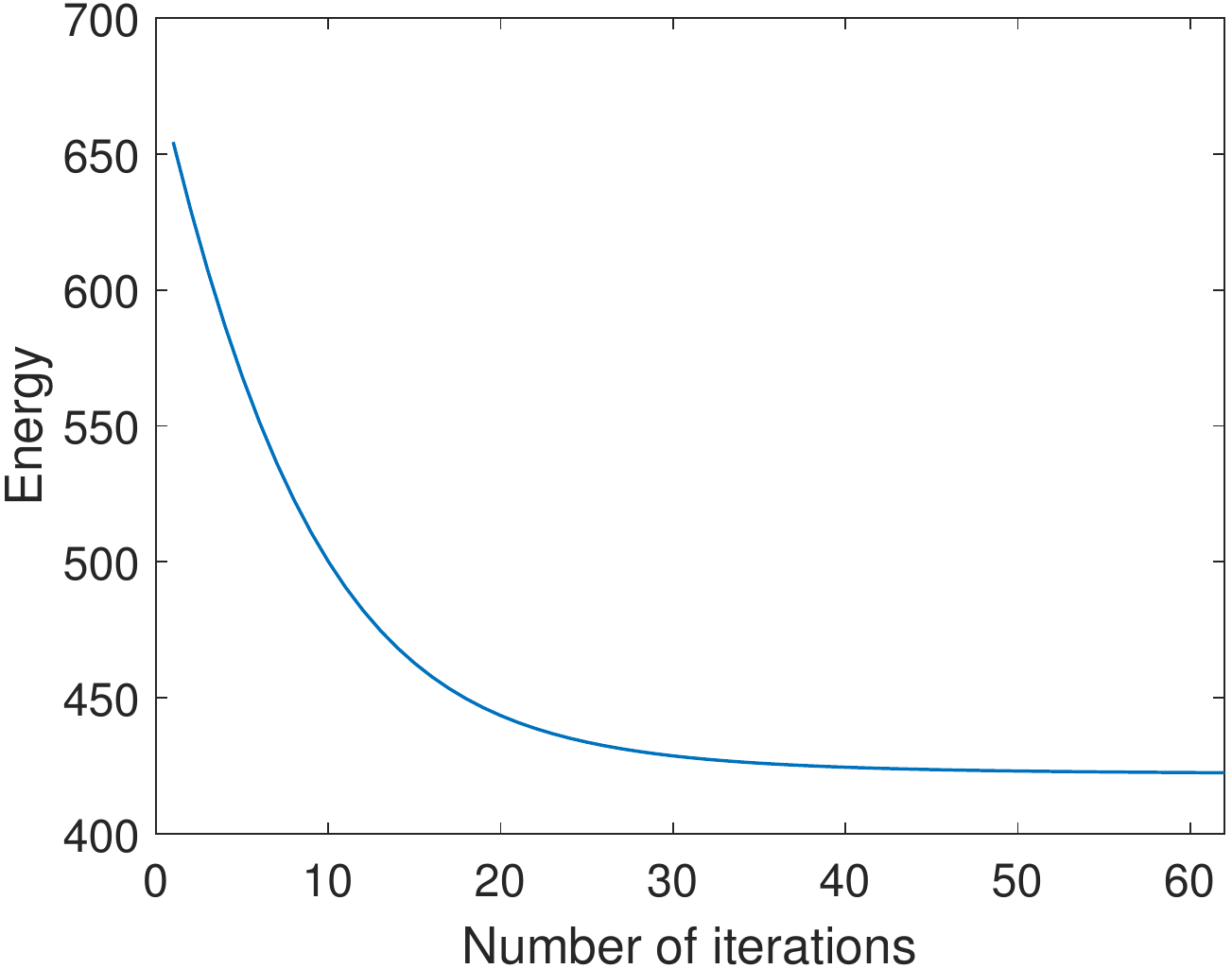}\\
			\includegraphics[width=0.3\textwidth]{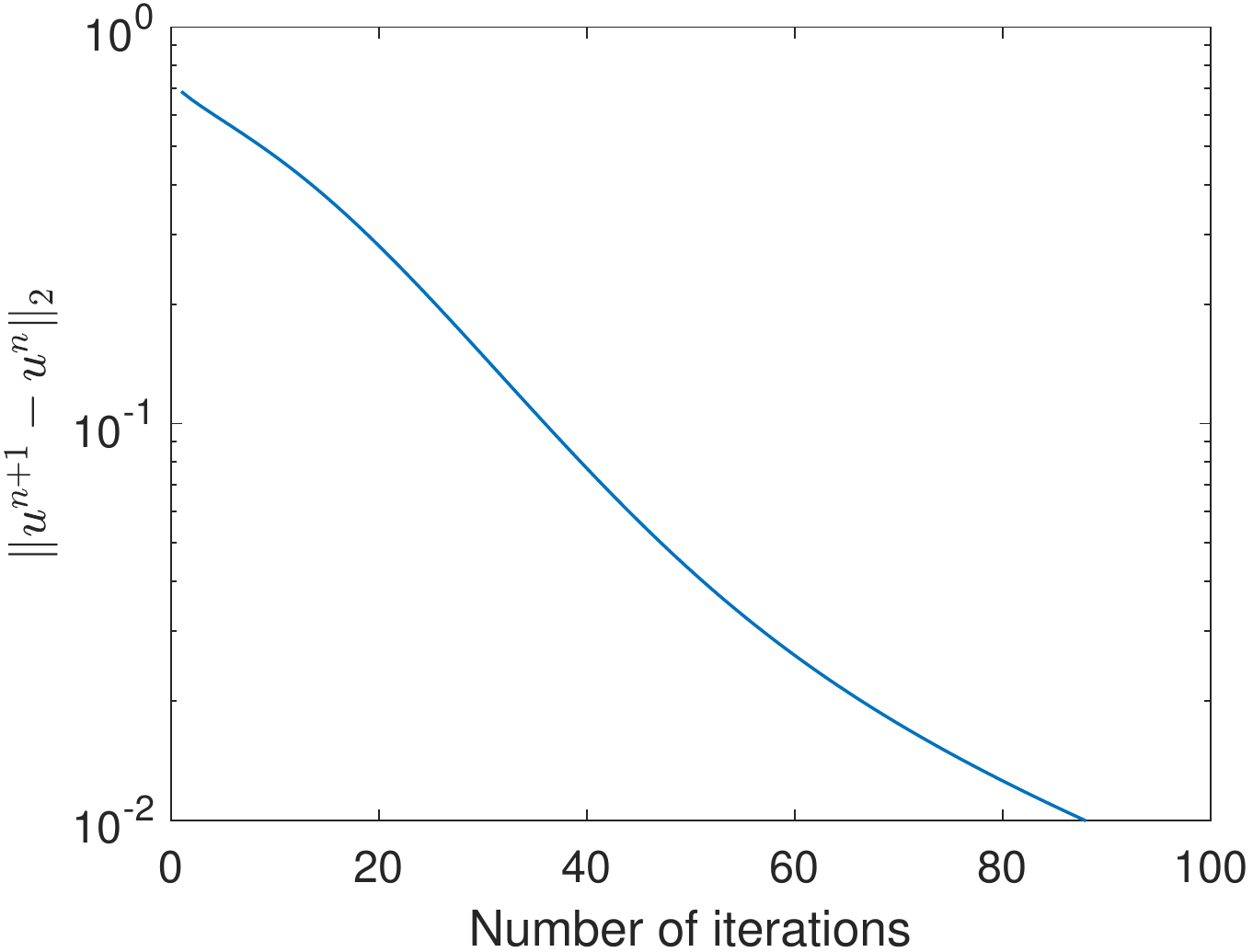}&
			\includegraphics[width=0.3\textwidth]{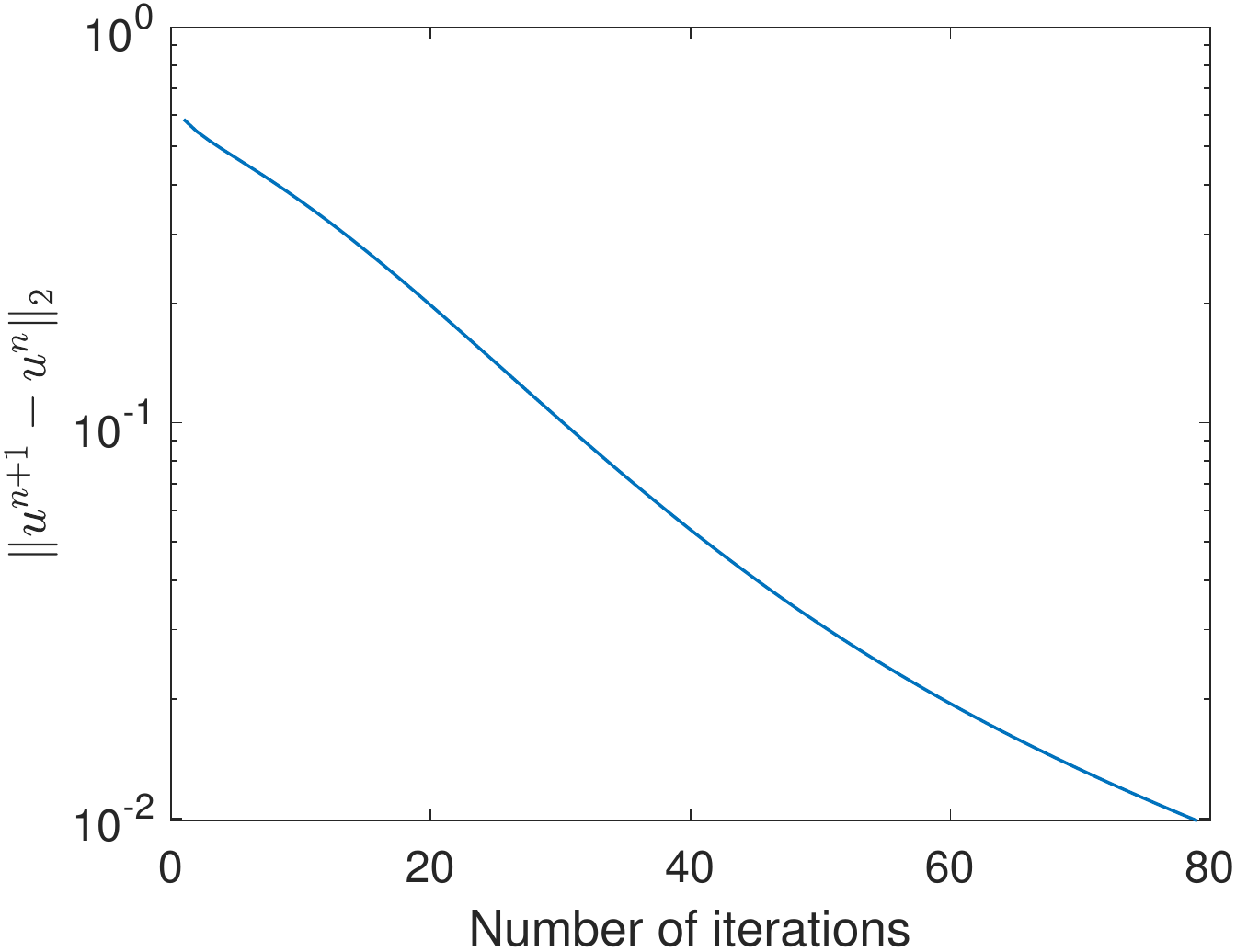}&
			\includegraphics[width=0.3\textwidth]{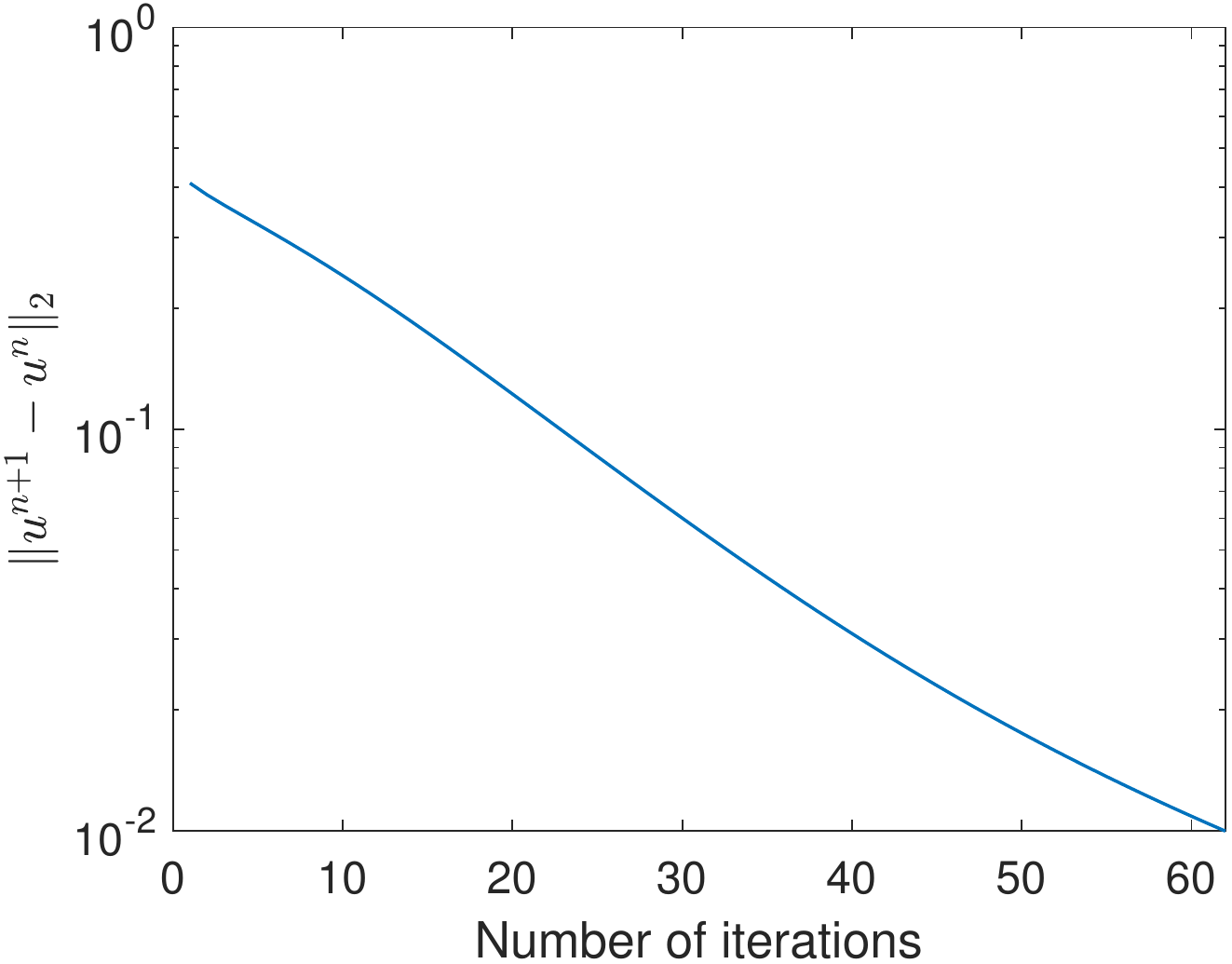}
		\end{tabular}
		\caption{(Possion noise with $P=500$. $\alpha=0.01,\eta=0.3,\beta=0.005$.) The evolution of (first row) the energy and (second row) $\|u^{n+1}-u^n\|_2$ w.r.t. the number of iterations for results in Figure \ref{fig.Poisson.light}. (a) the lighthouse. (b) the stop sign and (c) vegetables.}
		\label{fig.Poisson.light.ener}
	\end{figure}
	
	\begin{figure}[t!]
		\centering
		\begin{tabular}{ccc}
			(a) &(b) &(c)\\
			\includegraphics[height=0.28\textwidth]{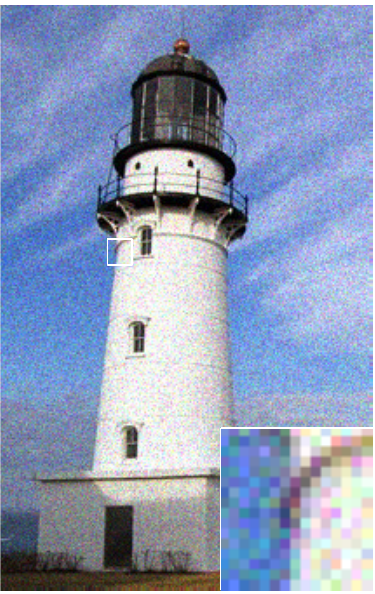}&
			\includegraphics[height=0.28\textwidth]{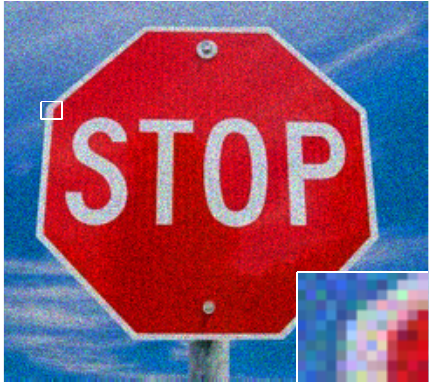}&
			\includegraphics[height=0.22\textwidth]{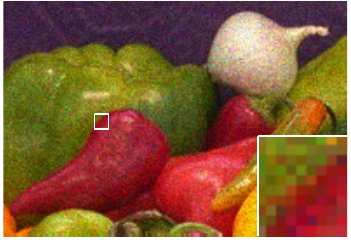}\\
			\includegraphics[height=0.28\textwidth]{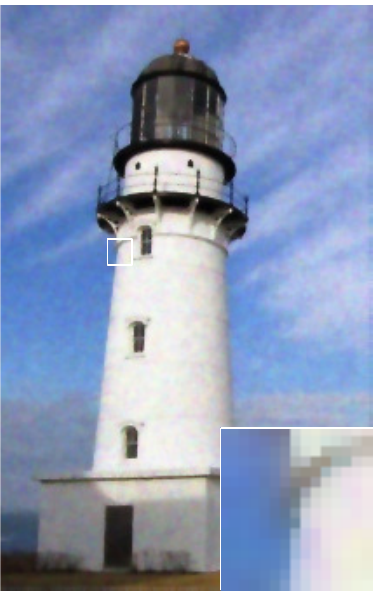}&
			\includegraphics[height=0.28\textwidth]{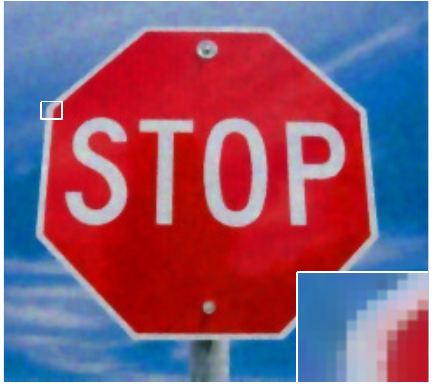}&
			\includegraphics[height=0.22\textwidth]{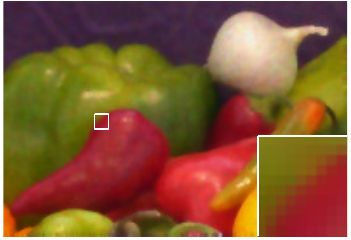}
		\end{tabular}
		\caption{(Possion noise with $P=100$. $\alpha=0.01,\eta=1,\beta=0.005$.) Image denoising by the proposed method on (a) the lighthouse, (b) the stop sign and (c) vegetables. The first row shows noisy images. The second row shows denoised images.}
		\label{fig.Poisson.heavy}
	\end{figure}
	\begin{figure}[t!]
		\centering
		\begin{tabular}{ccc}
			(a) &(b) &(c)\\
			\includegraphics[width=0.28\textwidth]{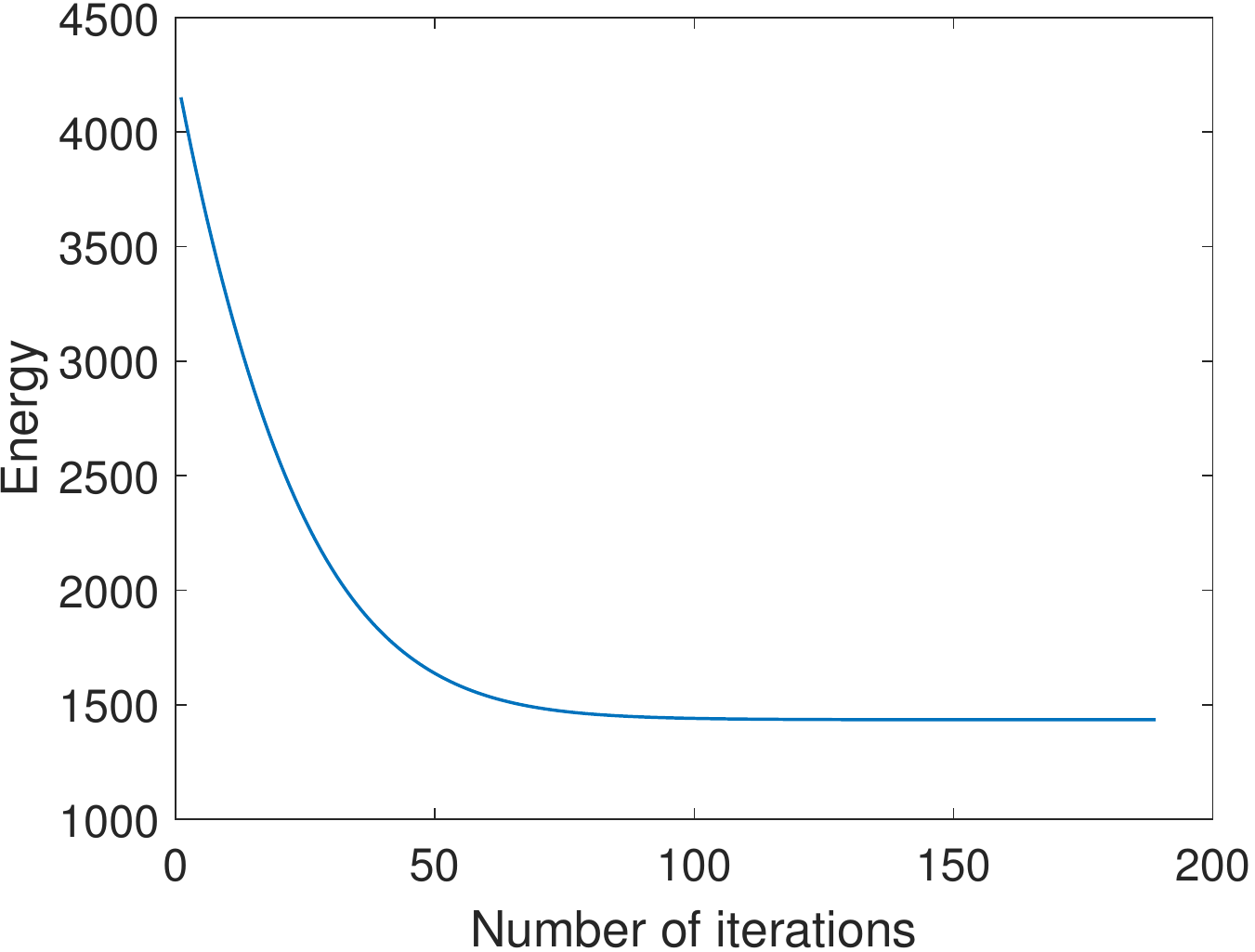}&
			\includegraphics[width=0.28\textwidth]{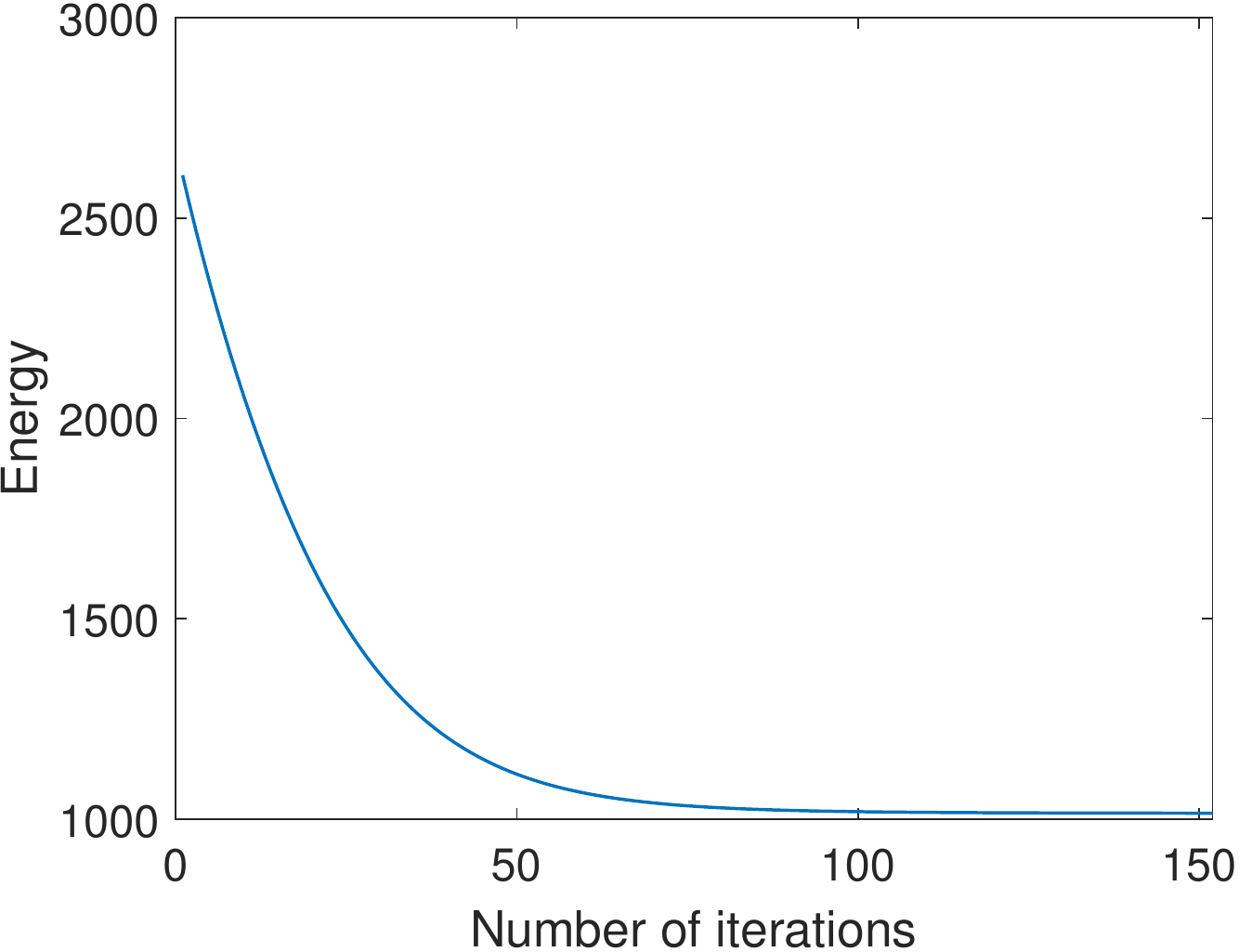}&
			\includegraphics[width=0.28\textwidth]{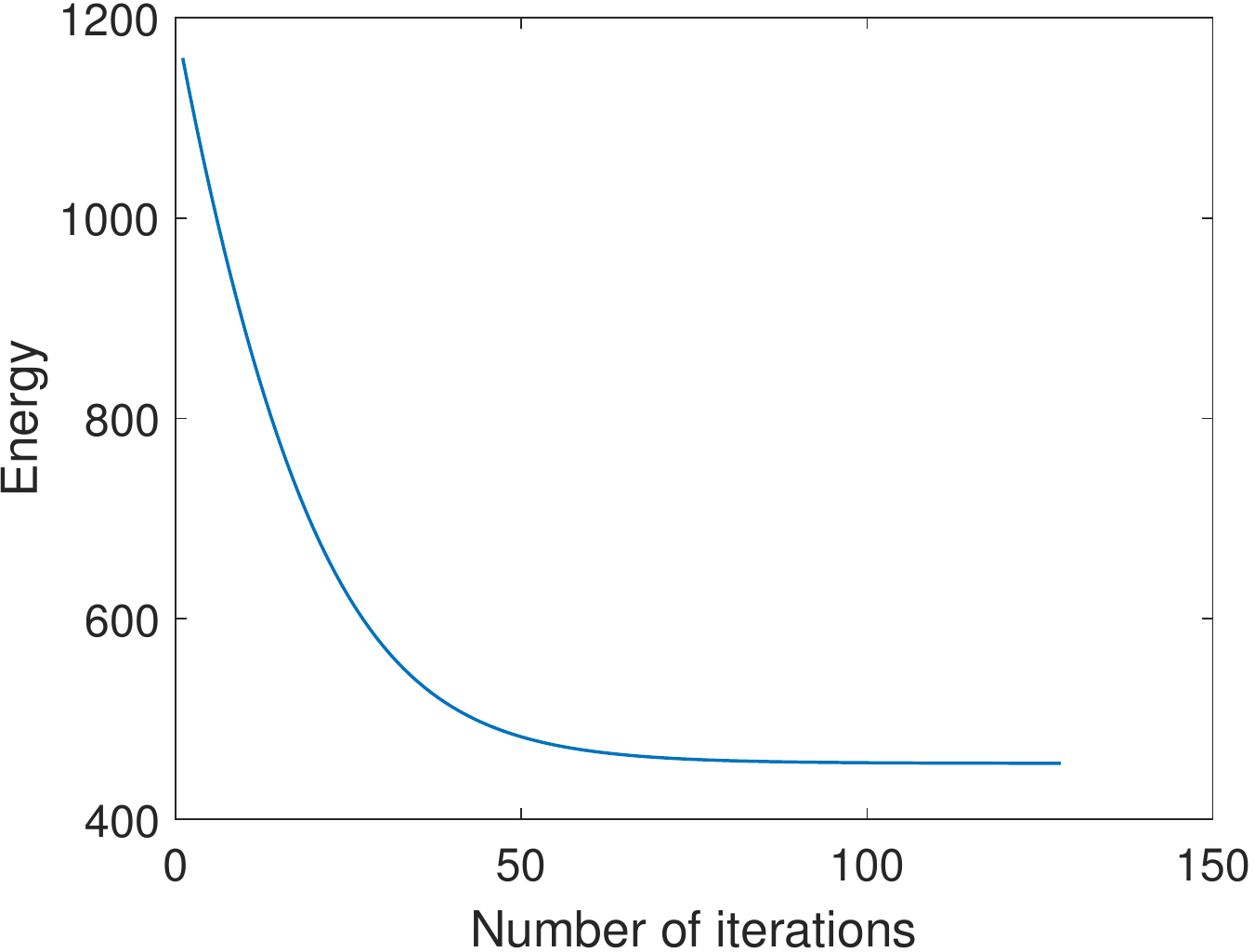}
		\end{tabular}
		\caption{(Possion noise with $P=100$. $\alpha=0.01,\eta=1,\beta=0.005$.) The evolution of the energy w.r.t. the number of iterations for results in Figure \ref{fig.Poisson.heavy}. (a) the lighthouse, (b) the stop sign and (c) vegetables.}
		\label{fig.Poisson.heavy.ener}
	\end{figure}
	\begin{figure}[t!]
		\centering
		\begin{tabular}{cc}
			\includegraphics[height=0.28\textwidth]{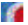}&
			\includegraphics[height=0.28\textwidth]{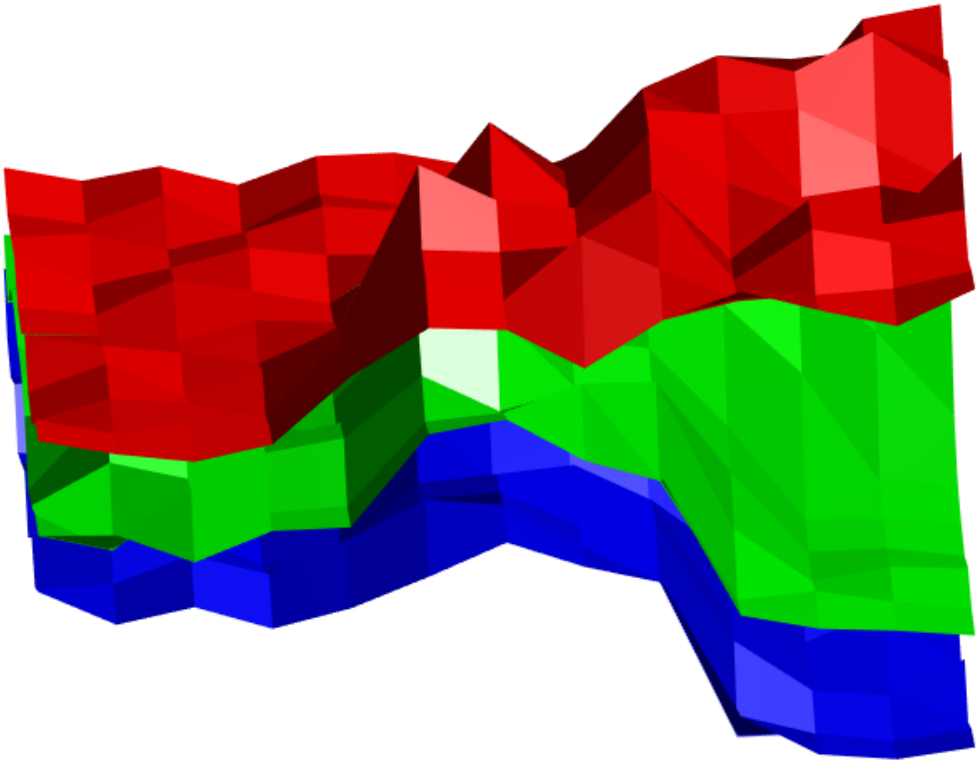}\\
			\includegraphics[height=0.28\textwidth]{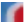}&
			\includegraphics[height=0.28\textwidth]{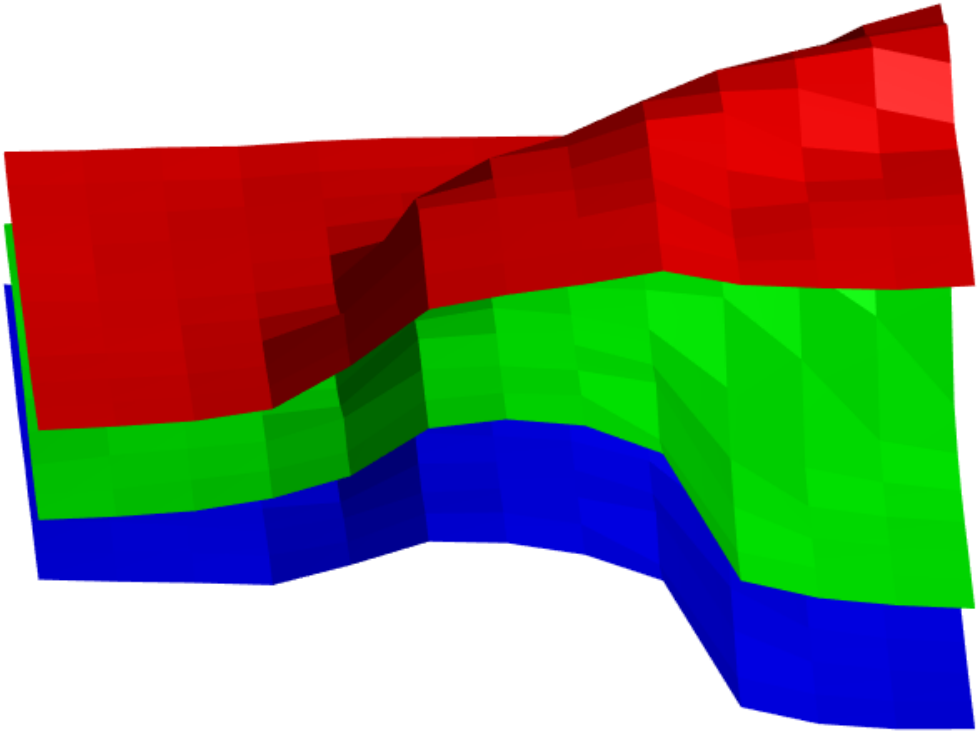}
		\end{tabular}
		\caption{(Poisson noise with $P=100$. Smoothing effect.) Left: zoomed region of the (first row) noisy and (second row) denoised stop sign in Figure \ref{fig.Poisson.heavy}. Right: the surface plot of the left images. The RGB channels correspond to the red, green and blue surface, respectively.}
		\label{fig.Poisson.stop.surf}
	\end{figure}
	
	\begin{table}[t!]
		\centering
		(a)\\
		\begin{tabular}{c|cc|cc|cc|cc}
			\hline
			& \multicolumn{2}{c|}{Proposed} & \multicolumn{2}{c|}{Polyakov} & \multicolumn{2}{c|}{CTV} & \multicolumn{2}{c}{VTV}\\
			\hline
			Stop Sign & 32.05 & (0.99) &28.52&(0.98) &30.36 & (0.98) & 31.73 & (0.97)\\
			\hline
			Vegetables & 31.25 &(0.98) & 30.04 & (0.98) & 30.60 &(0.97) &  31.04 &(0.98)\\
			\hline
		\end{tabular}\vspace{0.2cm}\\
		(b)\\
		\begin{tabular}{c|cc|cc|cc|cc}
			\hline
			& \multicolumn{2}{c|}{Proposed} & \multicolumn{2}{c|}{Polyakov} & \multicolumn{2}{c|}{CTV} & \multicolumn{2}{c}{VTV}\\
			\hline
			Stop Sign& 152 & (52.47) & 73&(2.03) &103 & (2.05) & 307 & (7.94)\\
			\hline
			Vegetables & 128 &(18.86) & 85 & (0.94) & 111 &(0.67) &  243 &(2.79)\\
			\hline
		\end{tabular}
		\caption{(Poisson noise with $P=100$. Comparison of different methods.) (a) The PSNR (SSIM) value of the denoised images; (b) The number of iterations (CPU time in seconds) of different methods. }
		\label{tab.Poisson100.comparison}
	\end{table}
	
	\subsection{Image denoising for Poisson noise}
	\label{sec.numerical.Poisson}
	We explore the performance of the proposed method to denoise Poisson noise with $P=500$.
	Our examples are (a) the lighthouse, (b)  stop sign and (c) vegetables, see the first row in Figure \ref{fig.Poisson.light}.
	The noisy images and denoised images with $\alpha=0.01,\eta=0.3,\beta=0.005$ are shown as the second row and third row in Figure \ref{fig.Poisson.light}, respectively.
	Similar to the performance on Gaussian noise, our method keeps sharp edges.
	The evolution of the energy and $\|u^{n+1}-u^n\|_2$ of these examples are shown in Figure \ref{fig.Poisson.light.ener}.
	All energies achieve their minimum within 50 iteration.
	
	Then, we add heavy Poisson noise with $P=100$.
	The noisy and denoised images are shown in Figure \ref{fig.Poisson.heavy}. In this set of experiments, $\alpha=0.01,\eta=1,\beta=0.005$ is used.
	We also show the evolution of energy against the number of iterations in Figure \ref{fig.Poisson.heavy.ener}.
	All energies achieve their minimum in about 100 iterations.
	Figure \ref{fig.Poisson.stop.surf} shows the surface plot of the zoomed region of the stop sign in Figure \ref{fig.Poisson.heavy}.
	The surface plot of the denoised image is smooth while that of the noisy image is very oscillating.  We then compare the denoised images by the proposed model, the Polyakov action model, CTV and VTV in Table \ref{tab.Poisson100.comparison}, which presents the PSNR, SSIM, number of iterations and CPU time of each denoised image. In the proposed model, $\alpha=0.01, \eta=1, \beta=0.005$ is used. In CTV and VTV, we set $\lambda=6$ and $\lambda=0.1$ respectively. Among all methods, the proposed method provides results with the largest PSNR and SSIM value.
	
	\begin{figure}[t!]
		\centering
		\begin{tabular}{ccc}
			(a) & (b) & (c)\\
			\includegraphics[height=0.28\textwidth]{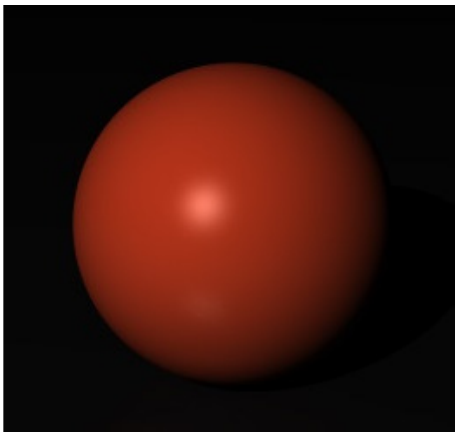}&
			\includegraphics[height=0.28\textwidth]{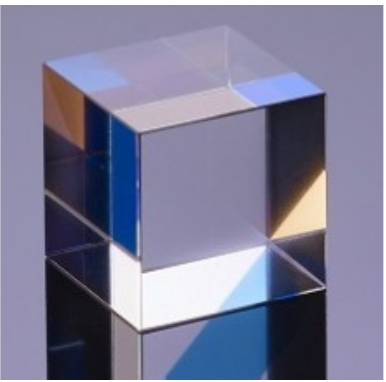}&
			\includegraphics[height=0.28\textwidth]{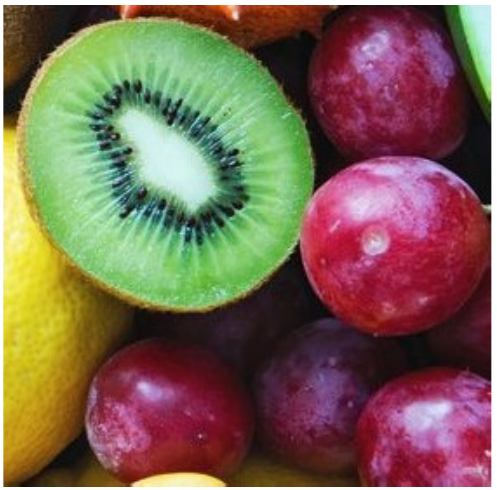}
		\end{tabular}
		\caption{(Effect of $\alpha$.): (a) The orange ball. (b) The crystal cube. (c) Fruits.}\label{fig.alpha.clean}
	\end{figure}
	
	\begin{figure}[t!]
		\centering
		\begin{tabular}{ccc}
			(a) &(b) &(c)\\
			\includegraphics[width=0.3\textwidth]{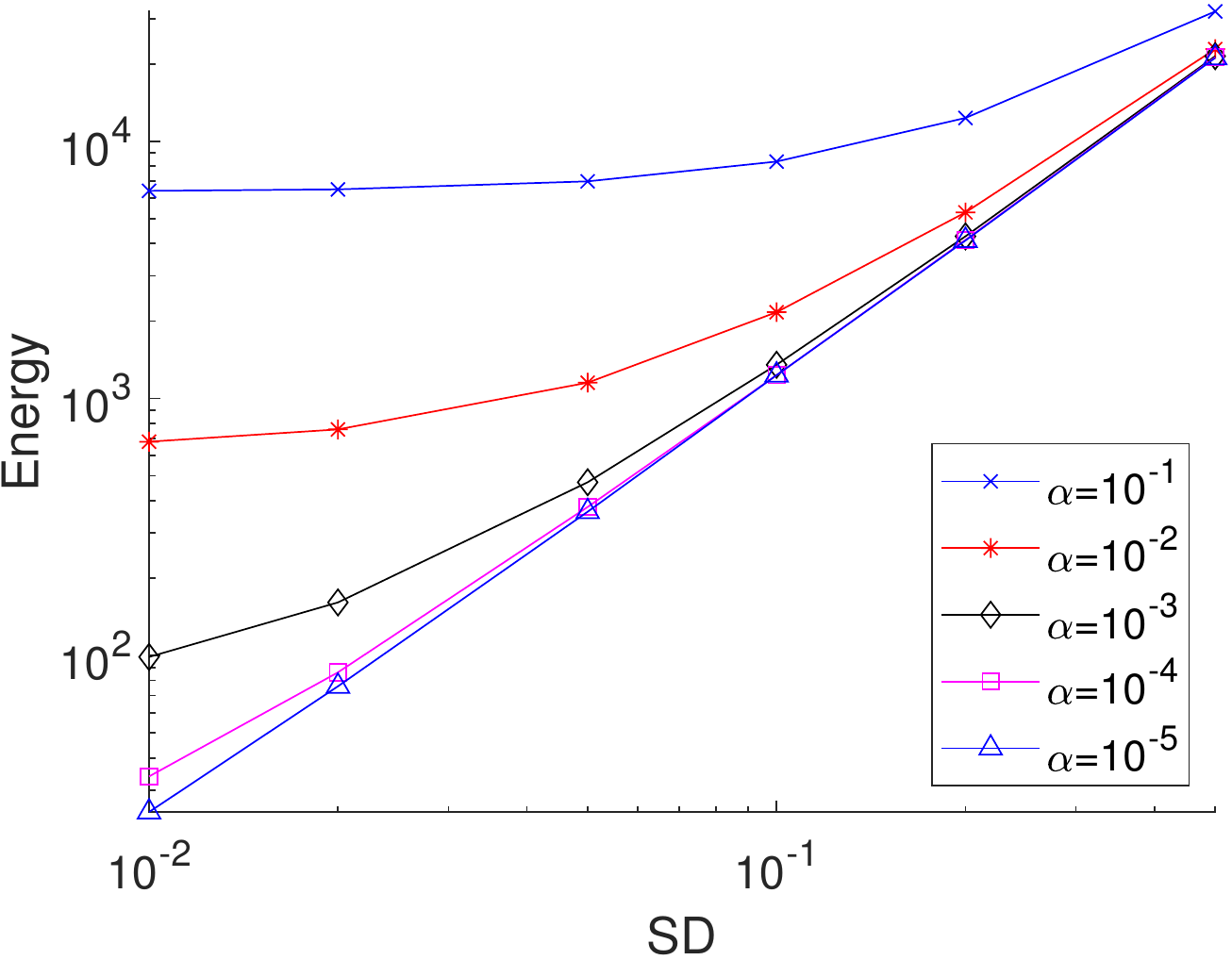}&
			\includegraphics[width=0.3\textwidth]{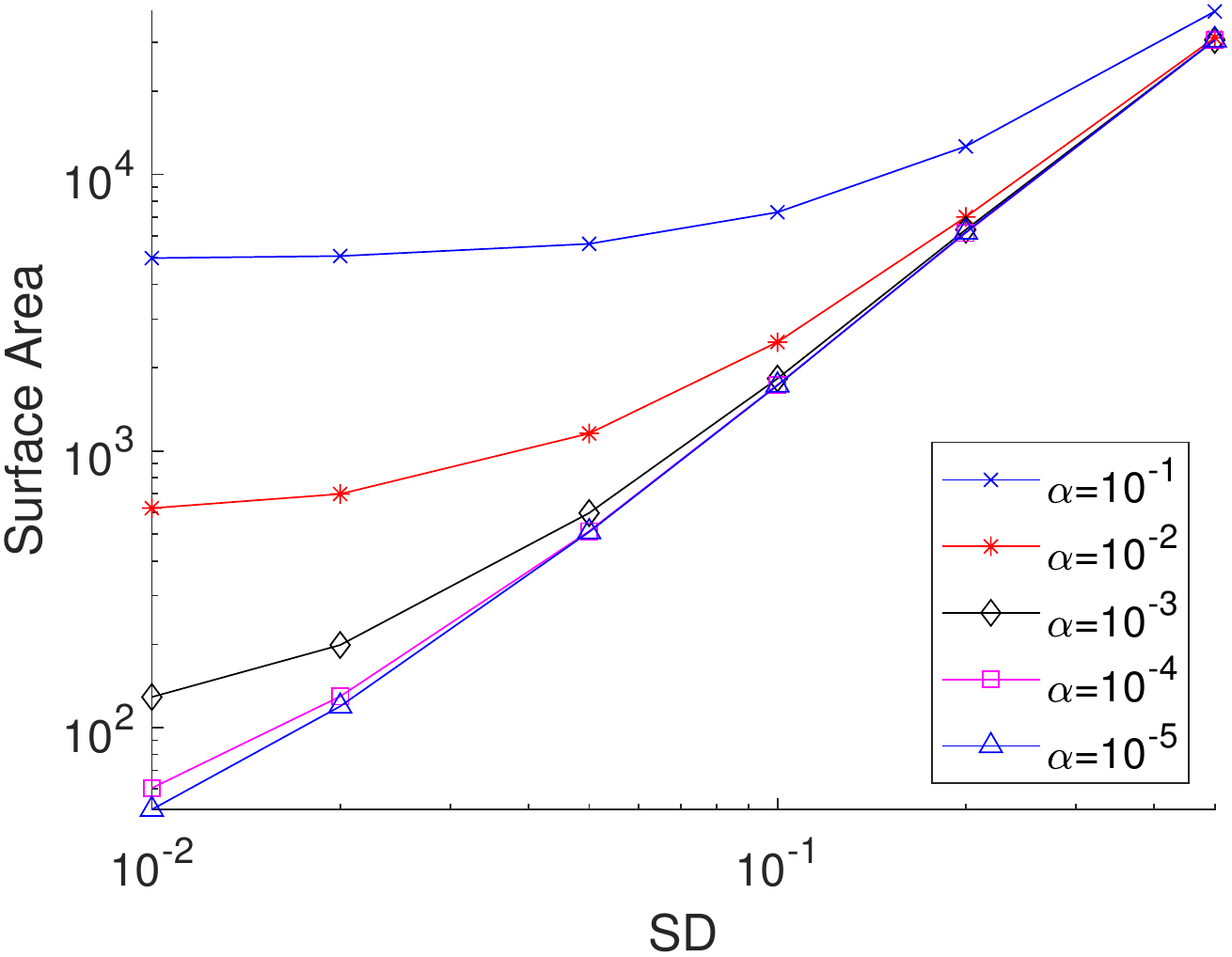}&
			\includegraphics[width=0.3\textwidth]{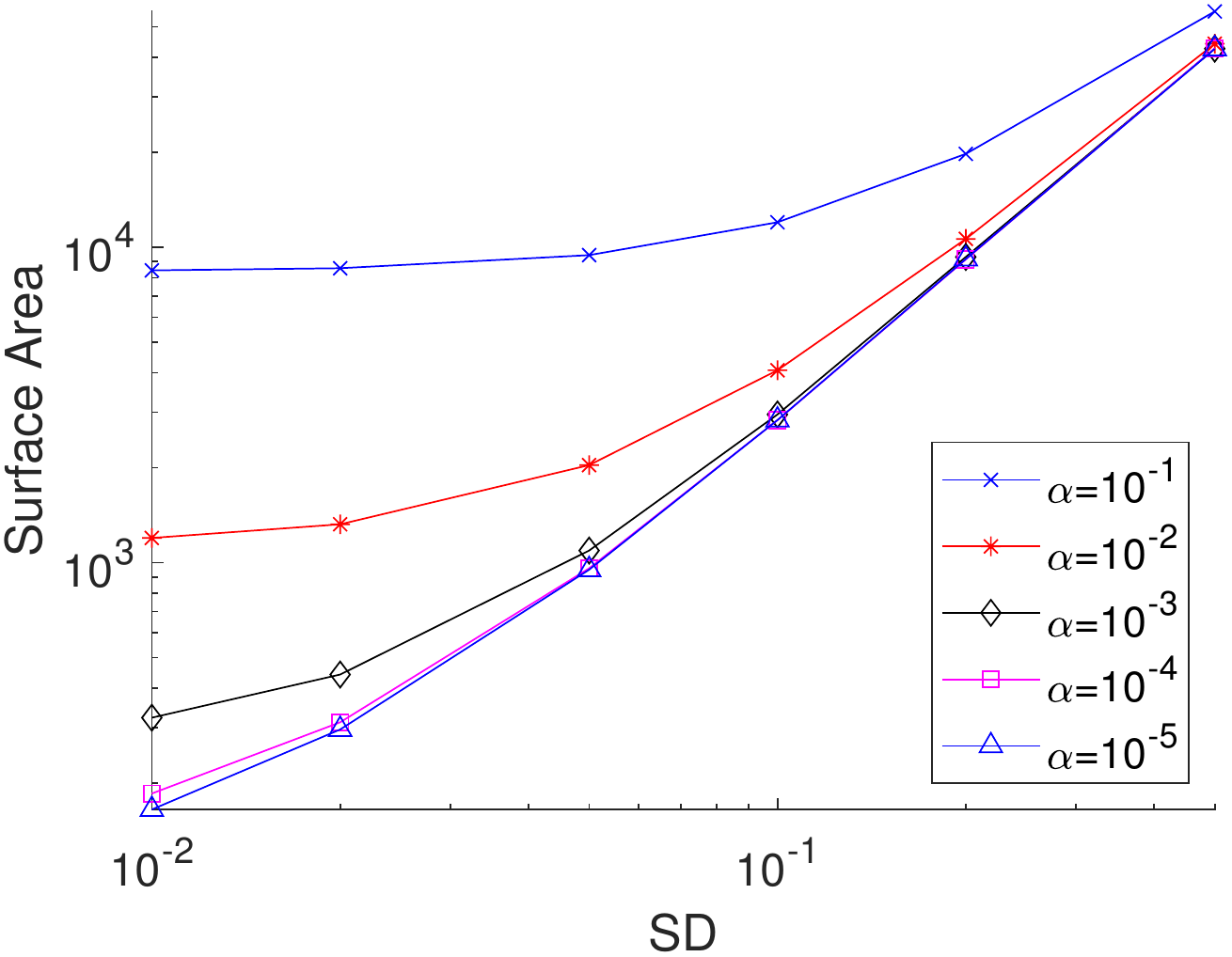}\\
			\includegraphics[width=0.3\textwidth]{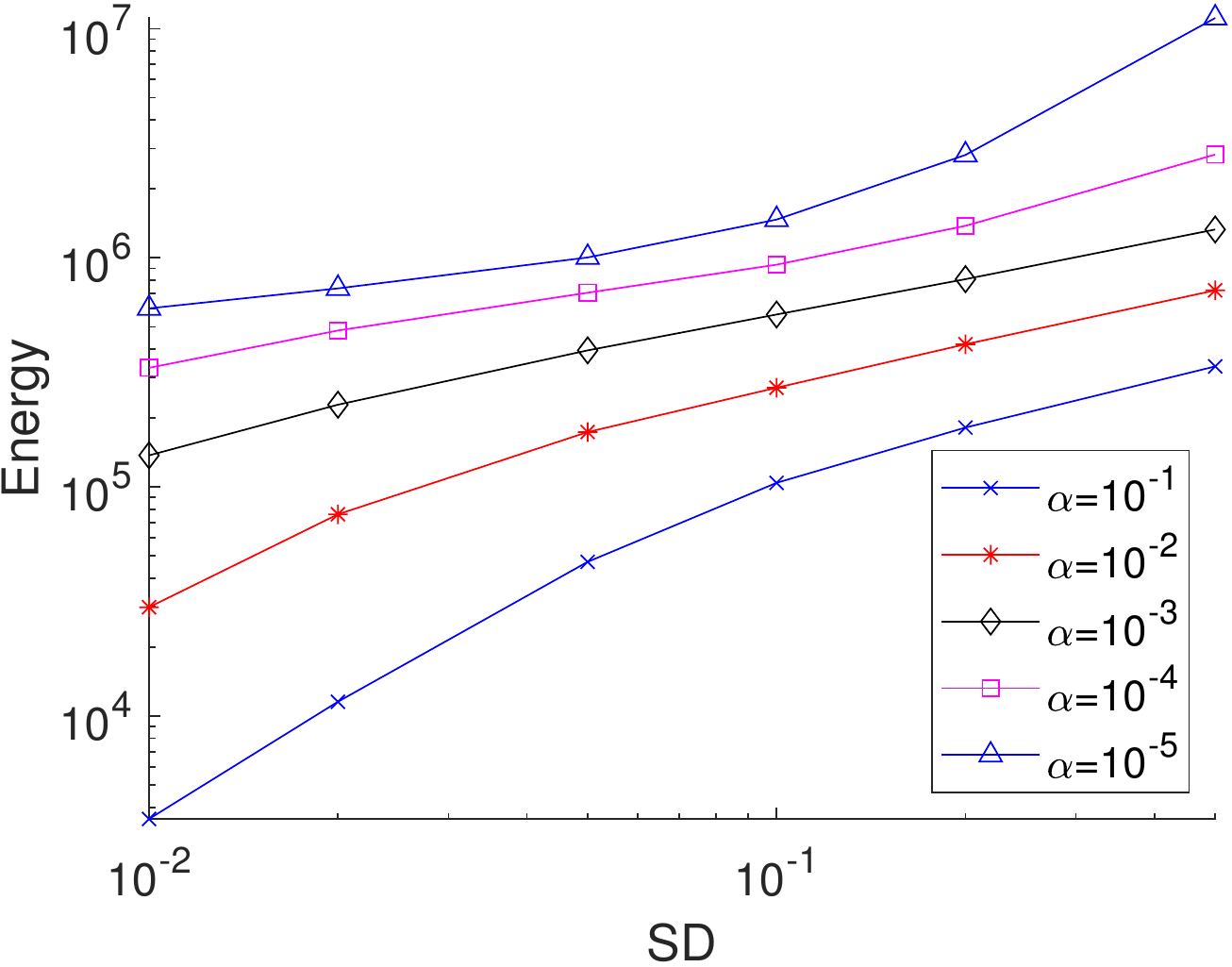}&
			\includegraphics[width=0.3\textwidth]{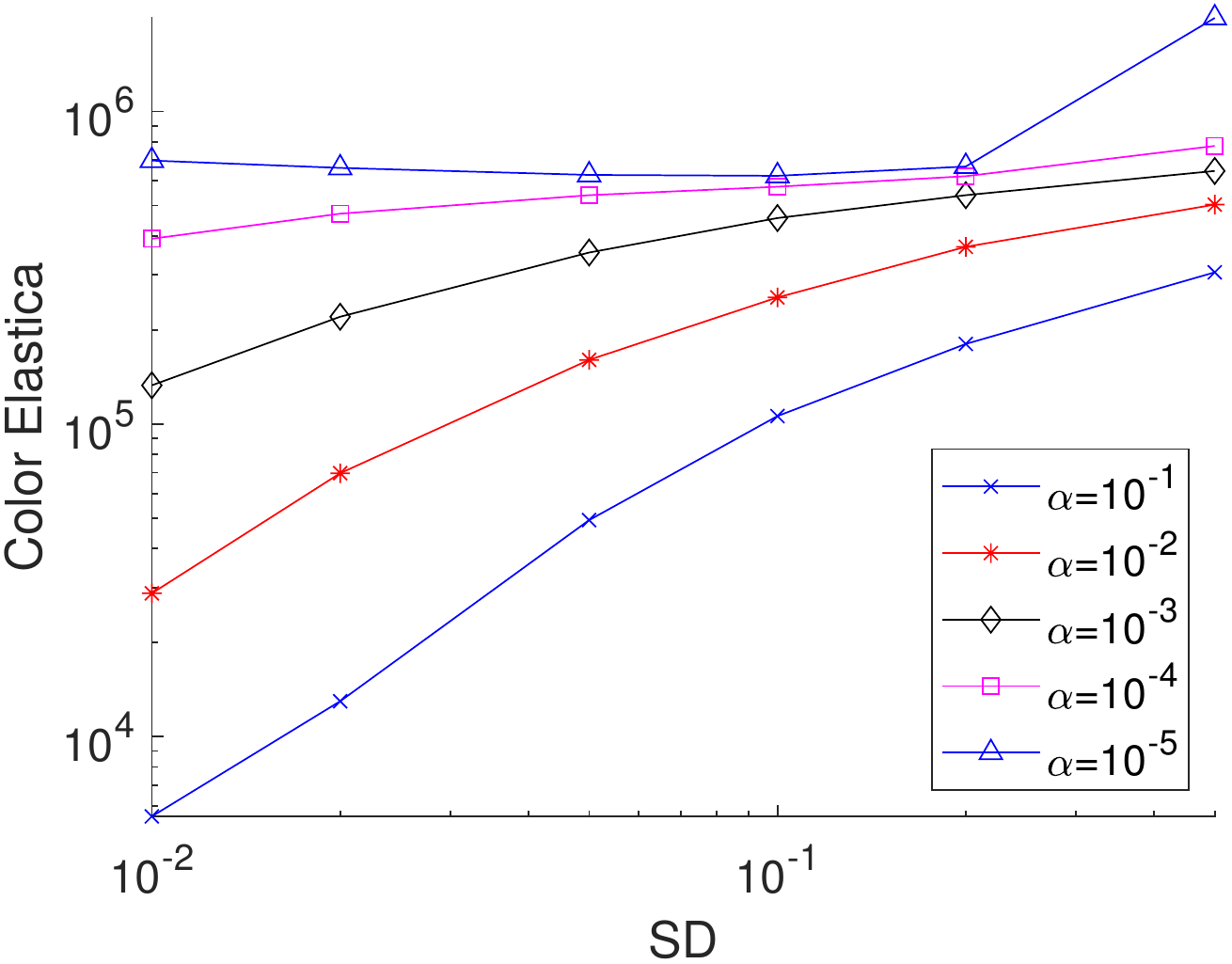}&
			\includegraphics[width=0.3\textwidth]{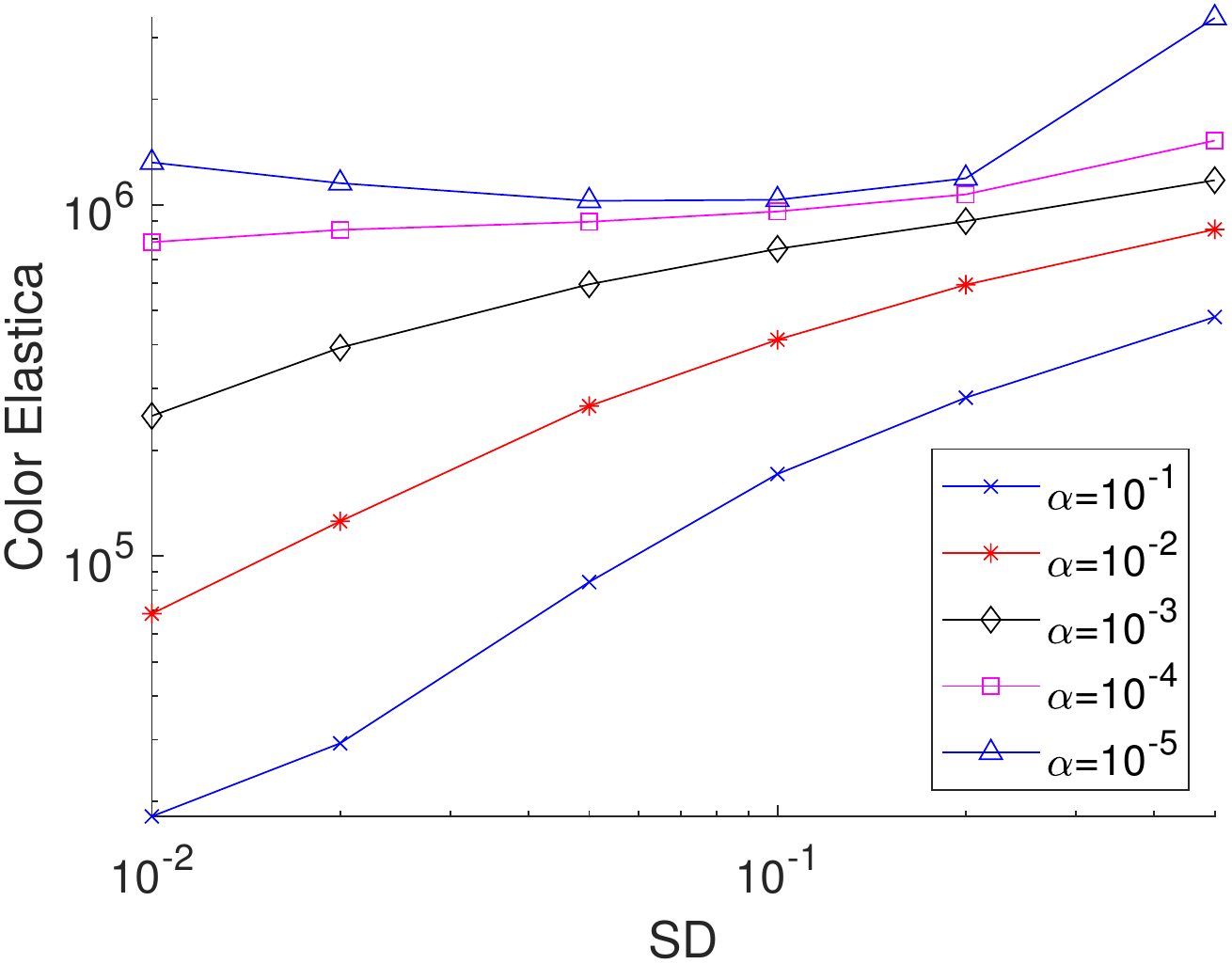}
		\end{tabular}
		\caption{(Effect of $\alpha$, Gaussian noise.) With $\alpha=10^{-1},10^{-2},10^{-3},10^{-4}$ and $10^{-5}$, plot of (first row) the surface area and (second row) the color elastica computed from noisy images w.r.t. $SD$. (a)-(c) corresponds to the orange ball, the crystal cube and fruits shown in Figure \ref{fig.alpha.clean}, respectively. Images have higher quality with smaller $SD$. 
		}
		\label{fig.alpha.G}
	\end{figure}
	
	\begin{figure}[t!]
		\centering
		\begin{tabular}{ccc}
			(a) &(b) &(c)\\
			\includegraphics[width=0.3\textwidth]{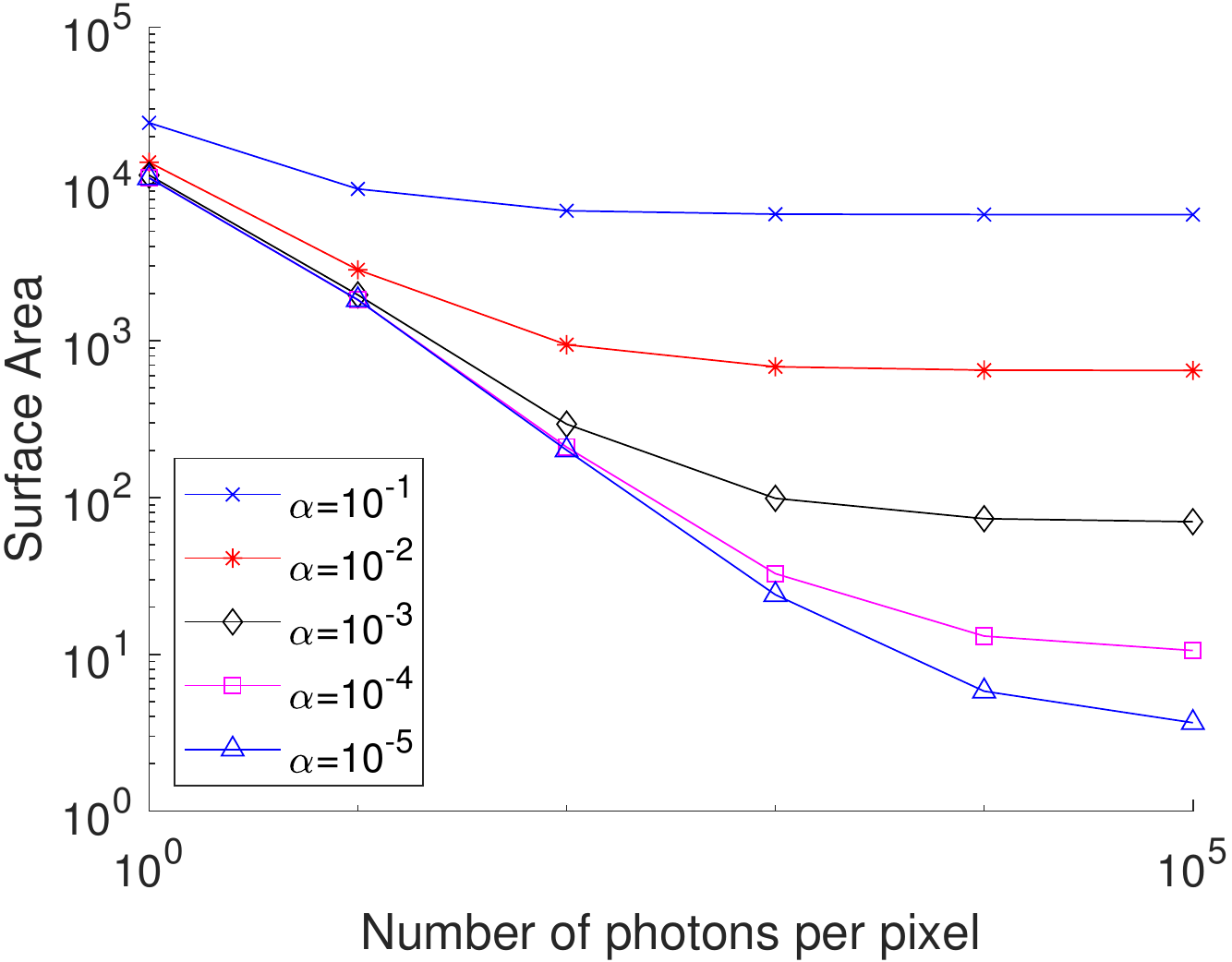}&
			\includegraphics[width=0.3\textwidth]{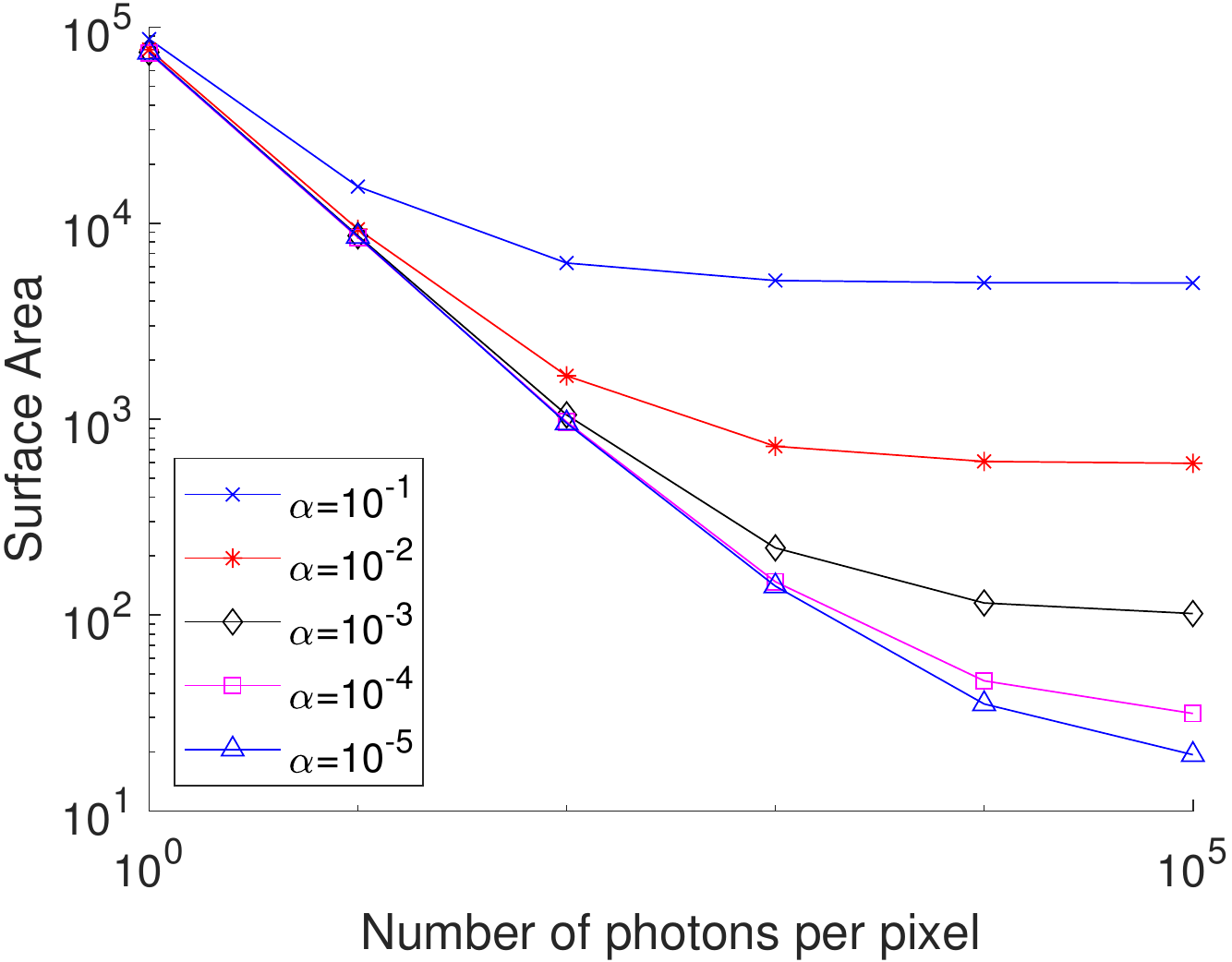}&
			\includegraphics[width=0.3\textwidth]{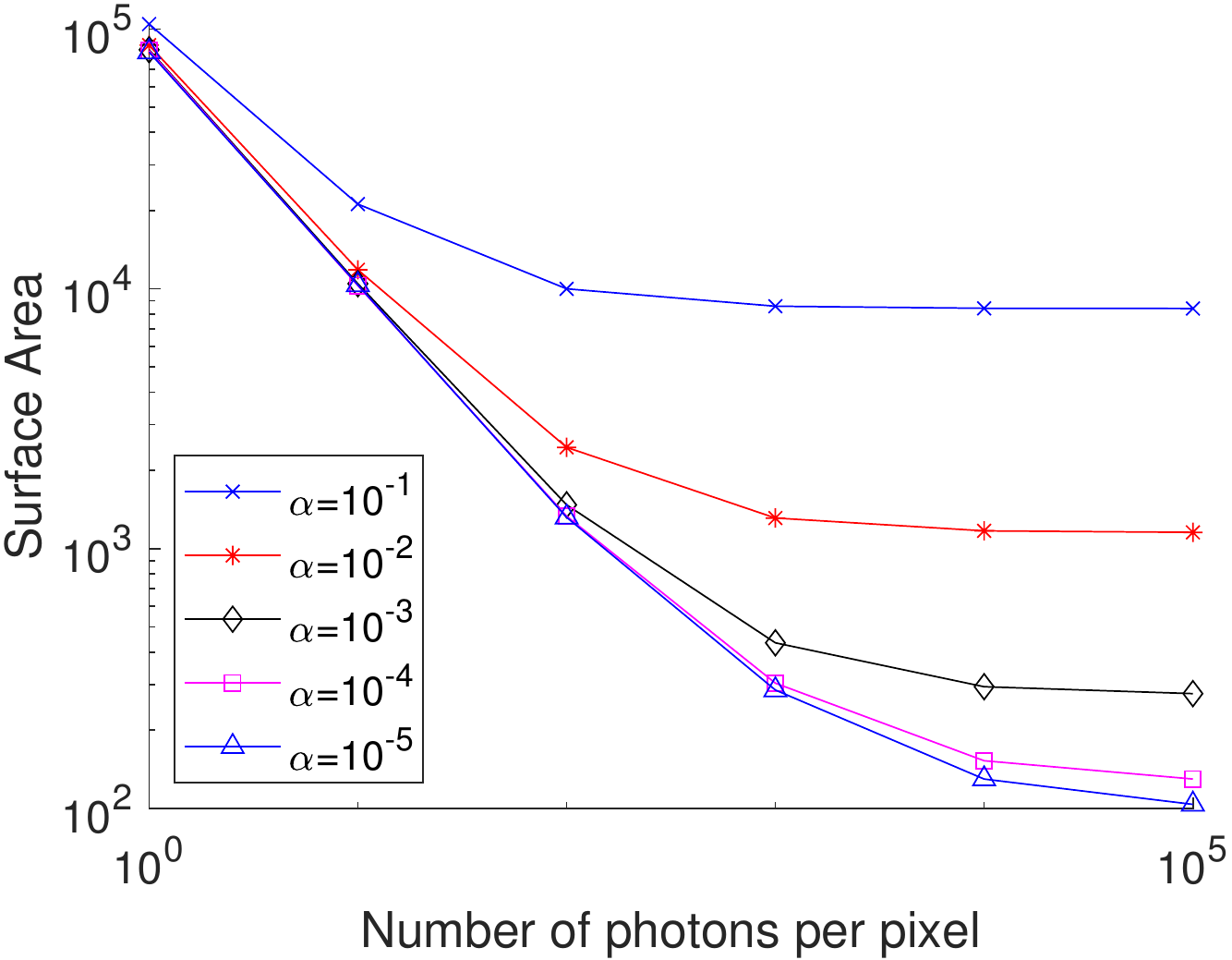}\\
			\includegraphics[width=0.3\textwidth]{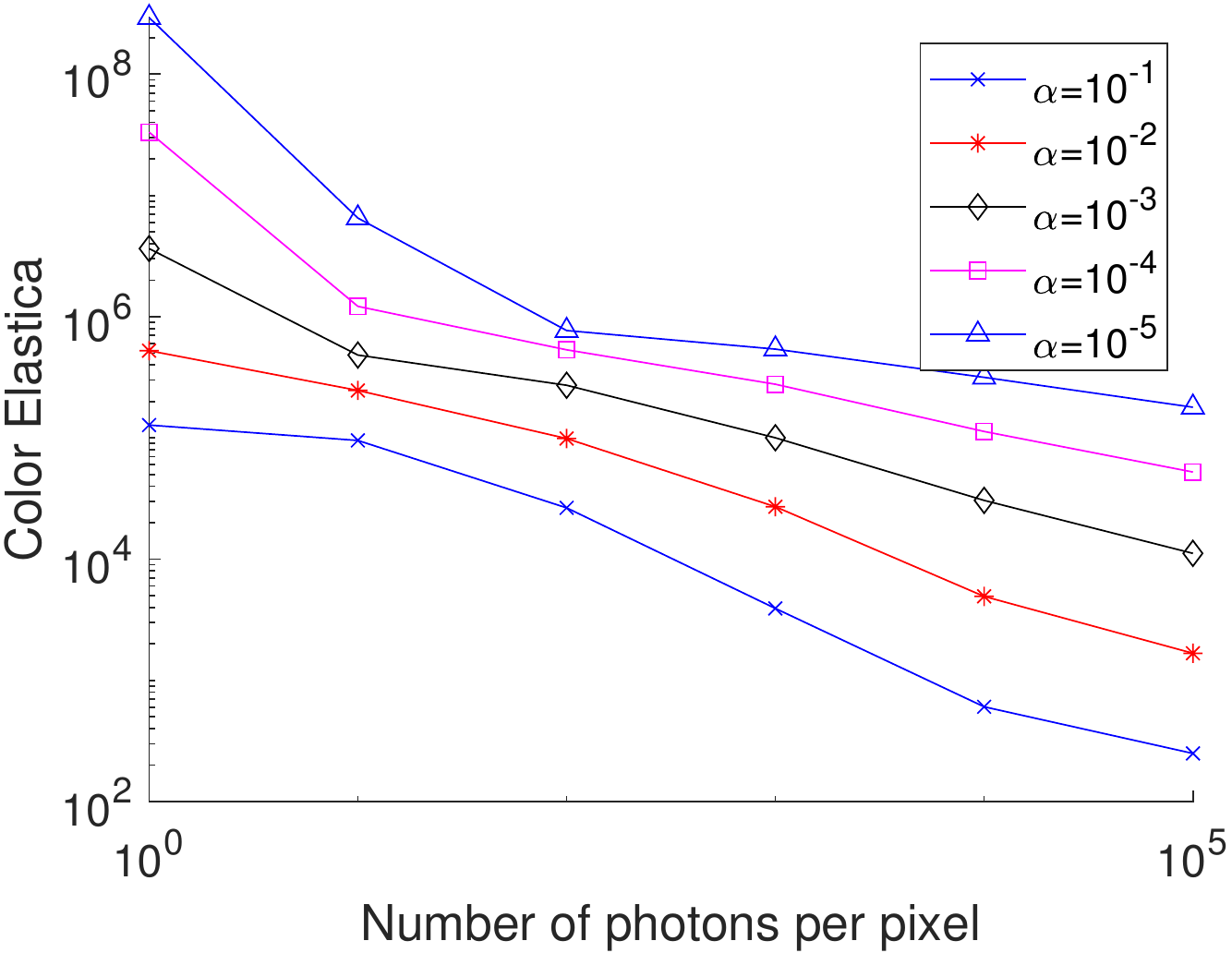}&
			\includegraphics[width=0.3\textwidth]{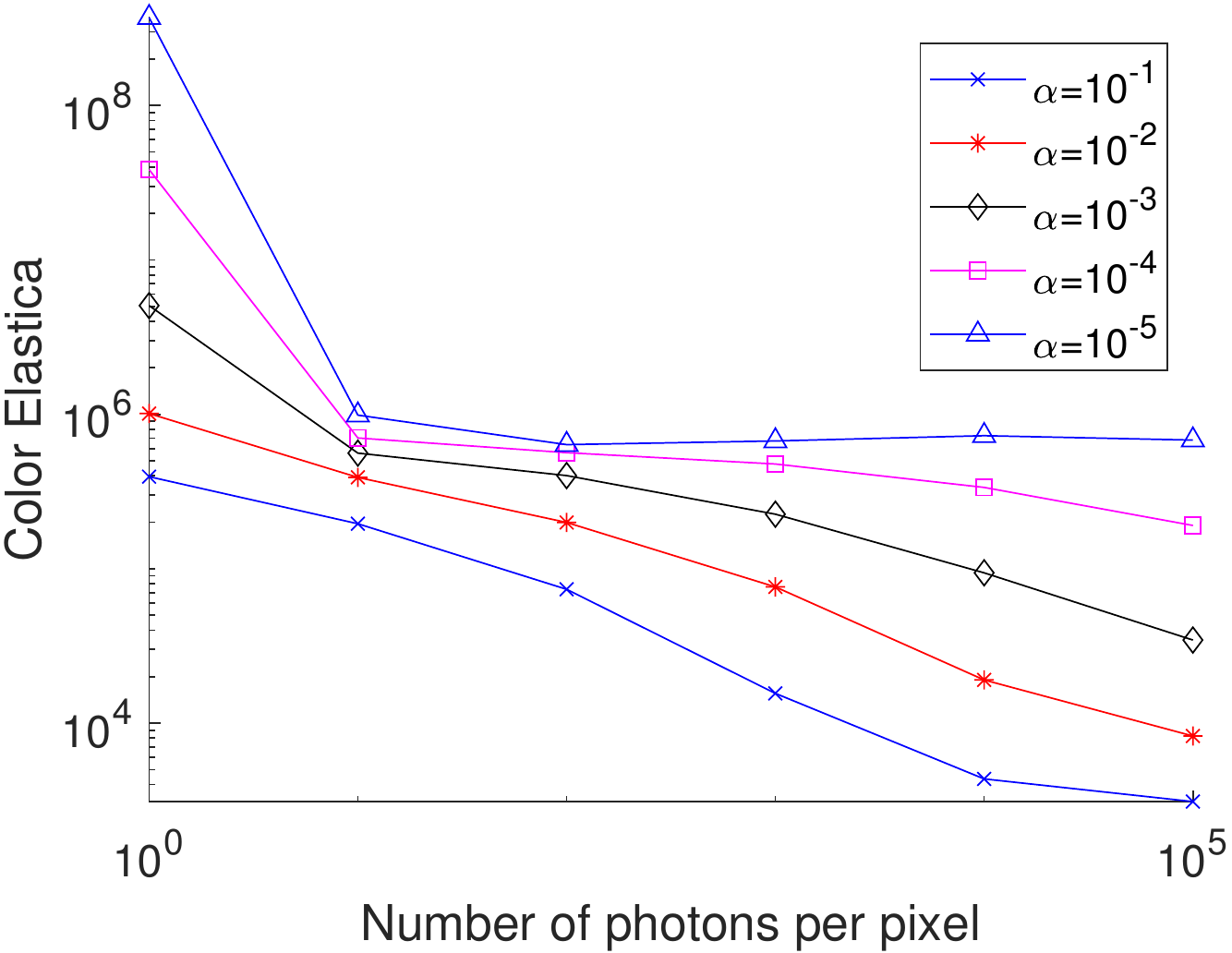}&
			\includegraphics[width=0.3\textwidth]{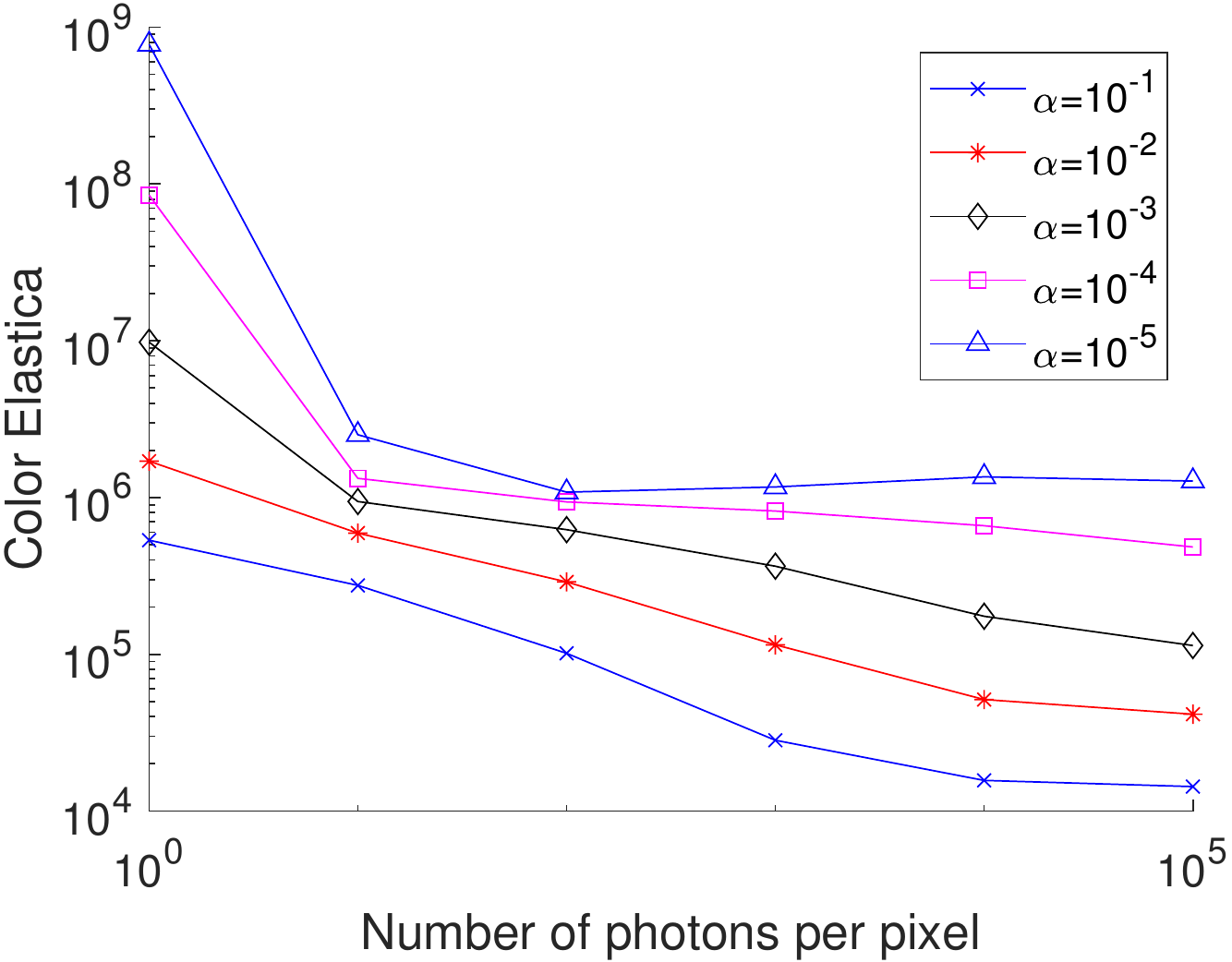}
		\end{tabular}
		\caption{(Effect of $\alpha$, Poisson noise.) With $\alpha=10^{-1},10^{-2},10^{-3},10^{-4}$ and $10^{-5}$, plot of (first row) the surface area and (second row) the color elastica computed from noisy images against $P$. (a)-(c) corresponds to the orange ball, the crystal cube and fruits shown in Figure \ref{fig.alpha.clean}, respectively. Images have higher quality with larger $P$.}
		\label{fig.alpha.P}
	\end{figure}
	
	\begin{figure}[t!]
		\centering
		\begin{tabular}{ccc}
			(a) & (b) & (c)\\
			\includegraphics[height=0.28\textwidth]{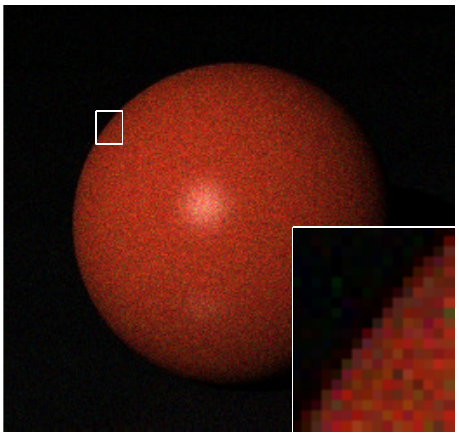}&
			\includegraphics[height=0.28\textwidth]{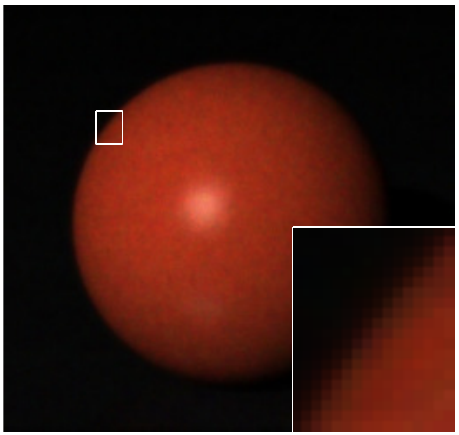}&
			\includegraphics[height=0.28\textwidth]{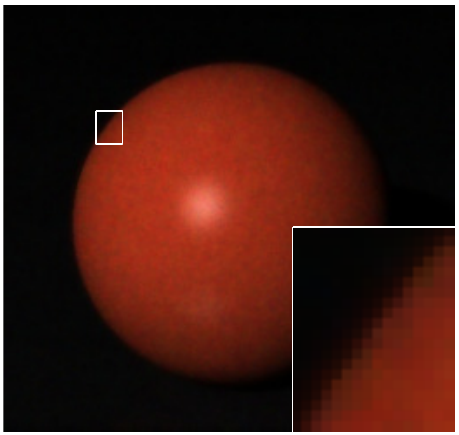}\\
			(d) & (e) & (f)\\
			\includegraphics[height=0.28\textwidth]{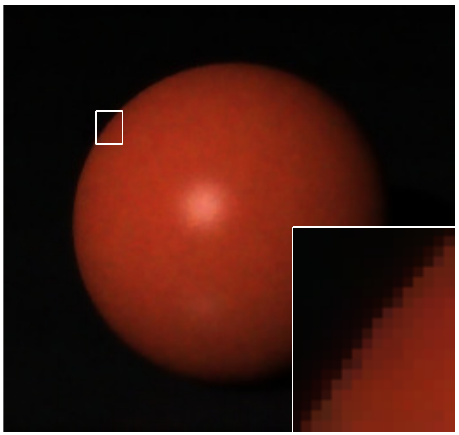}&
			\includegraphics[height=0.28\textwidth]{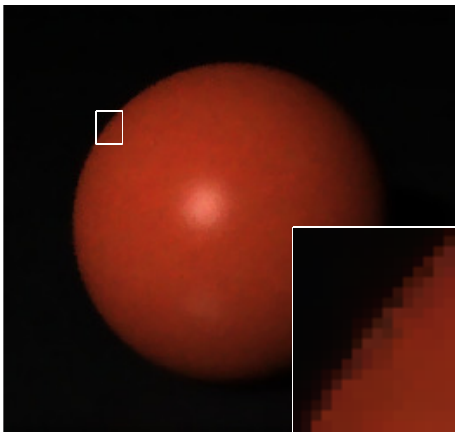}&
			\includegraphics[height=0.28\textwidth]{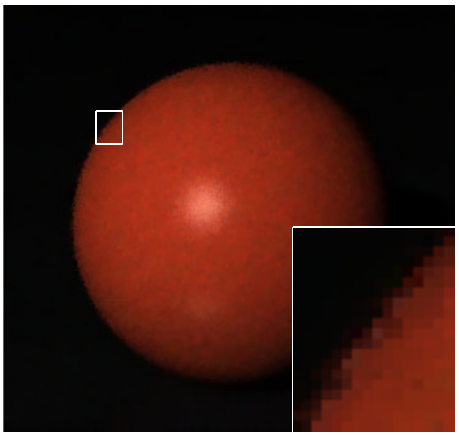}
		\end{tabular}
		\caption{(Effect of $\alpha$ with $ \beta=0$.) The image is corrupted by Poisson noise with $P=100$.  (a) is the noisy image. (b)-(f) show denoised images with (b) $\alpha=10^{-1},\eta=2$, (c) $\alpha=10^{-2},\eta=2$, (d) $\alpha=10^{-3},\eta=4$, (e) $\alpha=10^{-4},\eta=8$ and (f) $\alpha=10^{-5},\eta=10$.}\label{fig.alpha}
	\end{figure}
	
	\subsection{Effect of $\alpha$}
	The regularization term (first term) in (\ref{eq.model}) depends on two quantities: the surface area $\sqrt{g}$ and the color elastica $\sum_{k=1}^3 |\Delta_g u_k|^2$.
	Both terms are closely dependent on $\alpha$.
	In this section, we explore the effect of $\alpha$ by checking the behavior of both terms w.r.t. the noise level with $\alpha=10^{-1},10^{-2},10^{-3},10^{-4}$ and $10^{-5}$.
	The images used are shown in Figure \ref{fig.alpha.clean}: (a)  orange ball, (b) crystal cube and (c) fruits.
	
	The first experiment involves Gaussian noise for which the noise level is controlled by $SD$: images have better quality with smaller $SD$.
	In Figure \ref{fig.alpha.G}, we show the behavior of both terms w.r.t. $SD$.
	The first row shows the surface area and the second row shows the color elastica.
	Both terms have larger values on images with heavier noise (larger $SD$) for most choices of $\alpha$, which justifies the effectiveness of the proposed model (\ref{eq.model}).
	In Figure \ref{fig.alpha.G}, as SD gets larger, the surface area increases faster with smaller $\alpha$ while the color elastica increases faster with larger $\alpha$.
	To make both terms effective, $\alpha$ should not be too large or too small.
	
	We then repeat these experiments on Poisson noise in which the noise level is controlled by $P$, the number of photons.
	Images have better quality with larger $P$.
	The behavior of the surface area and energy w.r.t. $P$ with different $\alpha$ is shown in Figure \ref{fig.alpha.P}. Similar to the behavior for Gaussian noise, both terms have larger values with heavier noise (smaller $P$).
	As $\alpha$ gets larger, the surface area has larger slope while the color elastica has smaller slope, which again implies that $\alpha$ should not be too large or too small to make both terms effective.

	The above observation shows that the surface area is more effective with smaller $\alpha$. However, even if our model only contains the surface area term (i.e., $\beta=0$), we face the smoothness problem of edges if $\alpha$ is too small. In Figure \ref{fig.alpha}, we use the proposed model with  $\beta=0$ and various $\alpha$ to denoise the orange ball which is contaminated by Poisson noise with $P=100$. The noisy image, denoised image with $\alpha=10^{-1},10^{-2},10^{-3},10^{-4}$ and $10^{-5}$ are shown in (a)-(f), respectively. Since the scale of surface area changes as $\alpha$ varies, we need to set different $\eta$ for each $\alpha$. In our experiments, $\eta=2,2,4,8,10$ are used as $\alpha$ varies from $10^{-1}$ to $10^{-5}$. As $\alpha$ decreases, the edge of the orange ball in the denoised image becomes more oscillating. The cause of this observation may come from the model itself or the numerical algorithm, and is to be studied in the future.
	
	%
	
	\subsection{Effect of $\beta$}
\begin{figure}[t!]
		\centering
		\begin{tabular}{cccc}
			& (a) & (b) &\\&
			\includegraphics[width=0.22\textwidth]{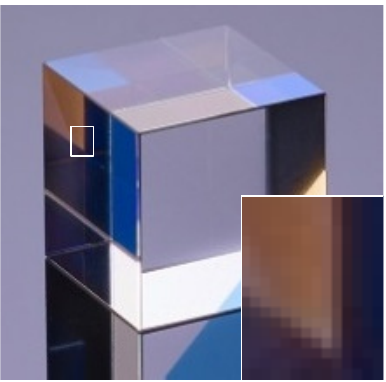}&
			\includegraphics[width=0.22\textwidth]{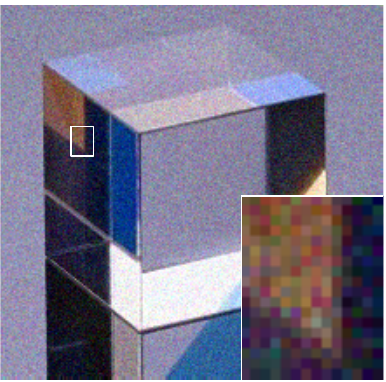}&\\
			(c) & (d) & (e)&(f)\\
			\includegraphics[width=0.22\textwidth]{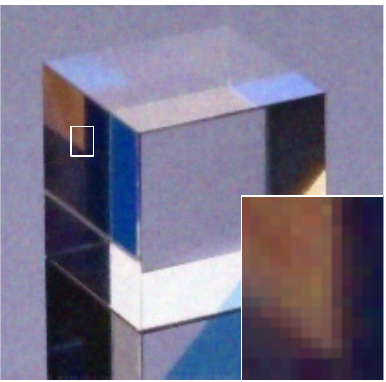}&
			\includegraphics[width=0.22\textwidth]{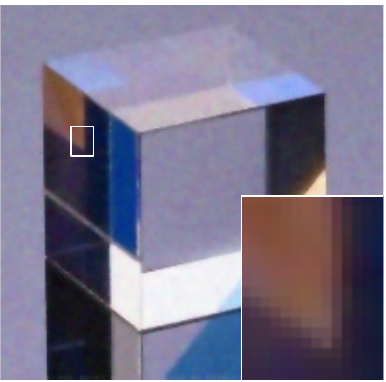}&
			\includegraphics[width=0.22\textwidth]{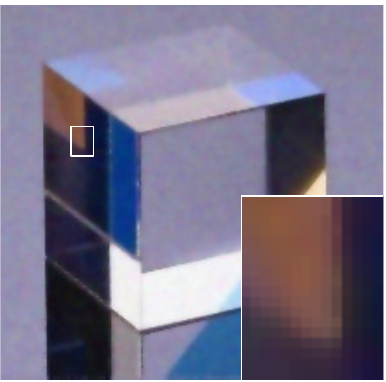}&
			\includegraphics[width=0.22\textwidth]{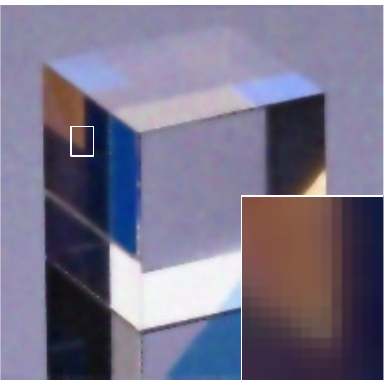}\\
			\includegraphics[width=0.22\textwidth]{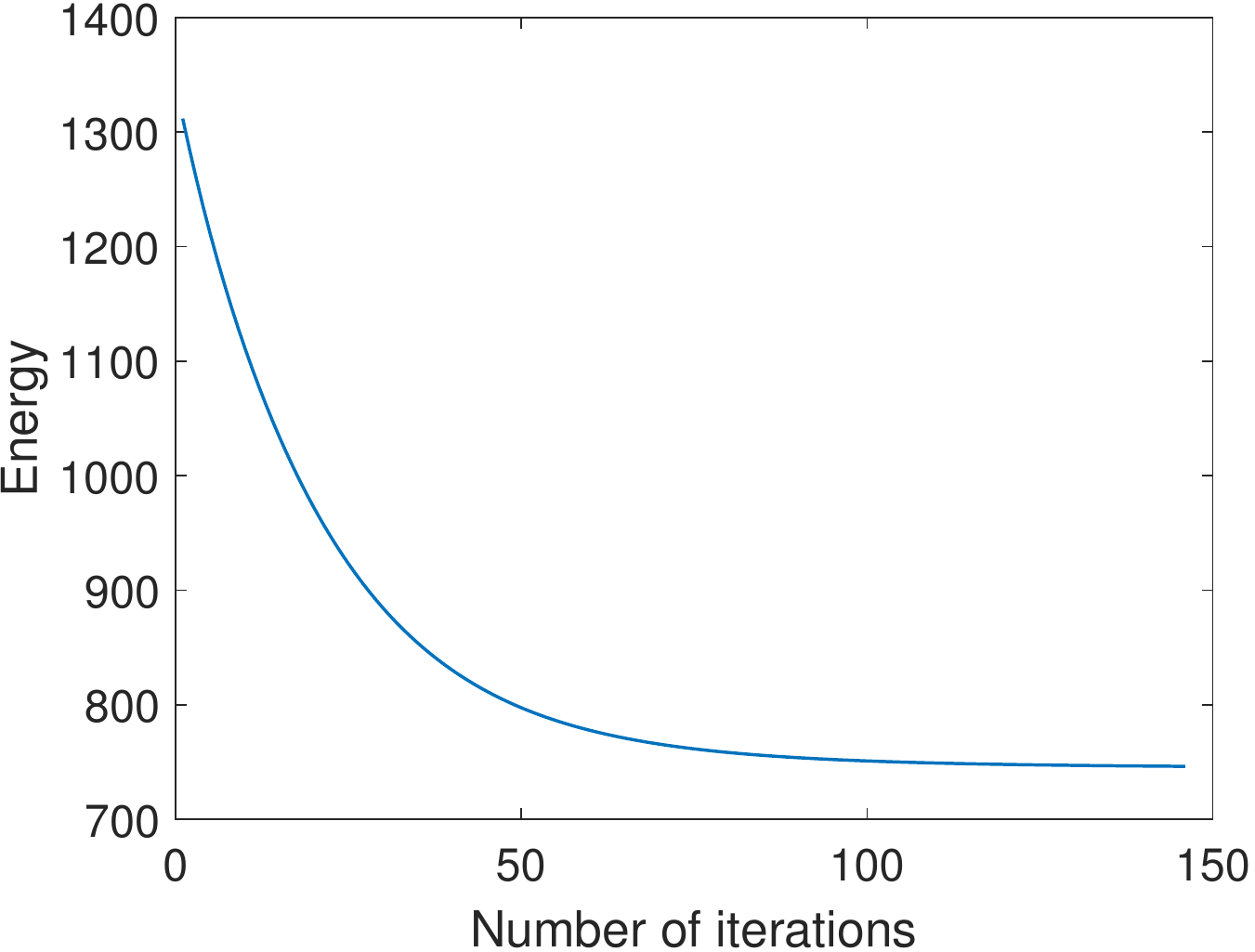}&
			\includegraphics[width=0.22\textwidth]{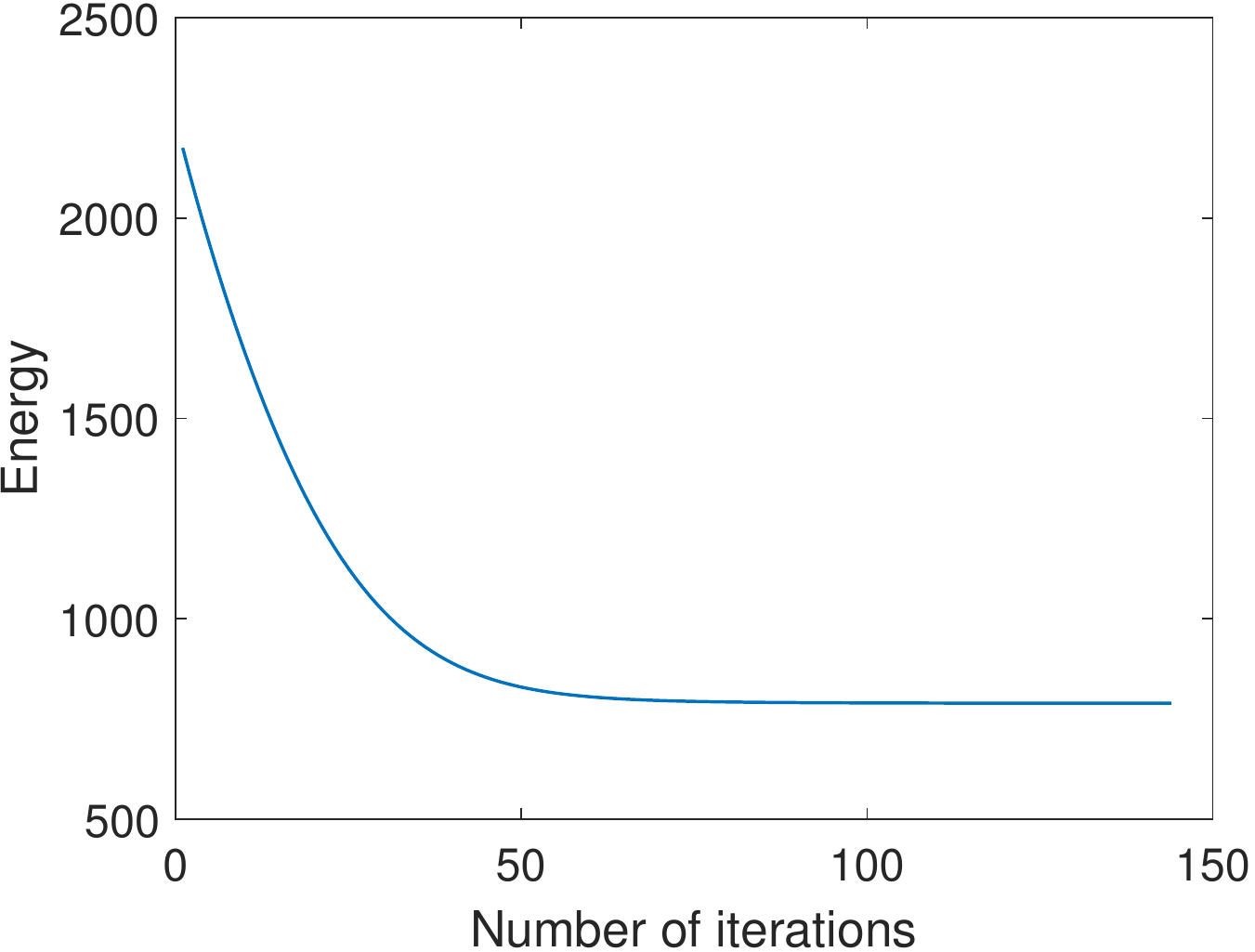}&
			\includegraphics[width=0.22\textwidth]{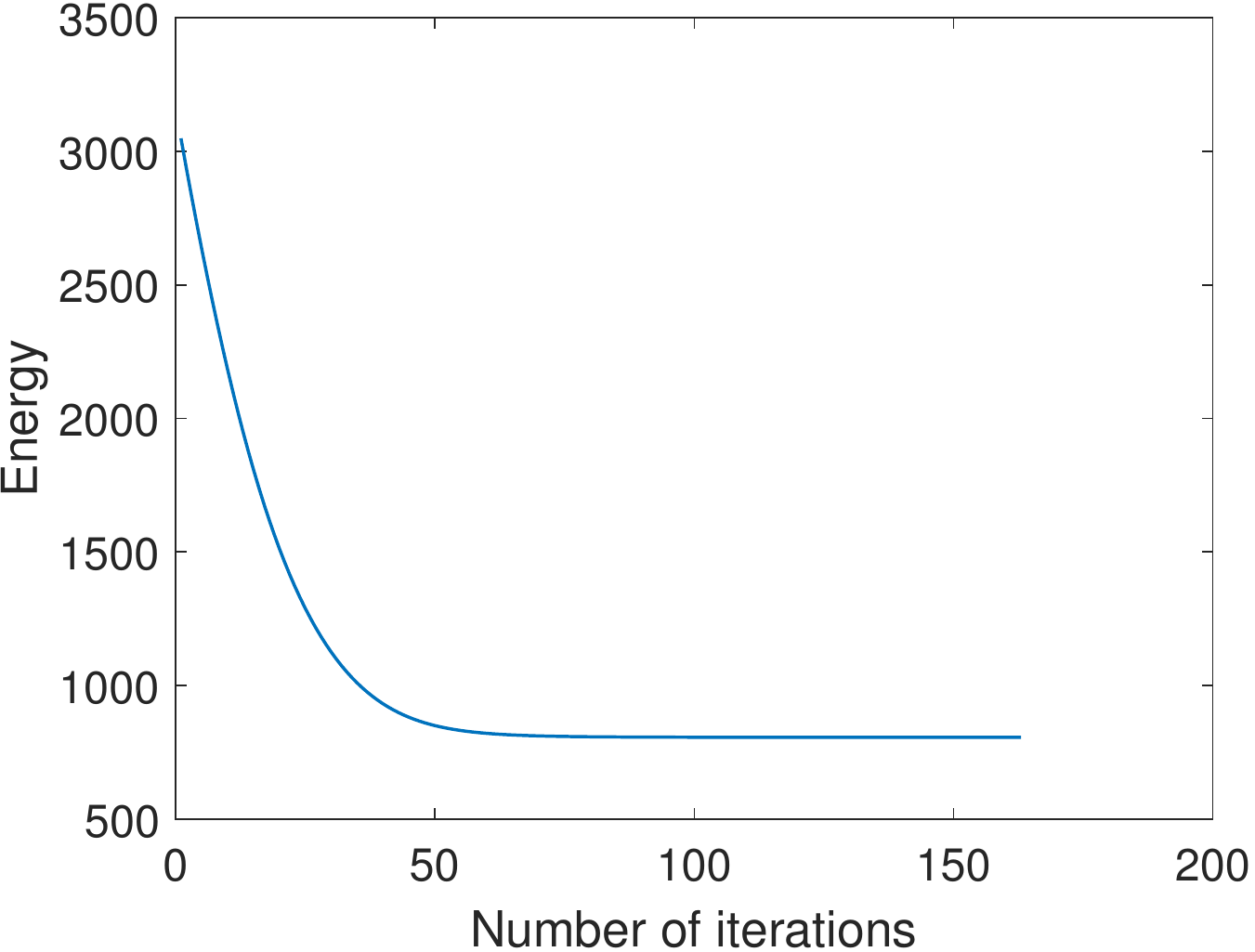}&
			\includegraphics[width=0.22\textwidth]{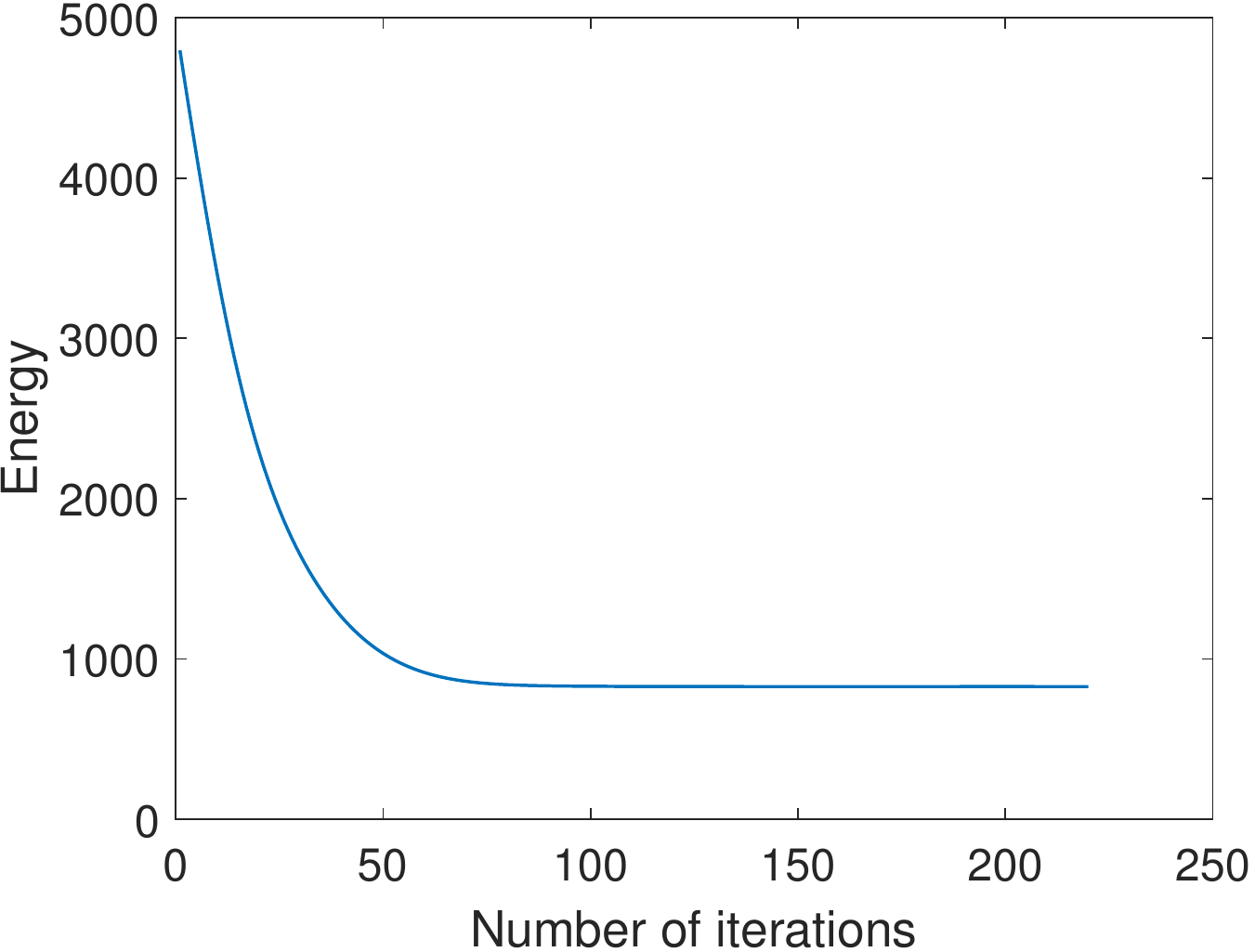}
		\end{tabular}
		\caption{(Effect of $\beta$, $\alpha=0.01,\eta=1$.) The image is corrupted by Gaussian noise with $SD=0.06$.  (a) is the clean image. (b) is the noisy image. (c)-(f) show denoised images and the evolution of the energy with (c) $\beta=0$, (d) $\beta=0.005$, (e) $\beta=0.01$, (f) $\beta=0.02$.}\label{fig.beta}
	\end{figure}

\begin{figure}[t!]
		\centering
		\begin{tabular}{cc}
			\includegraphics[height=0.28\textwidth]{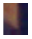}&
			\includegraphics[height=0.28\textwidth]{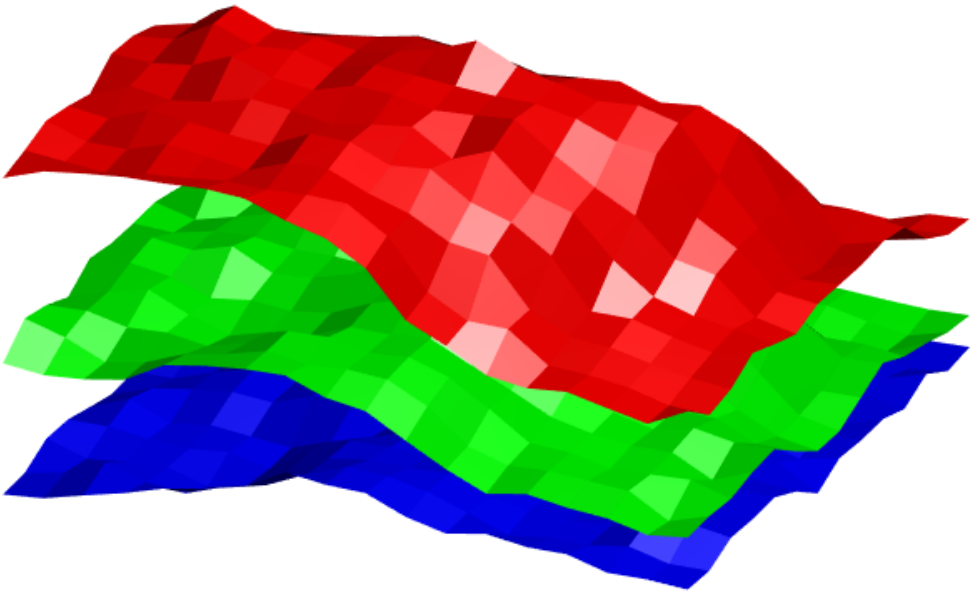}\\
			\includegraphics[height=0.28\textwidth]{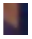}&
			\includegraphics[height=0.28\textwidth]{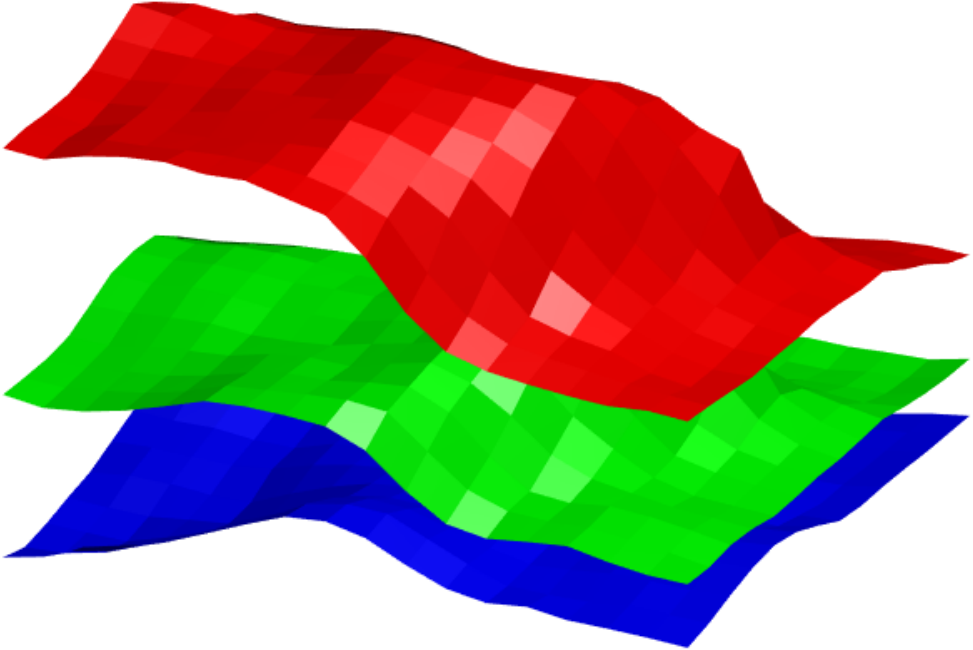}
		\end{tabular}
		\caption{(Effect of $\beta$, surface plot.) Comparison of the surface plot of results in Figure \ref{fig.beta} (c) and (d). Left: zoomed region of the denoised image in Figure \ref{fig.beta} with (first row) $\beta=0$ and (second row) $\beta=0.005$. Right: the surface plot of the left images. The RGB channels correspond to the red, green and blue surface, respectively.
		}
		\label{fig.beta.surf}
	\end{figure}

	We fix $\alpha=10^{-2},\eta=0.01$ and explore the effect of $\beta$ on our model.
	We use the image of crystal cube (shown in Figure \ref{fig.beta}(a)) with Gaussian noise with $SD=0.06$.
	We test our model with $\beta=0,0.005,0.1$ and $0.2$.
	The noisy and denoised images are shown in Figure \ref{fig.beta}.
	When $\beta=0$, model (\ref{eq.model}) reduces to the Polyakov action model and there are perturbations in the flat region of the denoised image, see Figure \ref{fig.beta}(c).
	As $\beta$ gets larger, the flat part of the denoised image becomes smoother while the edges are kept.
	Similar to the elastica model for gray-scale images, the color elastica term helps smoothing the flat region.
	In Figure \ref{fig.beta.surf} we compare the surface plot of the zoomed region of the denoised image with $\beta=0$ and $\beta=0.005$.
	The rendered surface  of the result with $\beta=0.005$ is indeed much smoother.
	The evolution of the energy are shown in the third row of Figure \ref{fig.beta}.
	The color elastica term helps the energy to achieve the minimum faster.
	With non-zero $\beta$, the energy of all experiments achieve the minimum within 80 iterations, while with $\beta=0$, it takes nearly 150 iterations for the energy to achieve its minimum.
	
	\begin{figure}[t!]
		\centering
		\begin{tabular}{ccc}
			(a) &(b) &(c)\\
			\includegraphics[width=0.3\textwidth]{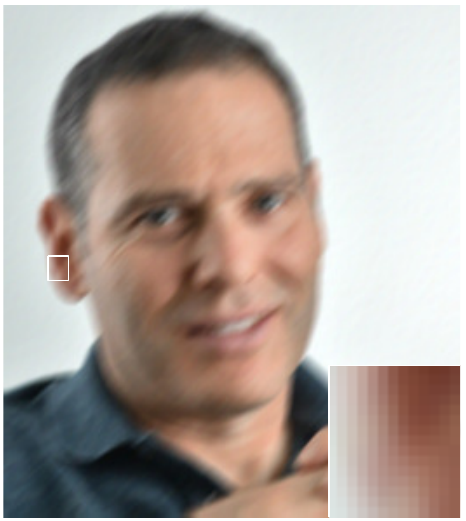}&
			\includegraphics[width=0.3\textwidth]{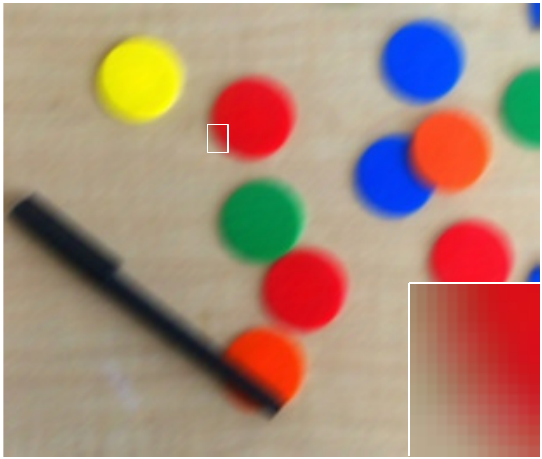}&
			\includegraphics[width=0.3\textwidth]{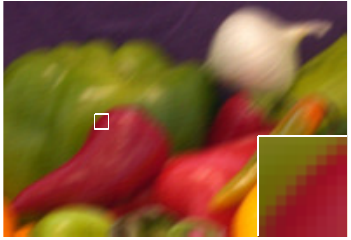}\\
			\includegraphics[width=0.3\textwidth]{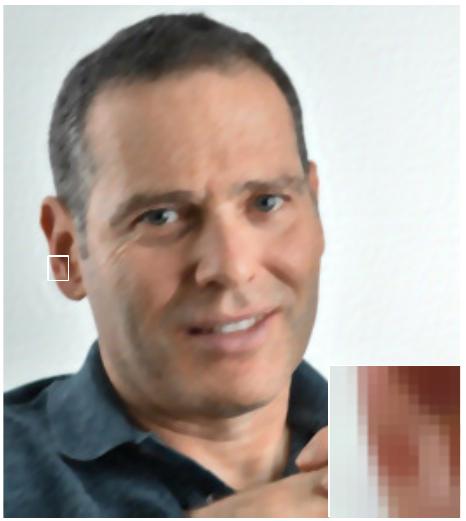}&
			\includegraphics[width=0.3\textwidth]{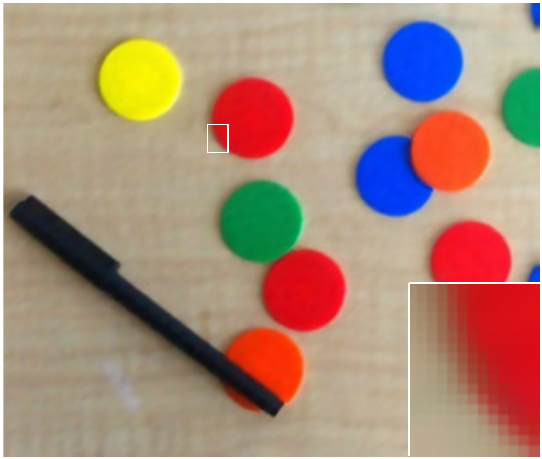}&
			\includegraphics[width=0.3\textwidth]{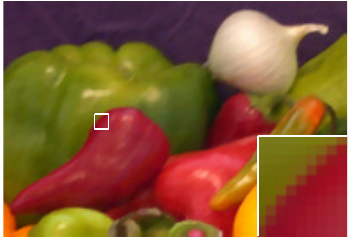}
		\end{tabular}
		\caption{(Image deblurring with $\alpha=0.001,\eta=0.05,\beta=5\times10^{-4}$.) Clean images are first blurred by motion kernel and then contaminated by noise. Image deblurred by the proposed method on (a) the portrait, (b) chips and (c) vegetables. The first row shows blurred images. The second row shows deblurred images.}
		\label{fig.blur}
	\end{figure}
	

	\subsection{An example on image deblurring}
	To conclude this section, We test the proposed model on image deblurring by minimizing (\ref{eq.model.deblurring}). The clean images are blurred by motion kernel and then contaminated by noise. In our algorithm, $\alpha=0.001,\eta=0.05,\beta=5\times10^{-4}$ is used. The clean images, blurred images and deblurred images are shown in Figure \ref{fig.blur}. The proposed model preserves color well.

	\section{Conclusion}
	\label{sec.conclusion}
	We proposed a color elastica model (\ref{eq.model}), which incorporates the Polyakov action and the squared magnitude of mean curvature of the image treated as a two dimensional manifold in spatial-color space, to regularize color images.
	The proposed model is a geometric extension of the elastica model (\ref{eq.greyelastica}): when applied to gray-scale images, it reduces to a variant of the Euler elastica model.
	We also proposed an operator-splitting method to solve (\ref{eq.model}).
	The proposed method is efficient and robust with respect to parameter choices.
	The effectiveness of the proposed model and the efficiency of the numerical method are demonstrated by several experiments in color image regularization.
	The mean curvature term helps to better smooth noisy almost uniform (yet with some color changes) regions in images while keeping the edges sharp.
	
		\section*{Acknowledgment} The authors would like to thank the anonymous reviewers of this article for helpful comments and suggestions.

	\bibliographystyle{siamplain}
	\bibliography{references}
	
\end{document}